\documentclass[10pt,journal,compsoc]{IEEEtran}

\ifCLASSOPTIONcompsoc
  \usepackage[nocompress]{cite}
\else
  \usepackage{cite}
\fi
\usepackage{paralist}
\usepackage{pifont}
\usepackage{listings}
\usepackage{amssymb}
\usepackage{times}
\usepackage{latexsym}
\usepackage{graphicx}
\usepackage{threeparttable}
\usepackage{multirow}
\usepackage{color}
\usepackage{algorithm}
\usepackage{algorithmic}
\usepackage{newfloat}
\usepackage{listings}
\usepackage{helvet}  
\usepackage{courier}  
\usepackage{graphicx} 
\usepackage{verbatim}
\usepackage{subfigure}
\usepackage{svg}
\usepackage{bm}
\usepackage{pifont}
\usepackage{amsthm}
\usepackage{amssymb}
\usepackage{amsmath}

\usepackage{changes} 
\definechangesauthor[name={Bin Ji}, color=red]{S.} 
\usepackage{booktabs}
\def\mathbi#1{\textbf{\em #1}}
\usepackage{colortbl}  
\usepackage{xcolor}
\usepackage{array} 
\newcommand{\red}[1]{\textcolor{red}{#1}}
\newcommand{\blue}[1]{\textcolor{blue}{#1}}
\ifCLASSINFOpdf
\else
\fi

\begin{document}

\title{Win-Win Cooperation: Bundling Sequence and Span Models for Named Entity Recognition}

\author{Bin Ji,~\IEEEmembership{Member,~IEEE,}
				Jing Yang,
				Shasha Li,~\IEEEmembership{Member,~IEEE,}
        		Jie Yu,~\IEEEmembership{Member,~IEEE,}   
        		Jun Ma,~
				Huijun Liu~
		 % <-this % stops a space
\IEEEcompsocitemizethanks{\IEEEcompsocthanksitem B. Ji, S. Li, J. Yu, J. Ma, and H. Liu are with the College of Computer, National University of Defense Technology, Changsha 410073, China.\protect\\
E-mail: \{jibin, yangjing, shashali, yj, majun, liuhuijun\}@nudt.edu.cn.
%\IEEEcompsocthanksitem J. Doe and J. Doe are with Anonymous University.
}% <-this % stops a space
\thanks{Manuscript received July xx, 2022; revised xx xx, 2022.}}

\markboth{Journal of \LaTeX\ Class Files,~Vol.~14, No.~8, August~2015}%
{Shell \MakeLowercase{\textit{et al.}}: Bare Advanced Demo of IEEEtran.cls for IEEE Computer Society Journals}

\IEEEtitleabstractindextext{%
\begin{abstract}
For Named Entity Recognition (NER), sequence labeling-based and span-based paradigms are quite different. Previous research has demonstrated that the two paradigms have clear complementary advantages, but few models have attempted to leverage these advantages in a single NER model as far as we know.
In our previous work, we proposed a paradigm known as Bundling Learning (BL) to address the above problem. The BL paradigm bundles the two NER paradigms, enabling NER models to jointly tune their parameters by weighted summing each paradigm's training loss. However, three critical issues remain unresolved: When does BL work? Why does BL work? Can BL enhance the existing state-of-the-art (SOTA) NER models? 
To address the first two issues, we implement three NER models, involving a sequence labeling-based model--SeqNER, a span-based NER model--SpanNER, and BL-NER that bundles SeqNER and SpanNER together. We draw two conclusions regarding the two issues based on the experimental results on eleven NER datasets from five domains. 
We then apply BL to five existing SOTA NER models to investigate the third issue, consisting of three sequence labeling-based models and two span-based models. Experimental results indicate that BL consistently enhances their performance, suggesting that it is possible to construct a new SOTA NER system by incorporating BL into the current SOTA system. 
Moreover, we find that BL reduces both entity boundary and type prediction errors.
In addition, we compare two commonly used labeling tagging methods as well as three types of span semantic representations. 
\end{abstract}

}

\maketitle

\IEEEdisplaynontitleabstractindextext

\IEEEpeerreviewmaketitle

\ifCLASSOPTIONcompsoc
\IEEEraisesectionheading{\section{Introduction}\label{sec:introduction}}
\else
\section{Introduction}
\label{sec:introduction}
\fi

\begin{figure}[]
\centering
\includegraphics[width=0.49\textwidth]{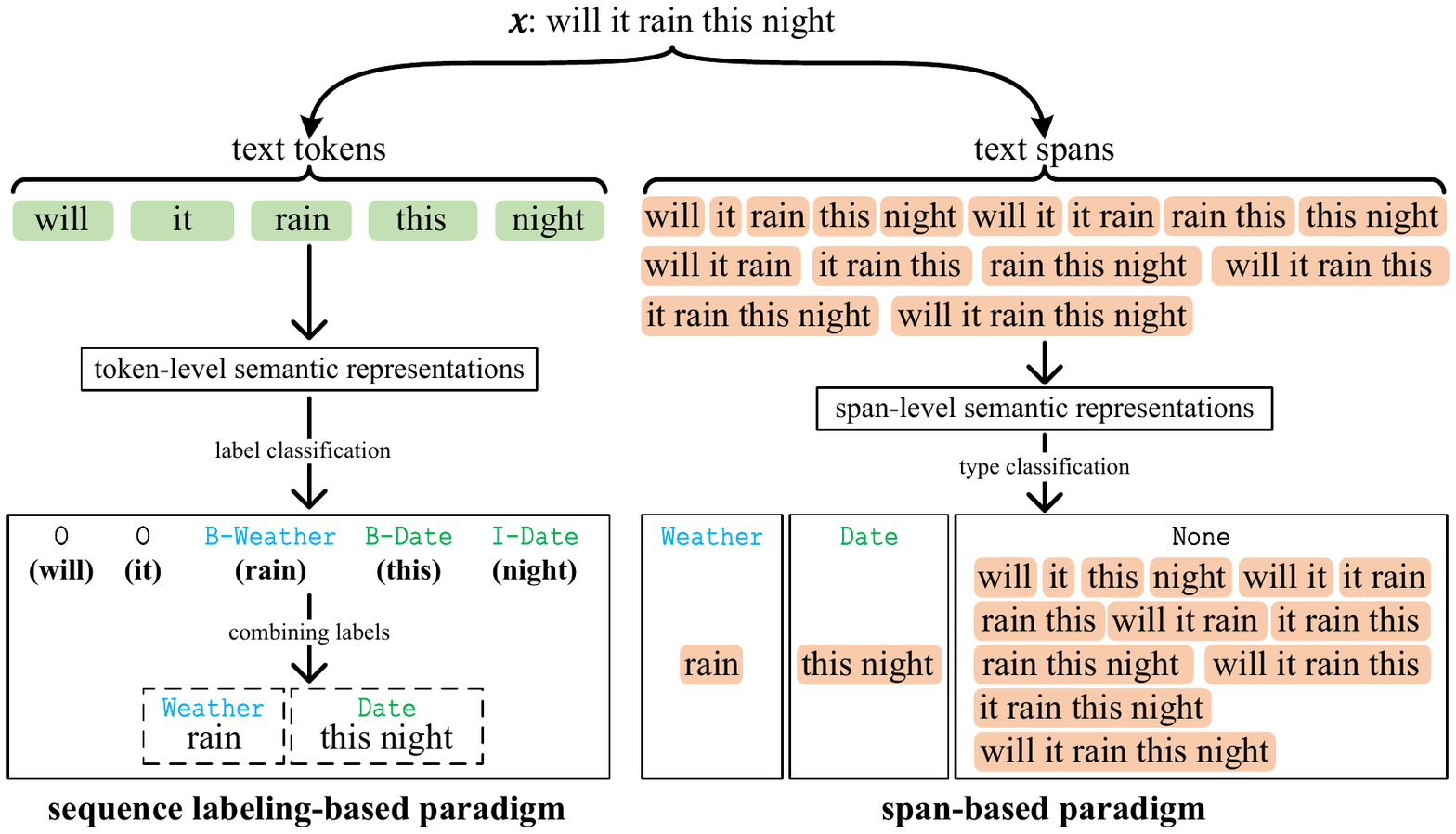} 
\caption{Comparisons of sequence labeling-based and span-based paradigms for the NER task. The {\texttt{B-Weather}}, {\texttt{B-Date}}, {\texttt{I-Date}}, and \texttt{O} are BIO labels, {\texttt{Weather}} and {\texttt{Date}} are two pre-defined entity types. \texttt{None} is an added type for spans that are not entities. {``rain''} is a \texttt{Weather} entity and {``this night''} is a \texttt{Date} entity.}\label{figure1}
\end{figure}

\IEEEPARstart{N}amed entity recognition (NER) is a fundamental task of Natural Language Processing (NLP) and is a precursor to many downstream NLP tasks such as relation extraction \cite{ji} and coreference resolution \cite{crqa}.
The NER task currently involves two mainstream technical routines: sequence labeling-based paradigm \cite{huang,mi_ba,ka_ca,flat,saner} and span-based paradigm \cite{tanqiu,lifei,ouchi,spanner,generic,hamner,Toshniwal}. 
As Figure \ref{figure1} shows, the sequence labeling-based paradigm formulates the problem as a sequence labeling task,  in which each text token is tagged with a label based on the token-level representation, and entities are derived from these labels. 
By contrast, the span-based paradigm formulates the problem as a span prediction task, which considers text spans as candidate entities and determines span types directly by carrying out type classifications on span-level representations.

Previous work \cite{spanner} has proven that the two NER paradigms have clear complementary advantages. For example, the sequence labeling-based paradigm performs better when entities are long, while the span-based paradigm is better at dealing with sentences with more Out-of-Vocabulary (OOV) words. However, few studies attempt to leverage these complementary advantages within a single NER model except for our recent study \cite{ji_kbs}.
In the study, we propose a novel Bundling Learning (BL) paradigm for the NER task. BL bundles sequence labeling-based and span-based models by (1) making them share a common encoding layer while keeping their decoding layers unchanged and (2) weighted summing their training losses to jointly tune the shared layer. Consequently, the BL paradigm permits us to model the NER task from both sequence labeling and span prediction perspectives, allowing both token-level and span-level semantic representations to be used simultaneously.
We apply the BL paradigm to two previously published models, namely SANER (a sequence labeling-based NER model) \cite{saner} and SpERT (a span-based joint entity-relation extraction model)  \cite{spert}.\footnote{For SANER, we bundle a span-based NER model with it. For SpERT, we bundle a sequence labeling-based NER model with its NER module.} Experimental results show that BL consistently improves the NER performance of SANER and SpERT. Nevertheless, we remain unexplored on three critical issues: (1) When does BL work? (2) Why does BL work? In other words, what exhaustive complementary advantages can BL leverage? (3) Can BL enhance the existing state-of-the-art (SOTA) NER models?

We investigate the above issues in this paper. To be specific, for issues (1) and (2), we first implement a span-based NER model--SpanNER and a sequence labeling-based NER model--SeqNER, and then we bundle them together to obtain a BL enhanced model--BL-NER. Next, We conduct detailed analyses with these models on eleven NER datasets. For issue (3), we examine whether BL can enhance the performance of five existing SOTA NER models. Moreover, we compare SeqNER, SpanNER, and BL-NER from the perspective of entity prediction error, which includes entity boundary error and entity type error.
In addition, we examine the effectiveness of CRF-based and softmax-based label tagging. As well as this, we investigate three different methods for obtaining span-level semantic representations, namely boundary, span-pooling, and hybrid.

Experimental results indicate that: (1) BL-NER outperforms both SeqNER and SpanNER when the two NER models perform closely on NER datasets. When the performance gap between the two NER models is large, BL-NER surpasses the NER model that performs worse but has disadvantages over the other model. (2) BL enables BL-NER to leverage the relative advantages of SeqNER and SpanNER when it comes to four attributes: entity length, text length, entity label consistency, and entity density. Additionally, the relative disadvantages of the two NER models lead to BL-NER's performance dropping on some attributes. (3) BL consistently produces performance gains in the five models, suggesting that BL may be used to construct a new SOTA NER system by applying it to the current SOTA system. In addition, qualitative experiments reveal that SeqNER generally suffers fewer entity boundary prediction errors than SpanNER, but more entity type prediction errors. It is a good thing that BL reduces both types of errors.

In conclusion, we summarize the contributions as follows:
(1) We revisit our proposed BL paradigm and clarify when it works. (2) We examine the exhaustive advantages that BL can leverage in detail. (3) We suggest a possible method to obtain new SOTA NER systems by applying BL to the existing SOTA systems.

\section{Related Work}
We roughly divide the majority of neural models for NER  into two categories: sequence labeling-based and span-based models, where the former adopts the sequence labeling-based paradigm and the latter uses the span-based paradigm.

\subsection{Sequence Labeling-based NER Models}
Sequence labeling-based models formulate NER as a sequence labeling task. 
Almost all these models adopt the encoder-decoder architecture, where the encoder encodes a token embedding for each text token, and the decoder tags each token with a label according to its embedding. They then obtain entities based on token labels.
Moreover, they generally use the Conditional Random Fields (CRF) based decoder, for CRF has proven to produce higher tagging accuracy in general \cite{huang}.  
Furthermore, they make attempt to use carefully designed neural architectures as their encoders, involving Recurrent Neural Network (RNN) \cite{linb,zheng,xin2018learning,martins2019joint}, Convolutional Neural Network (CNN) \cite{mrmep,12,strubell2017fast}, Transformer Encoder \cite{tener,hongbin,ushio2021t}, and Graph Neural Network (GNN) \cite{sui,carbonell,zhou2022panner}, with the goal of generate better token embeddings. 
Recently, Pre-trained Language Models (PLMs), such as ELMo \cite{peters},  BERT \cite{bert}, RoBERTa \cite{roberta}, and ALBERT \cite{albert}, are applied to this research field. 
These PLMs are taken as the encoders of a number of NER models \cite{souza,sapci2021focusing} directly due to their excellent encoding ability. In addition, some models make attempts to combine PLMs and other neural models as their encoders, such as ELMo+LSTM \cite{peters,jie2019dependency,dai2019using} and BERT+LSTM \cite{xu2021better}. 
In spite of the success of the above models, we demonstrate that they cannot benefit from the advantages of span-based models, such as the ability to recognize medium-length entities with higher accuracy. By contrast, the proposed BL paradigm makes it possible to leverage these advantages in sequence labeling-based models.

\subsection{Span-based NER Models}
It is a natural defect that sequence labeling-based models cannot deal with overlapped entities, for they are only allowed to tag each text token with one label. Span-based models are targeted in the scenario of recognizing overlapped entities, for they consider text spans as candidate entities that allow entity overlapping.
Sohrab et al. \cite{sohrab} propose the first span-based NER model as far as we know. The key idea of their model is to enumerate all possible spans as potential entities and class them with deep neural networks, where these networks first generate span representations and then use linear classifiers to classify them. The key idea has become a standard followed by the latter span-based models.
Fu et al. \cite{spanner} introduce BERT into their span-based model, and they further use the BiLSTM to obtain better span representations. Moreover, they investigate using their span-based model as a NER model combiner.
Li et al. \cite{lifei} propose a span-based model similar to Fu et al. \cite{spanner}. The difference is that they use the Attention-Guided Graph Convolutional Network (AGGCN) \cite{gcn} rather than BiLSTM to obtain better span representations. Moreover, their model allows double-checking overlapped entities and detecting discontinuous entities. 
Tan et al. \cite{tan} demonstrate that previous span-based models usually perform poorly in entity boundary detection. Thus they propose to train a boundary detection model and a span-based model jointly, where both models are built upon BERT embeddings. They consider the boundary and span classification results when determining whether a span is an entity.
Yu et al. \cite{yujie} propose a span-based model to deal with the problem of cascading label prediction errors that exist in sequence labeling-based models.
Moreover, Ouchi et al. \cite{ouchi} propose an instance-based model. At inference time, their model first enumerates all spans and then assigns a class label to each span based on its similar spans in the training set.
Yu et al. \cite{yujun} use a biaffine \cite{biaffine} model to score pairs of start and end tokens in a sentence, where text spans are restricted by these start-end pairs. Consequently, their model predicts whether these spans are entities according to the scores.
Additionally, span-based models for joint entity-relation extraction \cite{acl2019,coreference,dygie,dygie++,spert,ji} have been extensively studied, and their NER modules comply with the above span standard.
Span-based models have shown their effectiveness in the NER task. However, we demonstrate that it is hard for these models to benefit from advantages brought by sequence labeling-based models, such as better recognition performance for long entities.
Compared to them, the proposed BL paradigm enables span-based models to leverage these advantages.

\section{Models}

In this section, we first implement a BL enhanced NER model--BL-NER. To this end, we propose a sequence labeling-based NER model--SeqNER and a span-based NER model--SpanNER, then we obtain BL-NER by bundling the two models together (\textbf{Section  3.1}).
Next, we present a model discussion in \textbf{Section 3.2}, which aims to provide a better understanding of the BL paradigm.

\subsection{BL-NER}\label{section3.1}

\begin{figure}[]
\centering
\includegraphics[width=0.3\textwidth]{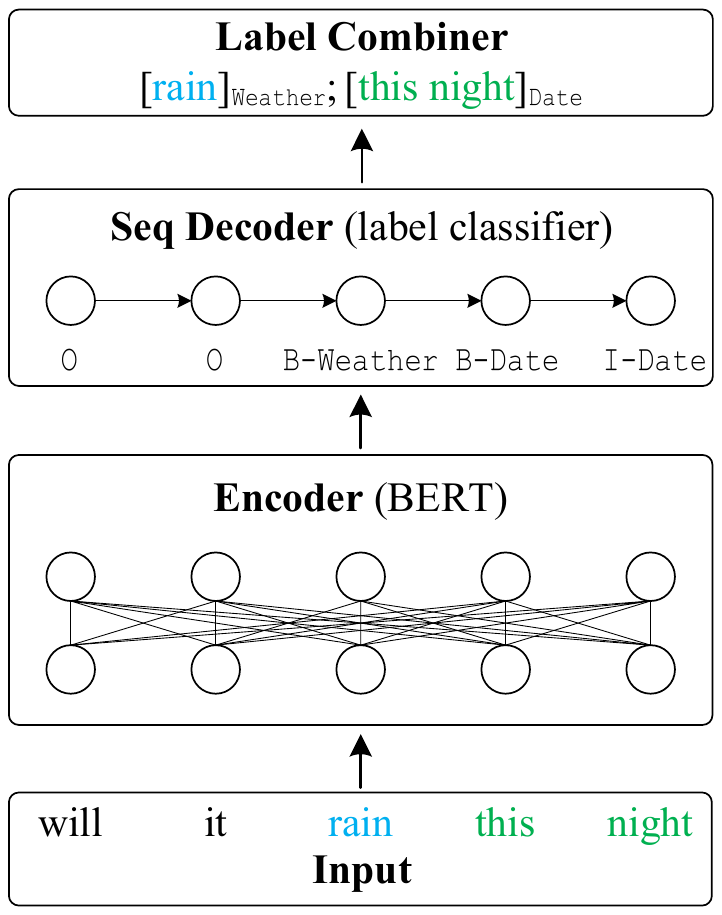} 
\caption{The architecture of SeqNER. Given an input text, the encoder first obtains the contextualized embeddings. Then the decoder tags each text token with a label according to the token embedding. For example, it tags the ``{will}'' with the label ``\texttt{O}'' and the ``{rain}'' with the label ``\texttt{B-Weather}''. At last, the Label Combiner obtains entities based on the token labels.}\label{seqner}
\end{figure}

\begin{figure}[]
\centering
\includegraphics[width=0.3\textwidth]{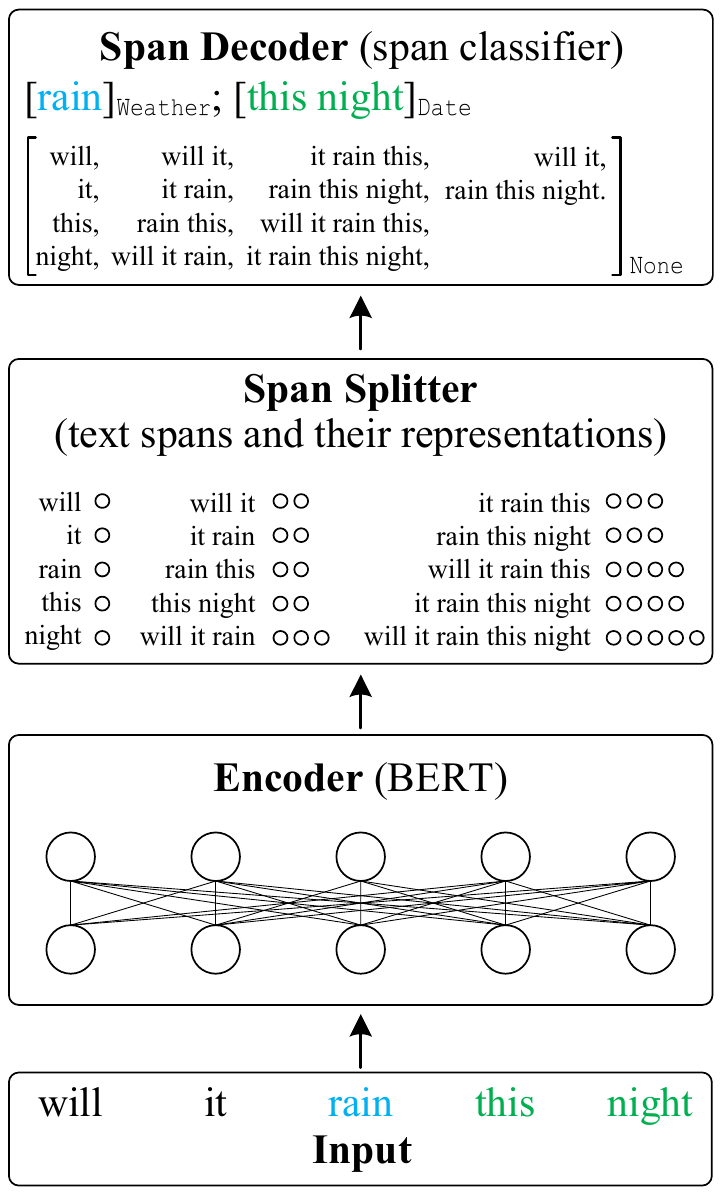} 
\caption{The architecture of SpanNER. Given an input text, the encoder first obtains its contextualized embeddings. Then the Span Spliter enumerates all text spans and obtain their semantic representations. At last, the Decoder conducts classifications on these representations. For example, the ``{rain}'' is classified as a ``\texttt{Weather}'' entity, the ``{this night}'' is classified as a ``\texttt{Date}'' entity, and all the other spans are classified as the ``\texttt{None}'' type.}\label{spanner}
\end{figure}

As mentioned above, we obtain BL-NER by bundling SeqNER and SpanNER. Therefore, we begin by illustrating the implementations of the two bundled models. As shown in Figure \ref{seqner}, Figure \ref{spanner}, and Figure \ref{bl-ner}, these three models all share a common model encoder, so we begin by illustrating the implementation of the encoder, followed by the two bundled models, and finally, BL-NER.

\subsubsection{The Common Encoder}\label{section3.1.1}
We use the BERT model \cite{bert} as the common encoder, for it can produce contextualized embeddings for input texts.
Given an input text $\mathcal{T}=t_0t_1t_2...t_n$, where $t_0$ is the specific [CLS] token required by BERT and $t_i$ denotes the $i$-$th$ text token, we pass it to the BERT model.
For each $t_i$ in $\mathcal{T}$, BERT first tokenizes it into several sub-tokens with the WordPiece vocabulary \cite{wordpiece} to avoid the Out-of-Vocabulary (OOV) problem. BERT therefore produces a sub-token sequence:
\begin{equation}
\mathcal{T}'=t_0t_{1}^1...t_{1}^jt_{2}^1...t_{2}^j......t_{n}^1...t_{n}^j, 
\end{equation}
where $t_i^j$ denotes the $j$-$th$ sub-token of the $i$-$th$ token and the value of $j$ varies from token to token. 

For each sub-token in $\mathcal{T}'$, its representation is the element-wise addition of WordPiece embedding, positional embedding, and segment embedding. Then a list of input embeddings $\mathbi{H} \in \mathbb{R}^{l \ast h}$ are obtained, where \textit{l} is the sequence length of $\mathcal{T}'$ and \textit{h} is the size of hidden units. A series of pre-trained Transformer \cite{transformer} blocks then project $\mathbi{H}$ into a BERT embedding matrix (denoted as $\mathbi{E}_{\mathcal{T}'}$):
\begin{equation}
{\mathbi{E}_{\mathcal{T}'} = [\mathbi{e}_0, \mathbi{e}_1^1,..., \mathbi{e}_1^j, \mathbi{e}_2^1,..., \mathbi{e}_2^j,......,  \mathbi{e}_{n}^1,...,\mathbi{e}_n^j]},
\end{equation}
where $\mathbi{E}_{\mathcal{T}'} \in \mathbb{R}^{(l*d)}$ and $d$ is the BERT embedding dimension.

For each $t_i$ in  $\mathcal{T}$, we obtain its BERT embedding $\mathbi{e}_i$ by applying the max-pooling function to the BERT embeddings of its sub-tokens:
\begin{equation}
\mathbi{e}_i = \textrm{max-pooling}\{\mathbi{e}_i^1,..., \mathbi{e}_i^j\},
\end{equation}
where $\mathbi{e}_i \in \mathbb{R}^{d}$.

Then we denote the BERT embedding matrix of $\mathcal{T}$ as follows:
\begin{equation}
{\mathbi{E}_{\mathcal{T}} = [\mathbi{e}_0, \mathbi{e}_1, \mathbi{e}_2,..., \mathbi{e}_n]}.
\end{equation}

Note that $\mathbi{e}_0$, the BERT embedding of the added [CLS] token, is used as the global-level representation of $\mathcal{T}$, for it is designed to incorporate the information of the whole text.

\subsubsection{Implementation of SeqNER}\label{sectionseqner}

\textbf{Setup.} Figure 2 shows the neural architecture of the SeqNER. 
Given an input text $\mathcal{T}=t_0t_1t_2...t_n$ and its label sequence $\mathcal{Y}=y_0y_1y_2...y_n$, the encoder obtains its contextualized embedding matrix $\mathbi{E}_{\mathcal{T}} = [\mathbi{e}_0, \mathbi{e}_1, \mathbi{e}_2,..., \mathbi{e}_n]$. Then the decoder tags each $t_i $ with a token label according to its representation $\mathbi{e}_i$. 
We use the BIO tagging scheme, which uses ``B''  to tag the beginning token of an entity, ``I'' to tag the other tokens of an entity, and ``O'' to tag tokens that do not belong to any entity, as the running example in Figure 2 shows. For the dataset containing $\mathcal{T}$, we assume it defines $m$ entity types. Thus it owns a total of ($2*m+1$) types of token labels, where each entity type has two label types. We use the symbol $\mathbb{L}$ to denote the token label set and the $|\mathbb{L}|=2*m+1$.

\noindent \textbf{Seq Decoder.} We refer to the decoder to tag token labels as Seq Decoder. We propose two methods for the label tagging: softmax-based tagging and CRF-based tagging. 

\underline{\textit{Softmax-based Label Tagging.} }
This method taggs each token label independently.
For each $\mathbi{e}_i$ in $\mathbi{E}_{\mathcal{T}}$, we first reduce its dimention using a Feed Forward Netwrok (FFN):
\begin{equation}
\hat{\mathbi{e}}_i = \mathbi{W}\mathbi{e}_i + \mathbi{b}, 
\end{equation}
where $\hat{\mathbi{e}}_i \in \mathbb{R}^{|\mathbb{L}|}$, $\mathbi{W}$ and $\mathbi{b}$ are trainable FFN parameters. We then pass $\hat{\mathbi{e}}_i$ to the softmax function, yielding probability distributions for the token $t_i$ on the token label set $\mathbb{L}$:
\begin{equation}
{\hat{\pmb{y}}_{i,j} = \frac{\exp (\hat{\mathbi{e}}_{i,j})}{\sum_{k=0}^{|\mathbb{L}|}\exp (\hat{\mathbi{e}}_{i,k})}}.
\end{equation}

The highest response in $\hat{\pmb{y}}_i$ indicates that the corresponding label type is considered activated.

During the model training, we tune model parameters by minimizing the following cross-entropy training loss:
\begin{equation}
{\mathcal{L}_{seqner\textrm{-}s} = -\frac{1}{N_{i}} \sum \limits_{i=1}^{N_{i}}{\pmb{y}}_{{i}}\log\hat{\pmb{y}}_{i}}, 
\end{equation}
where $N_{i}$ denotes the number of token instances. $\pmb{y}_i$ is the one-hot vector of gold label type for the token $t_i$.

\underline{\textit{CRF-based Label Tagging}.}
Instead of predicting each token label independently, the CRF-based labeling tagging considers the correlations between neighbor labels (a.k.a. label dependencies) and jointly decodes the optimal chain of labels by using sentence-level tag information. CRF uses a state transition matrix $\mathbi{A}$ to record the sentence-level tag information.
Label dependencies may play a vital role in sequence labeling. As the running example in Figure 2 shows, it not only makes no sense but also is illegal to tag the ``{night}'' with any other labels except for the ``\texttt{I-Date}''.

For the input text $\mathcal{T}$, CRF describes the probability of generating its whole label sequence as follows:
\begin{equation}
Pr(\hat{\mathcal{Y}}|\mathcal{T}) = \frac{\prod_{i=1}^n\psi (\hat{y}_{i-1},\hat{y}_i,\mathcal{T})}{\sum_{\mathcal{Y}' \in \textrm{Y} }\prod_{i=1}^n\psi ({y}'_{i-1},{y}'_i,\mathcal{T})},
\end{equation}
where $\hat{\mathcal{Y}}=\hat{y}_0\hat{y}_1\hat{y}_2...\hat{y}_n$ is a generic chain of label sequence. $\textrm{Y}$ is the set of all chains of label sequence. $\psi ({y}_{i-1},{y}_i,\mathcal{T})$ calculates the probability of tagging the token $t_i$ with the label $y_i$:
\begin{equation}
\psi ({y}_{i-1},{y}_i,\mathcal{T}) = (\mathbi{W}\mathbi{e}_i+\mathbi b) + \mathbi{A}_{y_{i-1},y_i},
\end{equation}
where we refer to the  ($\mathbi{W}\mathbi{e}_i+\mathbi b$) as the emission score, which is calculated by applying an FFN to the $\mathbi{e}_i$. And we refer to the $\mathbi{A}_{y_{i-1},y_i}$ as the transition score, which is the probability of tagging the token $t_i$ with the label $y_i$ under the condition that the token $t_{i-1}$ has been tagged with the label $y_{i-1}$.

During the model training, we minimize the following negative loglikelihood:
\begin{equation}
\mathcal{L}_{seqner-c} = -\log Pr(\hat{\mathcal{Y}}|\mathcal{T}). 
\end{equation}

The training objective and its gradients can be efficiently computed by dynamic programming. And during the model inference, we use the Viterbi algorithm to find the optimal chain of label sequence that maximizes the following likelihood:
\begin{equation}
\mathcal{Y}^* = \textrm{arg} \max \limits_{\mathcal{Y}' \in \textrm{Y}} Pr({\mathcal{Y}'}|\mathcal{T}).
\end{equation}

\noindent \textbf{Label Combiner.} This module obtains entities by combining consecutive token labels of the same entity type. For example, it obtains the \texttt{Date} entity--``this night'' by combining the two labels of \texttt{Date} type, namely the \texttt{B-Date} and \texttt{I-Date} tagged for the ``this'' and ``night'', respectively.

\subsubsection{Implementation of SpanNER}\label{sectionspanner}
\textbf{Setup.} Figure 3 shows the neural architecture of the SpanNER. Given an input text $\mathcal{T}=t_0t_1t_2...t_n$, the Encoder obtains its contextualized embedding matrix $\mathbi{E}_{\mathcal{T}} = [\mathbi{e}_0, \mathbi{e}_1, \mathbi{e}_2,..., \mathbi{e}_n]$. 
We add an additional \texttt{None} type to tag spans that are not entities. 
For the dataset containing $\mathcal{T}$, we use the symbol $\mathbb{T}$ to denote the set of the $m$ pre-defined entity types and the added \texttt{None} type, where $|\mathbb{T}|=m+1$. 

\noindent \textbf{Span Spliter.}
The Span Splitter first splits the input text into text spans and then obtains their semantic representations. 

Text spans are text segments and can be overlapped. For the text $\mathcal{T}$, we formulate the definition of text span as follows:
\begin{equation}
\mathcal{S} = t_it_{i+1},...,t_{i+j} \quad s.t. \quad 1 \leq i \leq i+j \leq n,
\end{equation}
where the length of $\mathcal{S}$ is ($j+1$). Note that we do not take $t_0$ (i.e., the added [CLS] token) into consideration. Additionally, we also set a length threshold $\epsilon$ to restrict the maximum span length, where $(j+1)\leq \epsilon$. 

We propose three methods to obtain span semantic representations, which we refer to as boundary, span-pooling, and hybrid.

\underline{\textit{Boundary}.} We concatenate the BERT embeddings of span head and tail tokens, and the span length embedding as the span semantic representation:
\begin{equation}
\mathbi{E}_\mathcal{S} = [\mathbi{e}_i; \mathbi{e}_{i+j}; \mathbi{W}_{j+1}],
\end{equation}
where $\mathbi{W}_{j+1}$ is the length embeddings for any span of length $(j+1)$. The span length embeddings are trained during model training.
If a span solely has one token, we duplicate the token embedding and take them as the head and tail token embedding, respectively.

\underline{\textit{Span-pooling}.} We first apply the max-pooling function to the BERT embeddings of all the span tokens. We then concatenate the max-pooling results, the global-level representation of the input text, and the span length embedding as the span semantic representation:
\begin{equation}
\mathbi{E}_\mathcal{S} = [\textrm{max-pooling}([e_i,e_{i+1},...,e_{i+j}]); \mathbi{e}_{0}; \mathbi{W}_{j+1}],
\end{equation}
where $\mathbi{e}_{0}$ is taken as the global-level representation of the text $\mathcal{T}$, as discussed in Section \ref{section3.1.1}.

\underline{\textit{Hybrid}.} We combine the {boundary} and {span-pooling} as the hybrid method, and we obtain the span semantic representation as follows:
\begin{equation}
\mathbi{E}_\mathcal{S} = [\mathbi{e}_i; \mathbi{e}_{i+j};\textrm{max-pooling}([e_i,e_{i+1},...,e_{i+j}]); \mathbi{e}_{0};\mathbi{W}_{j+1}].
\end{equation}

\noindent \textbf{Span Decoder.} We use a softmax-based decoder to conduct span classifications on span semantic representations. Specifically, we first pass $\mathbi{E}_\mathcal{S}$ through an FFN:
\begin{equation}
\hat{\mathbi{E}}_\mathcal{S} = \mathbi{W}\mathbi{E}_\mathcal{S} + \mathbi{b},
\end{equation}
where $\hat{\mathbi{E}}_\mathcal{S} \in \mathbb{R}^{|\mathbb{T}|}$, $\mathbi{W}$ and $\mathbi{b}$ are trainable FFN parameters. Then we feed $\hat{\mathbi{E}}_\mathcal{S}$ to the softmax function:
\begin{equation}
{\hat{\pmb{y}}_{\mathcal{S},j} = \frac{\exp (\hat{\mathbi{E}}_{\mathcal{S},j})}{\sum_{k=0}^{|\mathbb{T}|}\exp (\hat{\mathbi{E}}_{\mathcal{S},k})}},
\end{equation}
where $\hat{\pmb{y}}_\mathcal{S}$ is the predicted probabilities regarding all the types in $\mathbb{T}$. The highest response in $\hat{\pmb{y}}_\mathcal{S}$ indicates that the corresponding type is considered activated.

During the model training, we tune model parameters by minimizing the following cross-entropy loss:
\begin{equation}
{\mathcal{L}_{spanner} = -\frac{1}{N_{k}} \sum \limits_{k=1}^{N_{k}}{\pmb{y}_k}\log\hat{\pmb{y}}_k}, 
\end{equation}
\noindent where ${\pmb{y}}_{k}$ is the one-hot vector of gold span type. ${N_{k}}$ is the number of all span instances.

During the model inference, we remain spans that are predicted as entities.

\subsubsection{Implementation of BL-NER}\label{sectionblner}
Guided by the BL paradigm, we obtain BL-NER by bundling SeqNER and SpanNER together, as shown in Figure 4. 

\begin{figure}[h]
\centering
\includegraphics[width=0.486\textwidth]{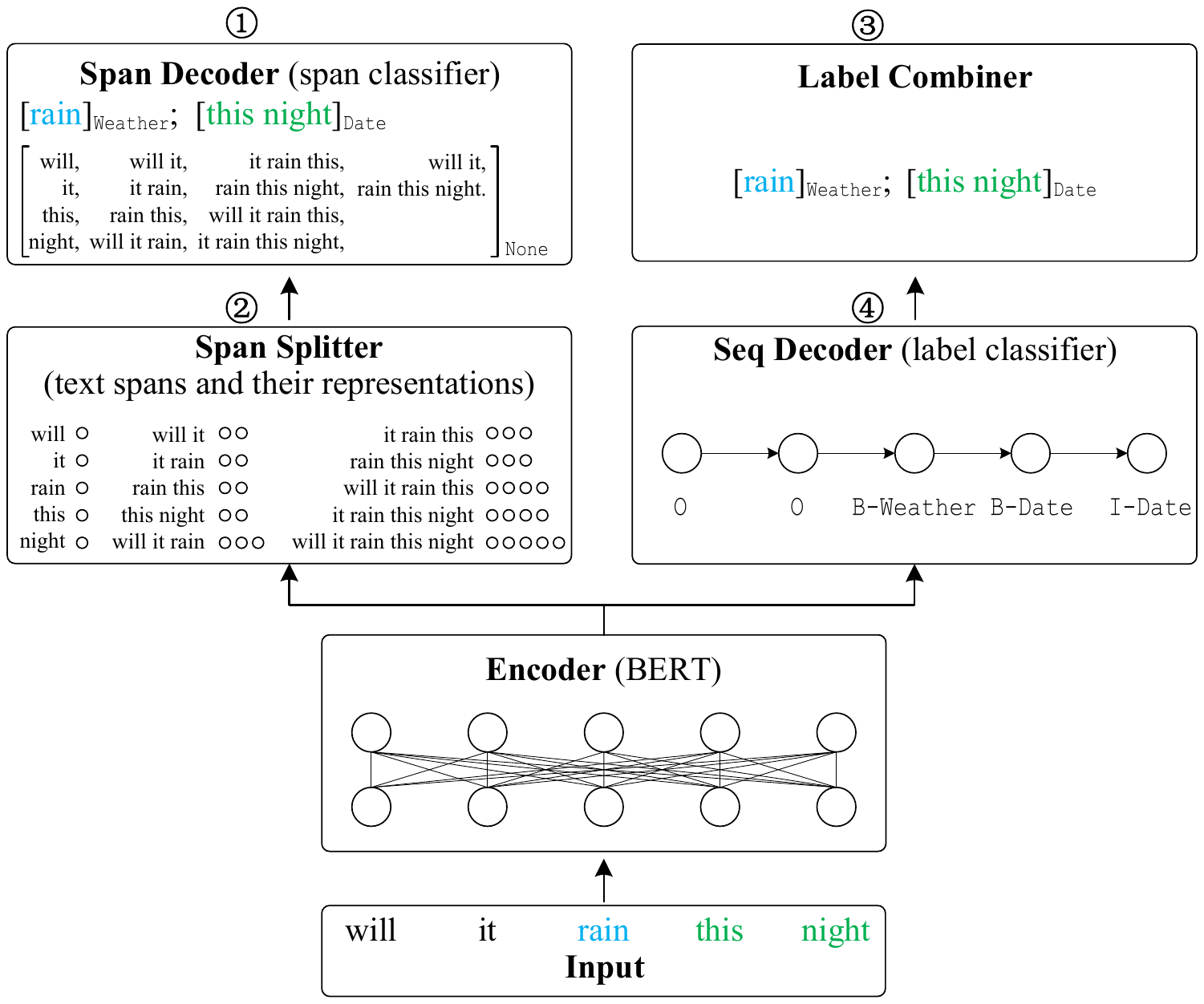} 
\caption{The architecture of BL-NER, which is achieved by bundling SeqNER and SpanNER together. The two bundled models share a common model encoder, and their losses are weighted summed to jointly tune the model parameters.}\label{bl-ner}
\end{figure}

Given an input text $\mathcal{T}=t_0t_1t_2...t_n$, the Encoder obtains its contextualized embedding matrix $\mathbi{E}_{\mathcal{T}} = [\mathbi{e}_0, \mathbi{e}_1, \mathbi{e}_2,..., \mathbi{e}_n]$.
Then $\mathbi{E}_{\mathcal{T}}$ is shared by the Span Splitter (\ding{193}) and the Seq Decoder (\ding{195}). Based on $\mathbi{E}_{\mathcal{T}}$ and $\mathcal{T}$, the Span Splitter and the Span Decoder (\ding{192}) conduct the same operations as they do in SpanNER (Section \ref{sectionspanner}). Similarly, the Seq Decoder (\ding{195}) and the Label Combiner (\ding{194}) do the same thing as they do in SeqNER (Section \ref{sectionseqner}).

During the model training, we weighted sum the traing losses of SeqNER and SpanNER to jointly tune the shared common encoder:
\begin{equation}
\mathcal{L}_{bl\textrm{-}ner} = \alpha * \mathcal{L}_{seqner} + (1-\alpha) * \mathcal{L}_{spanner},\label{equloss}
\end{equation}
where $\alpha$ is a loss weight.

Note that both the Span Decoder (\ding{192}) and the Label Combiner (\ding{194}) can output entities. During the model inference, we use only one of them to output entities with the goal of avoiding model conflicts. And if we use the Span Decoder, we refer to BL-NER as \textbf{BL-SpanNER}, which means that we use BL to enhance SpanNER by bundling SeqNER to it. Similarly, if we use the Label Combiner, we refer to BL-NER as \textbf{BL-SeqNER}, indicating that we use BL to enhance SeqNER by bundling SpanNER to it.

\subsection{Model Discussion}

\subsubsection{Why Bundle the Two NER models?}
The sequence labeling-based and span-based paradigms are regarded as two distinct technical routines for the NER tasks \cite{ji_kbs,spanner,ma2022decomposed}, but they exhibit similar performance across a number of datasets, such as SciERC \cite{coreference}, CoNLL2003 \cite{conll03}, and OntoNotes \cite{ontonotes}. Moreover, we demonstrate that the two paradigms depend on different granularities of semantic representation, i.e., the sequence labeling-based one is built upon token-level representations while the span-based one is based upon span-level representations. 
It is obvious that these two types of representations focus on different perspectives, namely, the token-level representation pays more attention to itself and the correlation with its neighbors, while the span-level representation prefers to consider a sequence of representations as a whole and is more concerned with the correlation among them.
The above difference results in quite different NER results on a number of attributes \cite{spanner}. For example, the sequence labeling-based paradigm performs better when entities are long, while the span-based paradigm is better at dealing with sentences with more OOV words.

The motivation of the BL paradigm is that a NER model will make use of the complementary advantages of the two NER paradigms if it can leverage both token-level and span-level representations simultaneously. 
When the two paradigms share a common encoder, and their losses are used to jointly tune the encoder's parameters, contextualized embeddings will be generated that incorporate token-level as well as span-level representations.

\subsubsection{Why Set the Span Length Threshold?}

A number of span-based NER models \cite{lifei,hamner,spanner} enumerate all text spans during the model training. However, we demonstrate that using all spans is not only unnecessary but also wasteful. For example, if the length ($n$) of the text $\mathcal{T}$ is 100, a total of 5,050 spans will be enumerated, which is calculated by $\frac{n*(n+1)}{2}$. However, only a very few spans are with lengths larger than 10, such as only 0.04\% of the entities in OntoNotes \cite{ontonotes}. Moreover, considering all spans needs vast GPU memory and takes a long time to train the model. By contrast, if we set the span length threshold $\epsilon$ to 10, only 955 spans will be enumerated, which is calculated by $\frac{\epsilon*(2n-\epsilon+1)}{2}$. 

However, the majority of the 955 spans (often more than 900) are not entities, which we refer to as negative spans and assign ``\texttt{None}'' type to them.
Eberts and Ulges \cite{spert} and Yu et al. \cite{yujie} have proven that a small number of negative spans is enough for the model training, and they propose a maximum negative sampling strategy to randomly sample a number of negative spans. We also use the proposed strategy. Given the text $\mathcal{T}$, we use the symbol $\mathbb{S}$ to represent the set of all negative spans restricted by $\epsilon$, and the symbol $\mathbb{D}$ to represent the set of sampled negative spans:
\begin{equation}
|\mathbb{D}| = \min \{|\mathbb{S}|, 100\},
\end{equation}
where $|\cdot|$ represents the number of spans in the set.

During the model inference, we enumerate all spans restricted by $\epsilon$ and predict their types.

\subsubsection{Heuristic Decoding}
We only take the flat NER task into consideration, which assumes no overlapped entities. We find that SpanNER and BL-SpanNER may predict overlapped entities. 
For those overlapped spans, we use the heuristic decoding method proposed by Fu et al. \cite{spanner} to remove them, where we keep the span with the highest prediction probability and drop the others.

\section{Experiments}

\subsection{Experimental Setup}

\subsubsection{Datasets}\label{sectiondatasets}
To provide empirical evidence for the effectiveness of the proposed BL paradigm, we conduct our experiments on eleven English datasets across the following five domains:

\noindent \textbf{General Domain.}
\begin{compactitem}
\item OntoNotes 5.0 (OntoNotes) \cite{ontonotes} dataset is a large corpus comprising various genres of text (news, broadcast, weblogs, usenet newsgroups, conversational telephone speech, talk shows). 
\end{compactitem}

\noindent \textbf{News Domain.}
\begin{compactitem}
\item CoNLL2003 (CoNLL03) \cite{conll03} dataset is one of the widely used NER datasets. Its annotation guideline is based on MUC Conventions \cite{conll03}. This dataset has been split into training, development, and test sets, with 14041, 3250, and 3453 sentences, respectively.
\item CoNLL++ \cite{conll++} is a re-annotated version of the CoNLL03 dataset. Wang et al. \cite{conll++} fix annotation errors of the test set and improve the quality of the training data through their CrossWeigh approach.
\item CoNLL2004 (CoNLL04) \cite{conll04} dataset consists of 1,441 sentences from news articles annotated with four entity types (Location, Organization, People, and Other).
\item ACE2004 (ACE04) \cite{ace04} and ACE2005 (ACE05) \cite{ace05} English dataset composes of news articles in multi-domain such as broadcast, newswire, and weblog. Each of them contains seven entity categories. In addition, we remove the overlapped entities from the two datasets and solely conduct experiments on those entities not overlapped.
\end{compactitem}

\noindent \textbf{Medical Domain.}
\begin{compactitem}
\item BC5CDR \cite{bc5cdr} dataset is proposed for the BioCreative V Chemical Disease Relation task and is composed of 1500 PubMed articles. It contains chemical, disease, and chemical-disease interaction entities.
\item NCBI \cite{ncbi} is a disease corpus of the National Center for Biotechnology Information (NCBI). The dataset is a collection of 793 PubMed abstracts annotated at the mention and concept level. 
\end{compactitem}

\noindent \textbf{Scientific Domain.}
\begin{compactitem}
\item SciERC \cite{coreference} dataset is collected from 500 AI paper abstracts and defines scientific terms and relations specially for scientific knowledge graph construction.
\end{compactitem}

\noindent \textbf{Social Domain.}
\begin{compactitem}
\item WNUT2016 (W16) \cite{w16} is a dataset focusing on NER over Twitter. Its training and development data are taken from prior work on Twitter NER \cite{ritter2015weakly, baldwin2015shared}, and its test data is new and is collected from general Twitter data and domain-specific Twitter data.
 
\item WNUT2017 (W17) \cite{w17} focuses on identifying unusual, previously unseen entities. It covers data from multiple social media platforms. Specifically, its training set uses tweets, its development set is based on YouTube comments, and its testing set combines content from Reddit and
StackExchange. The cross domain nature of the dataset establishes an additional challenge to the task.
\end{compactitem}

For ACE04 and ACE05, we regard an entity mention as correct if its label and the head region of its span are identical to the ground truth. For the other datasets, we regard an entity mention as correct if its label and span match the ground truth.
We report more dataset details in Table \ref{dataset_detail}.

\renewcommand\tabcolsep{11.5pt}
\begin{table}[h]
\caption{Dataset details, where the ``\# Train'', ``\# Dev'', and ``\#Test'' columns record the number of entities in the training, development, and test sets, respectively; the ``\# Type'' column records the number of entity types.}\label{dataset_detail}
\centering
\begin{tabular}{lrrrc}
\toprule
\textbf{Dataset} & \textbf{\# Train} & \textbf{\# Dev} & \textbf{\# Test} & \textbf{\# Type}\\ \midrule
OntoNotes        & 81,828          & 11,066        & 11,257     & 18  \\
CoNLL03          & 23,499         & 5,942         & 5,648      &\ \ 4   \\
CoNLL++          & 23,499         & 5,942         & 5,702     &\ \  4    \\
CoNLL04          & 3,377           & 893          & 1,079     &\ \   4   \\
ACE04            & 12,055          & 1,422         & 1,618     &\ \   7   \\
ACE05            & 15,052          & 2,088         & 1,875     &\ \   7   \\
BC5CDR           & 9,385           & 9,353         & 9,809      &\ \  2   \\
NCBI             & 5,429           & 923          & 941    &\ \   1     \\
SciERC           & 5,598           & 811          & 1,685    &\ \    6   \\
W16              & 2,104           & 661          & 3,473     &  10   \\
W17              & 1,975           & 835          & 1,079     &\ \  6    \\ \bottomrule
\end{tabular}
\end{table}

\noindent \textbf{Discussion.} We would like to emphasize a \textbf{specific characteristic} of W16 and W17: the two datasets focus on detecting and classifying novel and emerging named entities in noisy text. Thus their test sets include data from new domains or social platforms, such as new cybersecurity and mass shootings domains for W16, and new Reddit and StackExchange platforms for W17.
However, no new domains or platforms data is specifically included in the training or development sets. 

There is no doubt that the new data contained in the tests leads to a larger OOV density \cite{fu2020interpretable}, which is defined as follows:
\begin{equation}
\phi_\texttt{oDen} = \frac {\textrm{oov}(\mathcal{X})} {|\mathcal{X}|},
\end{equation}
where $\mathcal{X}$ is the sentences in the test set. $\textrm{oov}(\mathcal{X})$ is the number of words in $\mathcal{X}$ but not in the sentences of the training set. $|\mathcal{X}|$ is the number of all words in $\mathcal{X}$.

Fu et al. \cite{spanner} have proven that span-based NER models do better in sentences with more OOV words, which is also confirmed by our experiment, where SpanNER significantly outperforms SeqNER on the two datasets (see Section \ref{sectionmainresults}).

\subsubsection{Implementation Details}\label{details}
All the experiments are conducted on a single NVIDIA RTX 3090 GPU. We optimize all models using the AdamW for 20 epochs with a learning rate of 5$e^{-5}$, a drop out of 0.1,  a linear scheduler with a warm-up ratio of 0.1, and a weight decay of 1$e^{-2}$. 
Moreover, we report three dataset-specific parameters in Table \ref{parameter}.  Specifically,  we use the \texttt{bert-large-cased} \cite{scibert} as the BERT-large, the \texttt{biobert-pubmed-pmc} \cite{biobert} as the BioBERT, and the \texttt{scibert\_scivocab\_cased} \cite{bert} as the SciBERT.

In addition, we set the loss weight $\alpha$ to 0.1 for BL-NER and use the softmax-based label tagging for SeqNER, and we adopt the hybrid method to obtain span semantic representations. We will conduct detailed investigations on the above three parameters in Section \ref{sectionparameter} to reduce content redundancy here.

\renewcommand\tabcolsep{15.5pt}
\begin{table}[h]
\caption{Dataset-specific parameters.  
We determine the span length threshold ($\epsilon$) value for each dataset based on the principle that over 99\% of entities in the dataset are less than or equal to the value. 
}\label{parameter}
\centering
\begin{tabular}{lccr}
\toprule
\textbf{Dataset} & \textbf{$\epsilon$} & \textbf{batch size} & \textbf{BERT} \\ \midrule
OntoNotes        & 10        & 12                  & BERT-large    \\
CoNLL2003        &\ \ 5         & 32                  & BERT-large    \\
CoNLL++          &\ \  5         & 32                  & BERT-large    \\
CoNLL04          & 10        &\ \  2                   & BERT-large    \\
ACE04            & 10        &\ \  8                   & BERT-large    \\
ACE05            & 10        &\ \  8                   & BERT-large    \\
BC5CDR           &\ \  7         & 16                  & BioBERT     \\
NCBI             &\ \  7         & 16                  & BioBERT       \\
SciERC           & \ \ 8         & 12                  & SciBERT     \\
W16              &\ \  6         & 32                  & BERT-large    \\
W17              &\ \  6         & 24                  & BERT-large    \\ \bottomrule
\end{tabular}
\end{table}

\subsubsection{Evaluation Metrics}
We use the standard Precision (P), Recall (R), and micro-F1 to evaluate the model performance:
\begin{subequations}
\begin{normalsize}
\label{e2}
\begin{align}
\textrm{{P}} &=  \frac{\textrm{TP}}{\textrm{TP+FP}},\\
\textrm{{R}} &= \frac{\textrm{TP}}{\textrm{{TP}+{FN}}}, \\
\textrm{{F1}} &= \frac{{2*\textrm{P}*\textrm{R}}}{\textrm{{P}+{R}}},
\end{align}
\end{normalsize}
\end{subequations}
where TP, FP, and FN stand for true positive, false positive, and false negative, respectively.
For all results, we report the averaged Precision, Recall, and F1 based on five runs with different seeds.

\subsection{Main Results} \label{sectionmainresults}

\renewcommand\tabcolsep{9.5pt}
\begin{table}[h]
\centering
\caption{Performance comparisons between original NER models and their BL enhanced version. $\red{\uparrow}$ ($\blue{\downarrow}$) represents the BL paradigm increases (decreases) the model performance.}\label{mainresults}
\begin{tabular}{lrccc}
\toprule
\textbf{Dataset}           & \textbf{Model} & \textbf{P} & \textbf{R} & \textbf{F1}    \\ \midrule
\multirow{4.5}{*}{OntoNotes} 
								 & \cellcolor[HTML]{E7E7E7}SeqNER         & \cellcolor[HTML]{E7E7E7}89.56      & \cellcolor[HTML]{E7E7E7}88.44      & \cellcolor[HTML]{E7E7E7}89.00          \\
                           & \cellcolor[HTML]{E7E7E7}BL-SeqNER & \cellcolor[HTML]{E7E7E7}89.31      & \cellcolor[HTML]{E7E7E7}90.14      &\ \ \cellcolor[HTML]{E7E7E7} \textbf{89.72} \red{$\uparrow$} \\ \cmidrule{2-5} 
                           & \cellcolor[HTML]{C0C0C0}SpanNER        & \cellcolor[HTML]{C0C0C0}89.67      & \cellcolor[HTML]{C0C0C0}89.30      & \cellcolor[HTML]{C0C0C0}89.49          \\
                           & \cellcolor[HTML]{C0C0C0}BL-SpanNER & \cellcolor[HTML]{C0C0C0}88.95      & \cellcolor[HTML]{C0C0C0}90.64      &\ \ \cellcolor[HTML]{C0C0C0} \textbf{89.78} \red{$\uparrow$} \\ \midrule
\multirow{4.5}{*}{CoNLL2003} & \cellcolor[HTML]{E7E7E7}SeqNER         & \cellcolor[HTML]{E7E7E7}92.06      & \cellcolor[HTML]{E7E7E7}91.36       & \cellcolor[HTML]{E7E7E7}91.71          \\
                           &\cellcolor[HTML]{E7E7E7} BL-SeqNER &\cellcolor[HTML]{E7E7E7}92.12      &\cellcolor[HTML]{E7E7E7}92.64      &\ \ \cellcolor[HTML]{E7E7E7} \textbf{92.38}  \red{$\uparrow$}\\ \cmidrule{2-5} 
                           & \cellcolor[HTML]{C0C0C0}SpanNER        & \cellcolor[HTML]{C0C0C0}92.70      & \cellcolor[HTML]{C0C0C0}91.13      & \cellcolor[HTML]{C0C0C0}91.91          \\
                           &\cellcolor[HTML]{C0C0C0} BL-SpanNER &\cellcolor[HTML]{C0C0C0}92.52      &\cellcolor[HTML]{C0C0C0}92.50      &\ \ \cellcolor[HTML]{C0C0C0} \textbf{92.51} \red{$\uparrow$} \\ \midrule
\multirow{4.5}{*}{CoNLL++}   & \cellcolor[HTML]{E7E7E7}SeqNER         & \cellcolor[HTML]{E7E7E7}93.41      & \cellcolor[HTML]{E7E7E7}89.92      & \cellcolor[HTML]{E7E7E7}91.63          \\
                           &\cellcolor[HTML]{E7E7E7} BL-SeqNER &\cellcolor[HTML]{E7E7E7}93.60      &\cellcolor[HTML]{E7E7E7}92.81      &\ \ \cellcolor[HTML]{E7E7E7} \textbf{93.20}  \red{$\uparrow$}\\ \cmidrule{2-5} 
                           & \cellcolor[HTML]{C0C0C0}SpanNER        & \cellcolor[HTML]{C0C0C0}93.89      & \cellcolor[HTML]{C0C0C0}91.33       & \cellcolor[HTML]{C0C0C0}92.59          \\
                           &\cellcolor[HTML]{C0C0C0} BL-SpanNER &\cellcolor[HTML]{C0C0C0}93.80      &\cellcolor[HTML]{C0C0C0}92.79      &\ \ \cellcolor[HTML]{C0C0C0} \textbf{93.29}  \red{$\uparrow$}\\ \midrule
\multirow{4.5}{*}{CoNLL04}   & \cellcolor[HTML]{E7E7E7}SeqNER         & \cellcolor[HTML]{E7E7E7}91.45      & \cellcolor[HTML]{E7E7E7}91.20      & \cellcolor[HTML]{E7E7E7}90.62          \\
                           &\cellcolor[HTML]{E7E7E7} BL-SeqNER &\cellcolor[HTML]{E7E7E7}91.69      &\cellcolor[HTML]{E7E7E7}91.01      &\ \ \cellcolor[HTML]{E7E7E7} \textbf{91.35}  \red{$\uparrow$}\\ \cmidrule{2-5} 
                           & \cellcolor[HTML]{C0C0C0}SpanNER        & \cellcolor[HTML]{C0C0C0}91.88      & \cellcolor[HTML]{C0C0C0}89.92      & \cellcolor[HTML]{C0C0C0}90.89          \\
                           &\cellcolor[HTML]{C0C0C0} BL-SpanNER &\cellcolor[HTML]{C0C0C0}92.42      &\cellcolor[HTML]{C0C0C0}90.87     &\ \ \cellcolor[HTML]{C0C0C0} \textbf{91.64} \red{$\uparrow$} \\ \midrule
\multirow{4.5}{*}{ACE04}     & \cellcolor[HTML]{E7E7E7}SeqNER         & \cellcolor[HTML]{E7E7E7}88.03           & \cellcolor[HTML]{E7E7E7}85.64           & \cellcolor[HTML]{E7E7E7}86.82               \\
                           &\cellcolor[HTML]{E7E7E7} BL-SeqNER &\cellcolor[HTML]{E7E7E7}88.45           &\cellcolor[HTML]{E7E7E7}87.38           &\ \ \cellcolor[HTML]{E7E7E7}  \textbf{87.91}   \red{$\uparrow$}            \\ \cmidrule{2-5} 
                           & \cellcolor[HTML]{C0C0C0}SpanNER        &  \cellcolor[HTML]{C0C0C0}87.12          &\cellcolor[HTML]{C0C0C0}88.15            & \cellcolor[HTML]{C0C0C0}87.63               \\
                           &\cellcolor[HTML]{C0C0C0} BL-SpanNER &\cellcolor[HTML]{C0C0C0}89.23          &\cellcolor[HTML]{C0C0C0}87.96           &\ \  \cellcolor[HTML]{C0C0C0}  \textbf{88.59}     \red{$\uparrow$}         \\ \midrule
\multirow{4.5}{*}{ACE05}     & \cellcolor[HTML]{E7E7E7}SeqNER         & \cellcolor[HTML]{E7E7E7}84.45           & \cellcolor[HTML]{E7E7E7}87.94           & \cellcolor[HTML]{E7E7E7}86.16               \\
                           &\cellcolor[HTML]{E7E7E7} BL-SeqNER &\cellcolor[HTML]{E7E7E7}86.98          &\cellcolor[HTML]{E7E7E7}87.06          &\ \ \cellcolor[HTML]{E7E7E7}  \textbf{87.02}       \red{$\uparrow$}        \\ \cmidrule{2-5} 
                           & \cellcolor[HTML]{C0C0C0}SpanNER        &  \cellcolor[HTML]{C0C0C0}88.96          &  \cellcolor[HTML]{C0C0C0}86.94          &   \cellcolor[HTML]{C0C0C0}87.94             \\
                           &\cellcolor[HTML]{C0C0C0} BL-SpanNER &\cellcolor[HTML]{C0C0C0}88.78          & \cellcolor[HTML]{C0C0C0}89.26          &\ \ \cellcolor[HTML]{C0C0C0} \textbf{89.02}        \red{$\uparrow$}        \\ \midrule
\multirow{4.5}{*}{BC5CDR}    & \cellcolor[HTML]{E7E7E7}SeqNER         & \cellcolor[HTML]{E7E7E7}90.47      & \cellcolor[HTML]{E7E7E7}89.12      & \cellcolor[HTML]{E7E7E7}89.79          \\
                           &\cellcolor[HTML]{E7E7E7} BL-SeqNER &\cellcolor[HTML]{E7E7E7}89.78      &\cellcolor[HTML]{E7E7E7}90.91      &\ \ \cellcolor[HTML]{E7E7E7} \textbf{90.34} \red{$\uparrow$} \\ \cmidrule{2-5} 
                           & \cellcolor[HTML]{C0C0C0}SpanNER        & \cellcolor[HTML]{C0C0C0}89.37      & \cellcolor[HTML]{C0C0C0}90.46      & \cellcolor[HTML]{C0C0C0}89.91          \\
                           &\cellcolor[HTML]{C0C0C0} BL-SpanNER &\cellcolor[HTML]{C0C0C0}90.05      &\cellcolor[HTML]{C0C0C0}90.74      &\ \ \cellcolor[HTML]{C0C0C0} \textbf{90.40} \red{$\uparrow$} \\ \midrule
\multirow{4.5}{*}{NCBI}      & \cellcolor[HTML]{E7E7E7}SeqNER         & \cellcolor[HTML]{E7E7E7}88.91      & \cellcolor[HTML]{E7E7E7}89.49     & \cellcolor[HTML]{E7E7E7}89.20          \\
                           &\cellcolor[HTML]{E7E7E7} BL-SeqNER &\cellcolor[HTML]{E7E7E7}88.57      &\cellcolor[HTML]{E7E7E7}91.73      &\ \ \cellcolor[HTML]{E7E7E7} \textbf{90.12} \red{$\uparrow$} \\ \cmidrule{2-5} 
                           & \cellcolor[HTML]{C0C0C0}SpanNER        & \cellcolor[HTML]{C0C0C0}90.30      & \cellcolor[HTML]{C0C0C0}89.72     & \cellcolor[HTML]{C0C0C0}90.01          \\
                           &\cellcolor[HTML]{C0C0C0} BL-SpanNER &\cellcolor[HTML]{C0C0C0}89.96      &\cellcolor[HTML]{C0C0C0}90.99      &\ \ \cellcolor[HTML]{C0C0C0} \textbf{90.49} \red{$\uparrow$} \\ \midrule
\multirow{4.5}{*}{SciERC}    & \cellcolor[HTML]{E7E7E7}SeqNER         & \cellcolor[HTML]{E7E7E7}67.26      & \cellcolor[HTML]{E7E7E7}72.05      & \cellcolor[HTML]{E7E7E7}69.57          \\
                           &\cellcolor[HTML]{E7E7E7} BL-SeqNER &\cellcolor[HTML]{E7E7E7}68.98      &\cellcolor[HTML]{E7E7E7}71.51      &\ \ \cellcolor[HTML]{E7E7E7} \textbf{70.22} \red{$\uparrow$} \\ \cmidrule{2-5} 
                           & \cellcolor[HTML]{C0C0C0}SpanNER        & \cellcolor[HTML]{C0C0C0}70.02      & \cellcolor[HTML]{C0C0C0}70.91     & \cellcolor[HTML]{C0C0C0}70.46          \\
                           &\cellcolor[HTML]{C0C0C0}BL-SpanNER &\cellcolor[HTML]{C0C0C0}70.32      &\cellcolor[HTML]{C0C0C0}71.99      &\ \ \cellcolor[HTML]{C0C0C0} \textbf{71.14}  \red{$\uparrow$}\\ \midrule
\multirow{4.5}{*}{W16}       & \cellcolor[HTML]{E7E7E7}SeqNER         &\cellcolor[HTML]{E7E7E7}60.23            &\cellcolor[HTML]{E7E7E7}55.06            &\cellcolor[HTML]{E7E7E7}57.53                \\
                           &\cellcolor[HTML]{E7E7E7} BL-SeqNER &\cellcolor[HTML]{E7E7E7}59.57 &\cellcolor[HTML]{E7E7E7}57.78           &\ \ \cellcolor[HTML]{E7E7E7}  \textbf{58.66}   \red{$\uparrow$}            \\ \cmidrule{2-5} 
                           & \cellcolor[HTML]{C0C0C0}SpanNER        &\cellcolor[HTML]{C0C0C0}63.26         & \cellcolor[HTML]{C0C0C0}58.23           &  \cellcolor[HTML]{C0C0C0}\textbf{60.64}              \\
                           &\cellcolor[HTML]{C0C0C0} BL-SpanNER &\cellcolor[HTML]{C0C0C0}61.43           &\cellcolor[HTML]{C0C0C0}57.37           &\ \ \cellcolor[HTML]{C0C0C0}  59.86        \blue{$\downarrow$}       \\ \midrule
\multirow{4.5}{*}{W17}       & \cellcolor[HTML]{E7E7E7}SeqNER         & \cellcolor[HTML]{E7E7E7}44.39      & \cellcolor[HTML]{E7E7E7}62.24      & \cellcolor[HTML]{E7E7E7}51.82          \\
                           &\cellcolor[HTML]{E7E7E7} BL-SeqNER &\cellcolor[HTML]{E7E7E7}47.26 &\cellcolor[HTML]{E7E7E7}59.35      &\ \ \cellcolor[HTML]{E7E7E7} \textbf{52.62}  \red{$\uparrow$}\\ \cmidrule{2-5} 
                           & \cellcolor[HTML]{C0C0C0} SpanNER        & \cellcolor[HTML]{C0C0C0}60.32      & \cellcolor[HTML]{C0C0C0}50.19      & \cellcolor[HTML]{C0C0C0}\textbf{54.79} \\
                           &\cellcolor[HTML]{C0C0C0} BL-SpanNER &\cellcolor[HTML]{C0C0C0}56.63 &\cellcolor[HTML]{C0C0C0}51.35      &\ \ \cellcolor[HTML]{C0C0C0} 53.86   \blue{$\downarrow$}        \\ \bottomrule
\end{tabular}
\end{table}

\begin{table}[]
\caption{Performance comparisons with previous strong baselines.}\label{sota}
\centering
\begin{tabular}{llccc}
\toprule
\textbf{Dataset}          & \multicolumn{1}{c}{\textbf{Model}}                                    & \textbf{P}                    & \textbf{R}                    & \textbf{F1}                            \\ \midrule
                          & Multi-turn QA \cite{qa}                              & 89.00                          & 86.60                          & 87.80                                  \\
                          & SpERT \cite{spert}                                   & 88.25                         & 89.64                         & 88.94                                  \\
                          & Table-Sequence \cite{wa_lu} & -     & -     & 90.10          \\
                          & TriMF \cite{trimf}                                   & 90.26                         & 90.34                         & 90.30                                  \\
                          & TablERT \cite{tablert}                               & -                             & -                             & 91.30                                  \\ \cmidrule{2-5} 
\multirow{-6}{*}{CoNLL04} & BL-SpanNER                                    & 92.42 & 90.87 & \textbf{91.64} \\ \midrule
                          & BERT-CRF \cite{bertcrf}                              & -                             & -                             & 86.00                                  \\
                          & RDANER \cite{rdaner}                                 & -                             & -                             & 87.38                                  \\
                          & SparkNLP \cite{sparknlp}                             & -                             & -                             & 89.73                                  \\
                          & ELECTRAMed \cite{electramed}                         & 88.76                         & 91.34                         & 90.03                                  \\
                          & BioLinkBERT \cite{linkbert}                              & -                             & -                             & 90.22                                 
                          
\\ \cmidrule{2-5} 
\multirow{-6}{*}{BC5CDR}  & BL-SpanNER                                                            & 90.05                         & 90.74                         &\textbf{90.40}                         \\ \midrule
                          & ELECTRAMed \cite{electramed}                         & 85.87                         & 89.29                         & 87.54                                  \\
                          & RDANER \cite{rdaner}                                 & -                             & -                             & 87.89                                  \\
                          & BioLinkBERT \cite{linkbert}                          & -                             & -                             & 88.76                                  \\
                          & CL-KL \cite{clkl}                                    & -                             & -                             & 89.24                                  \\
                          & SciFive-Base \cite{scifive}                          & 88.65                         & 90.14                         & 89.39                                  \\
                          & BioBERT \cite{biobert}                               & 88.22                         & 91.25                         & 89.71                                  \\ \cmidrule{2-5} 
\multirow{-7}{*}{NCBI}    & BL-SpanNER                                                            & 89.96                         & 90.99                         & \textbf{90.49}                         \\ \midrule
								& SciBERT \cite{scibert}                                                                      & -                              &                              - &67.57                                        \\
                          & PURE \cite{chendanqi}                                                                      & -                              &                              - &68.90                                        \\
                          & RDANER \cite{rdaner}                                                                      &   -                            &                              - &68.96                         \\
                          &  SpERT \cite{spert}                                                                     &  70.87                             &69.79                               &70.33                                        
                                          
\\ \cmidrule{2-5} 
\multirow{-5.5}{*}{SciERC}  & BL-SpanNER                                                            & 70.32                         & 71.99                         & \textbf{71.14}                         \\ \midrule
								&CambridgeLTL \cite{cambridgeltl}                                                                       &60.77 &46.07  &52.41                                     \\
                          &InferNER \cite{inferner}                                                                       &-                               &-                               &53.48                                        \\
                          & SANER \cite{saner}                                                                      & -                              &-                               &55.01                                        \\
                          &CL-LK \cite{clkl}                                                                       &-                               &                              - &58.98                              \\ 
								&S-NER \cite{yujie}                                                                       &-                               &                              - &60.12                              \\

\cmidrule{2-5} 
\multirow{-6.5}{*}{W16}     & SpanNER                                                            & 63.26                              &58.23                               &\textbf{60.64}                                        \\\bottomrule
\end{tabular}
\end{table}

As mentioned in Section \ref{sectionblner}, we refer to BL-NER as BL-SpanNER when using the Span Decoder (Figure \ref{bl-ner}-\ding{192}) to output entities. Similarly, we refer to BL-NER as BL-SeqNER when using the Label Combiner (Figure \ref{bl-ner}-\ding{194}) to output entities. In this section, we conduct experiments to investigate the effectiveness of the BL paradigm by comparing SeqNER to BL-SeqNER, as well as SpanNER to BL-SpanNER.
We report the comparison results in Table \ref{mainresults}, from which we observe that: 

(1) BL-SeqNER consistently outperforms SeqNER across the eleven datasets. Specifically, BL-SeqNER brings an averaged +0.88\% F1 score and a maximum of +1.57\% F1 score (on CoNLL++) compared to SeqNER. These gains reveal that the BL paradigm always boosts the performance of SeqNER. 

(2) BL-SpanNER beats SpanNER on the first-nine datasets. To be more precise, BL-SpanNER beats SpanNER by an averaged +0.67\% F1 score and a maximum of +0.96\% F1 score (on ACE04) across the nine datasets. BL-SpanNER instead decreases -0.78\% and -0.93\% F1 scores on W16 and W17, respectively. 

(3) SpanNER shows a consistent superiority to SeqNER across the eleven datasets, especially on W16 and W17. Statistically, SpanNER outperforms SeqNER by an averaged +0.70\% F1 score across the first-nine datasets. We attribute these results primarily to the fact that SpanNER uses the effective hybrid method to obtain span representations, which is an overall consideration of the structure, boundary, length, and context of spans.
And the performance gains are +3.11\% and +2.97\% on W16 and W17, respectively. {We demonstrate that the large OOV density of the two datasets has the most significant effect on the results (see the discussion in Section \ref{sectiondatasets})}.

Based on the above observations,
we draw a conclusion regarding \textbf{when the BL paradigm works}:
\begin{compactitem}
\item \textbf{\textsc{Conclusion}} \#1:  The BL paradigm enhances the performance of both SeqNER and SpanNER when the two NER models perform closely on NER datasets. 
However, if there is a large performance gap between the two NER models on NER datasets, BL will boost the model that performs poorly while decreasing the other model.
\end{compactitem}

The above conclusion is also consistent with human intuition.
Moreover, we demonstrate that these performance gains are due to the fact that the BL paradigm allows the NER models to leverage their complementary advantages. We will present detailed analyses of the exhaustive complementary advantages in Section \ref{sectionwhy}.

In addition, BL-SpanNER achieves competing results on CoNLL04, BC5CDR, NCBI, and SciERC. These results also validate the effectiveness of the BL paradigm. 
Moreover, SpanNER creates the current best performance on W16. We report these performance comparisons in Table \ref{sota}.

\subsection{Why Does the BL Paradigm Work?}\label{sectionwhy}
This section examines in detail the reasons for the success of the BL paradigm.
We select the largest dataset of each domain for the investigation, and we actually select both W16 and W17 of the social domain due to their specific characteristic. 
The selected datasets are OntoNotes (general), CoNLL03 (news), BC5CDR (medical), SciERC (scientific), and W16 and W17 (social), and we use their dev sets to report model performance.

\subsubsection{What Complementary Advantages Does BL Leverage?}\label{sectioncomplementary}

The holistic results in Table \ref{mainresults} show the effectiveness of the BL paradigm, but they cannot interpret the complementary advantages of SeqNER and SpanNER that BL can leverage.  Fu et al. \cite{fu2020interpretable} propose an interpretable evaluation idea by breaking the holistic performance into different buckets from a number of perspectives and using performance heatmaps to investigate relative advantages between two systems. In this section, we make attempts to explore the complementary advantages using their evaluation idea.

\noindent \textbf{Setup.} 
Following Fu et al. \cite{fu2020interpretable}, we break the holistic performance into four groups based on different attributes. To be specific, given an entity $e$ that belongs to a text $\mathcal{T}$, we define the following attribute feature functions:
\begin{subequations}
\begin{normalsize}
\begin{align}
\phi_\texttt{eLen} &= \textrm{len}(e),\\
\phi_\texttt{tLen} &= \textrm{len}(\mathcal{T}),\\
\phi_\texttt{eCon} &=\frac{|\{\varepsilon \mid \operatorname{label}(\varepsilon)=\operatorname{label}(e), \forall \varepsilon \in \mathcal{E}\}|}{|\mathcal{E}|},\\
\phi_\texttt{eDen} &= \frac{\textrm{ent}(\mathcal{T})}{\phi_\texttt{tLen}},
\end{align}
\end{normalsize}
\end{subequations}
where $\phi_\texttt{eLen}$ denotes entity length; $\phi_\texttt{tLen}$ denotes text length; $\phi_\texttt{eCon}$ denotes entity label consistency and $\mathcal{E}$ denotes all entities in the training set; $\phi_\texttt{eDen}$ denotes entity density and $\textrm{ent}(\mathcal{T})$ calculates the number of entities in the text $\mathcal{T}$. 

The $\phi_\texttt{eCon}$ attribute measures the consistency that a particular entity is labeled with a unified label. For example, if the ``rain'' in the running example of Figure \ref{figure1} has a higher entity label consistency, it means that the ``rain'' is frequently labeled as the ``\texttt{Weather}'' type in the training set.

Based on the values of the four attributes, we divide test entities into four buckets, namely extra-small (XS), small (S), large (L), and extra-large (XL). We report the detailed attribute intervals for the four attributes in Table \ref{attribute}.

\renewcommand\tabcolsep{9.5pt}
\begin{table}[h]
\caption{The attribute intervals of bucket XS, S, L, and XL for attributes \texttt{eCon}, \texttt{tLen}, \texttt{eCon}, and \texttt{eDen}.
}\label{attribute}
\centering
\begin{tabular}{lllll}
\toprule
\textbf{Bucket}   & \multicolumn{1}{c}{\texttt{eLen}} & \multicolumn{1}{c}{\texttt{tLen}} & \multicolumn{1}{c}{\texttt{eCon}} & \texttt{eDen}           \\ \midrule
XS & {[}1{]}                  & {[}1, 7{]}                & {[}0, 0.1{]}              & {[}0, 0.01{]}   \\
S  & {[}2{]}                  & {[}8, 16{]}               & (0.1, 0.5{]}              & (0.01, 0.025{]} \\
L  & {[}3, 4{]}                & {[}17, 31{]}              & (0.5, 0.9{]}              & (0.025, 0.05{]} \\
XL & {[}5,{ ]}                 & {[}32,{ ]}                & (0.9, 1{]}                & (0.05, 1{]}     \\ \bottomrule
\end{tabular}
\end{table}

We take all four models, i.e., SpanNER, BL-SpanNER, SeqNER, and BL-SeqNER, into consideration.
And we analyze the complementary advantages through three model pairs, i.e., $<$SpanNER, SeqNER$>$, $<$SpanNER, BL-SpanNER$>$, and $<$SeqNER, BL-SeqNER$>$. Specifically, (1) the $<$SpanNER, SeqNER$>$ enables us to find out the relative advantages of the two models; (2) the $<$SpanNER, BL-SpanNER$>$ allows us to conclude what advantages that BL delivers from SeqNER to SpanNER; (3) similarly, we can be aware of the advantages that BL transfers from SpanNER to SeqNER through the $<$SeqNER, BL-SeqNER$>$.
 For all four models, we report their bucket-wise F1 scores regarding the four defined attributes.

\renewcommand\tabcolsep{1.5pt}
\begin{table*}[h]
  \centering 
\caption{Performance heatmaps regarding complementary advantages. For clarity, SpanNER is taken as an standard, where each value in its heatmap entry is 0. For the other models, each value in heatmap entry $(i,j)$ represents the performance gap (measured by F1 score) between the model and {SpanNER} on $j$-th bucket, and \textbf{the green (red) area indicates that the model performs better (worse) than SpanNER.} \texttt{eLen}, \texttt{tLen}, \texttt{eCon}, and \texttt{eDen} represent different attributes. }
  \label{complementaryheatmap}
    \begin{tabular}{rrrrrrrrrrr rrrrrrrrrrr}
    \toprule
    \multicolumn{11}{c}{\textbf{OntoNotes}} & \multicolumn{11}{c}{\textbf{CoNLL03}}\\
    \cmidrule(lr){1-11}\cmidrule(lr){12-22}
    \multicolumn{4}{c}{\texttt{eLen}}      & \multicolumn{2}{c}{\texttt{tLen}} & \multicolumn{2}{c}{\texttt{eCon}} & \multicolumn{3}{c}{\texttt{eDen}} 
    & \multicolumn{4}{c}{\texttt{eLen}}      & \multicolumn{2}{c}{\texttt{tLen}} & \multicolumn{2}{c}{\texttt{eCon}} & \multicolumn{3}{c}{\texttt{eDen}}\\
    \cmidrule(lr){1-11}\cmidrule(lr){12-22}
           \multicolumn{4}{c}{\includegraphics[scale=0.182]{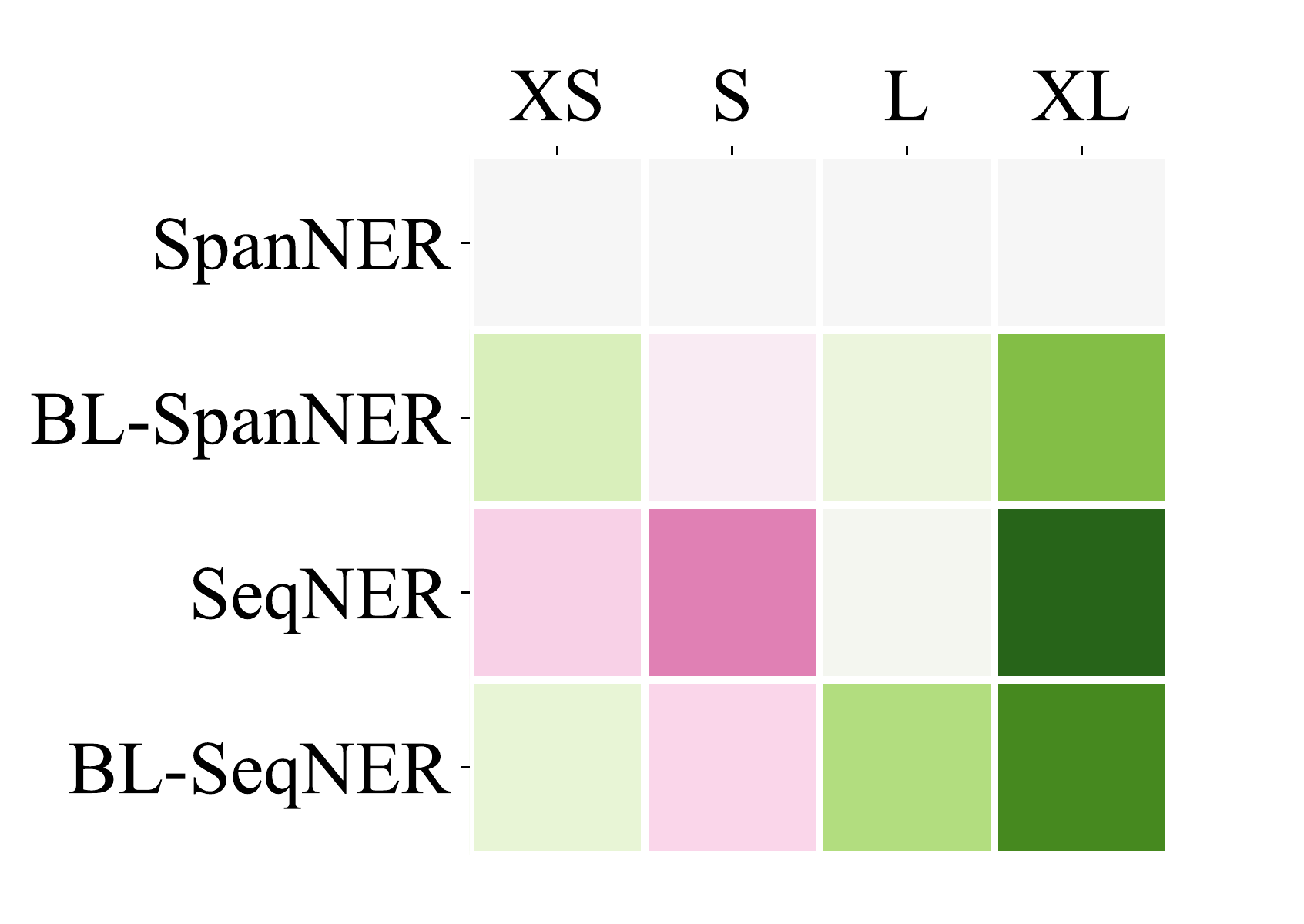}}  
        & \multicolumn{2}{c}{\includegraphics[scale=0.182]{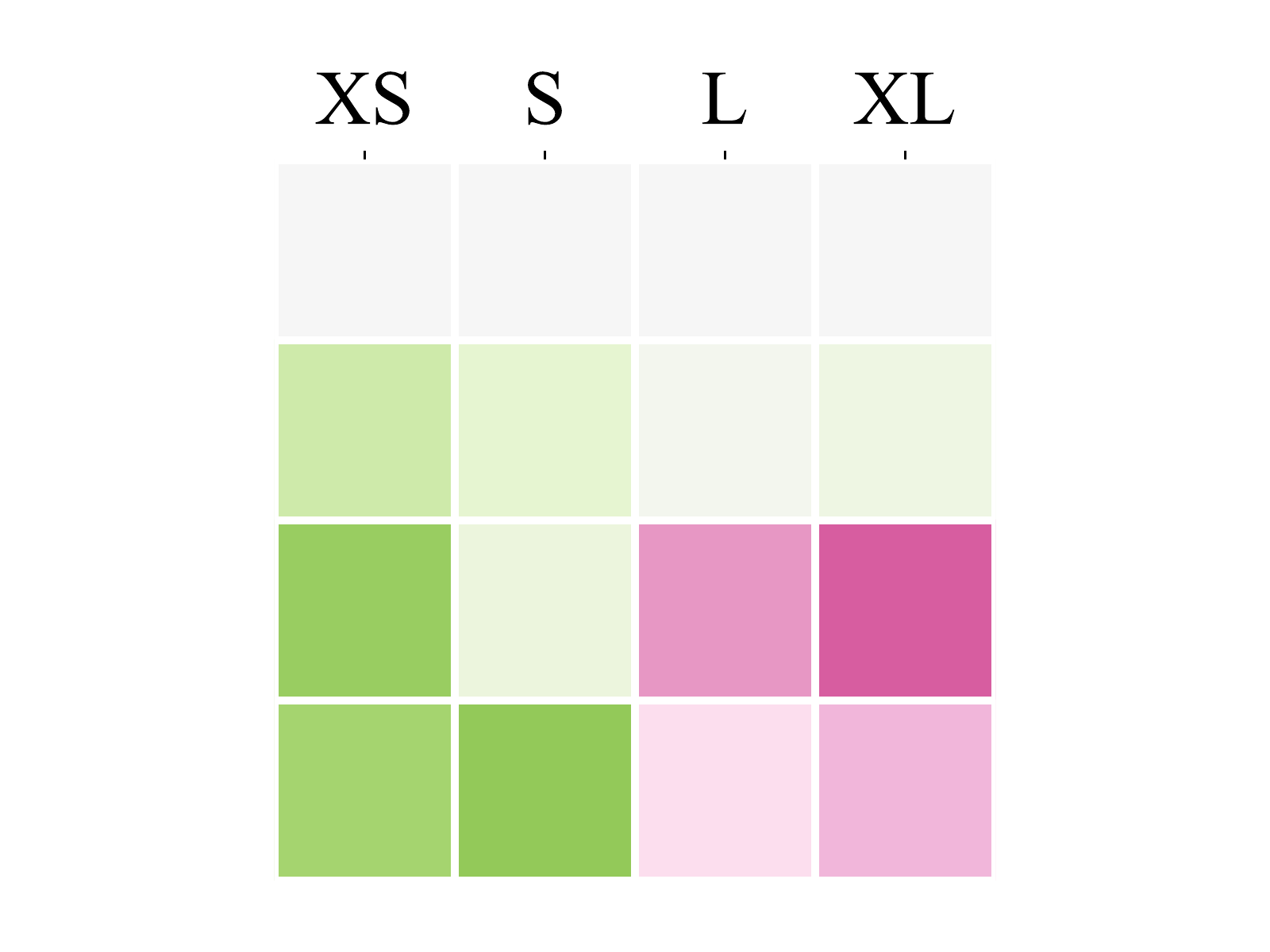}}                 
        & \multicolumn{2}{c}{\includegraphics[scale=0.182]{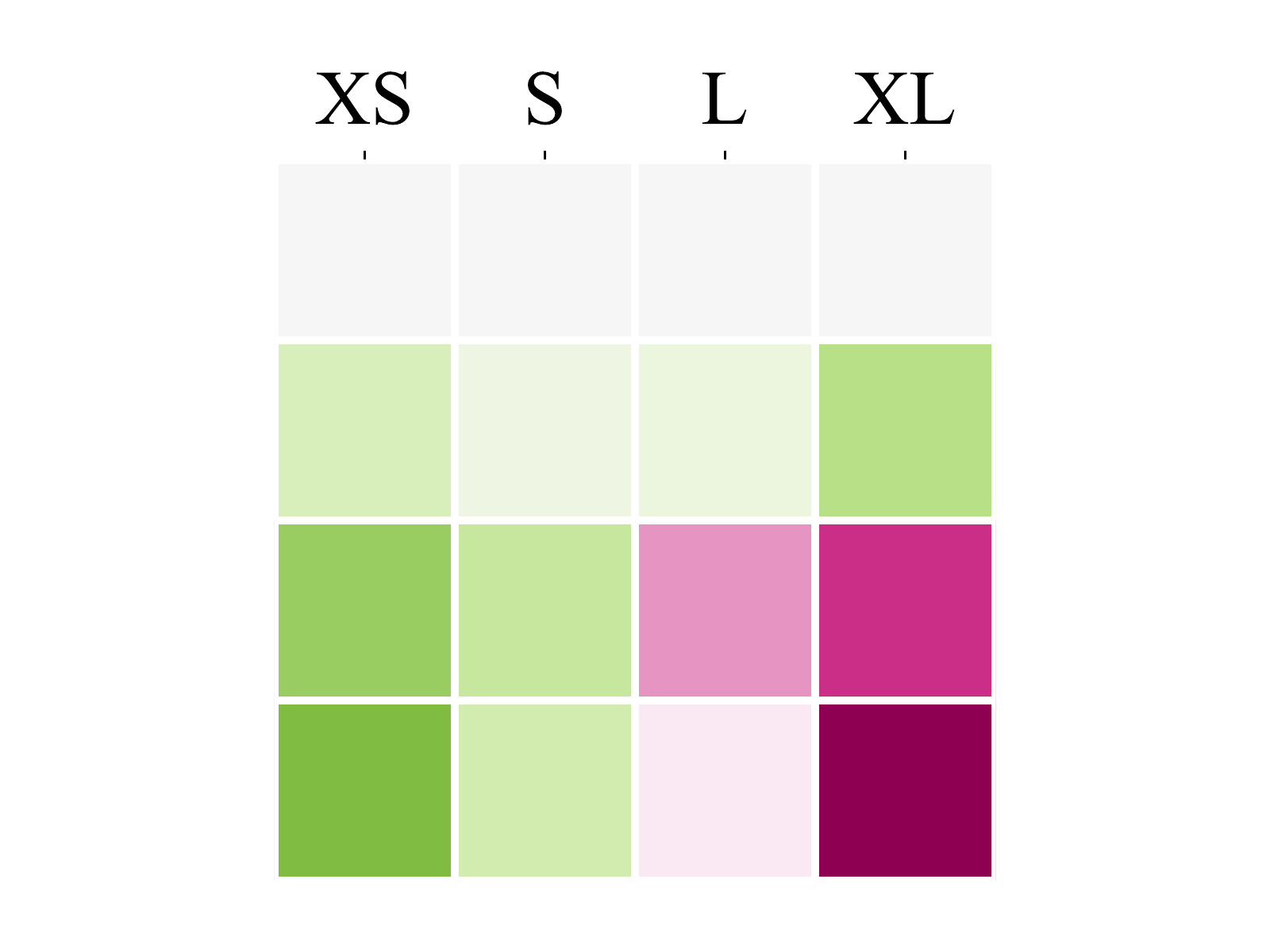}} 
        & \multicolumn{3}{c}{\includegraphics[scale=0.182]{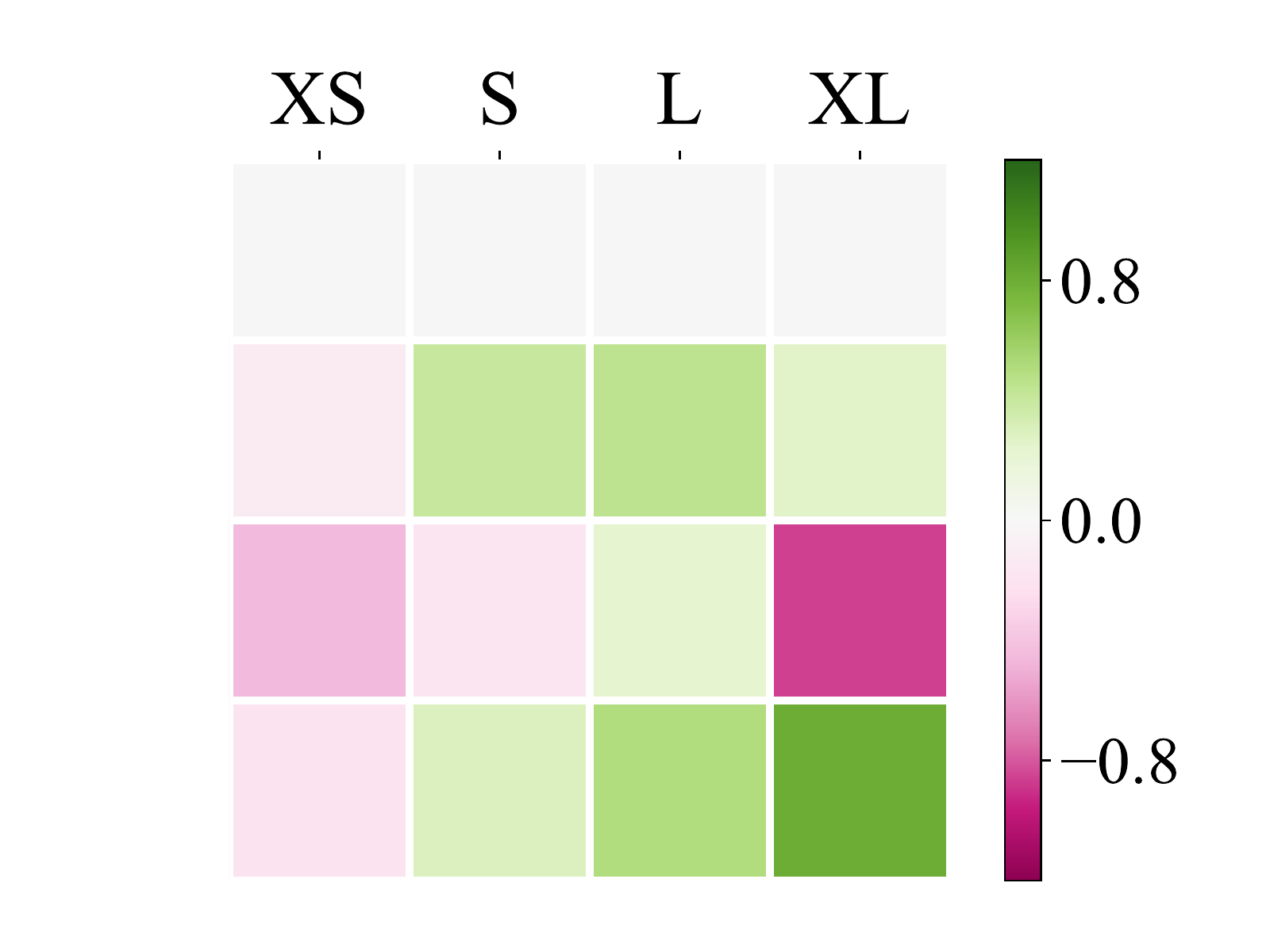}} 
     	  &\multicolumn{4}{c}{\includegraphics[scale=0.182]{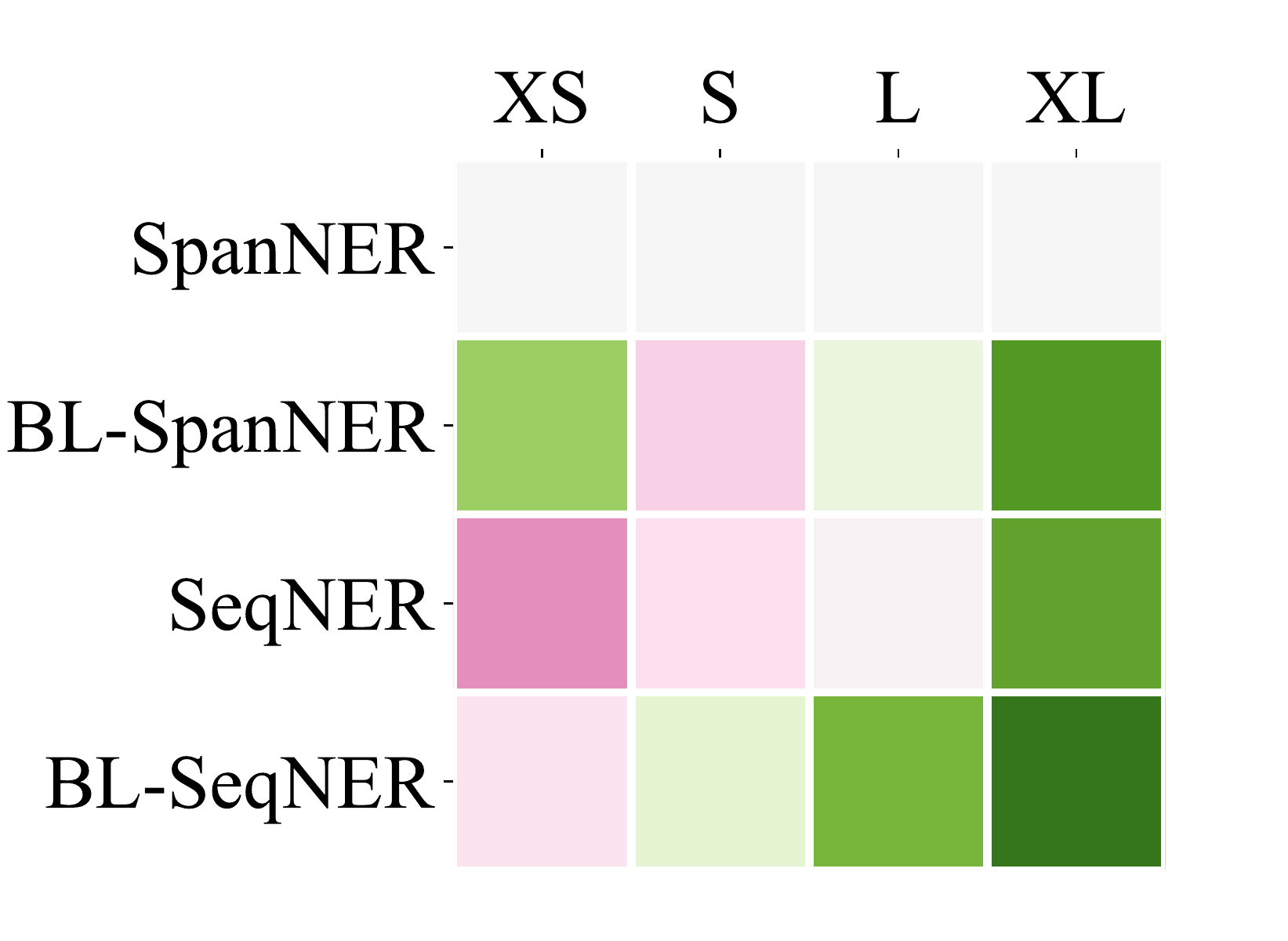}}  
        & \multicolumn{2}{c}{\includegraphics[scale=0.182]{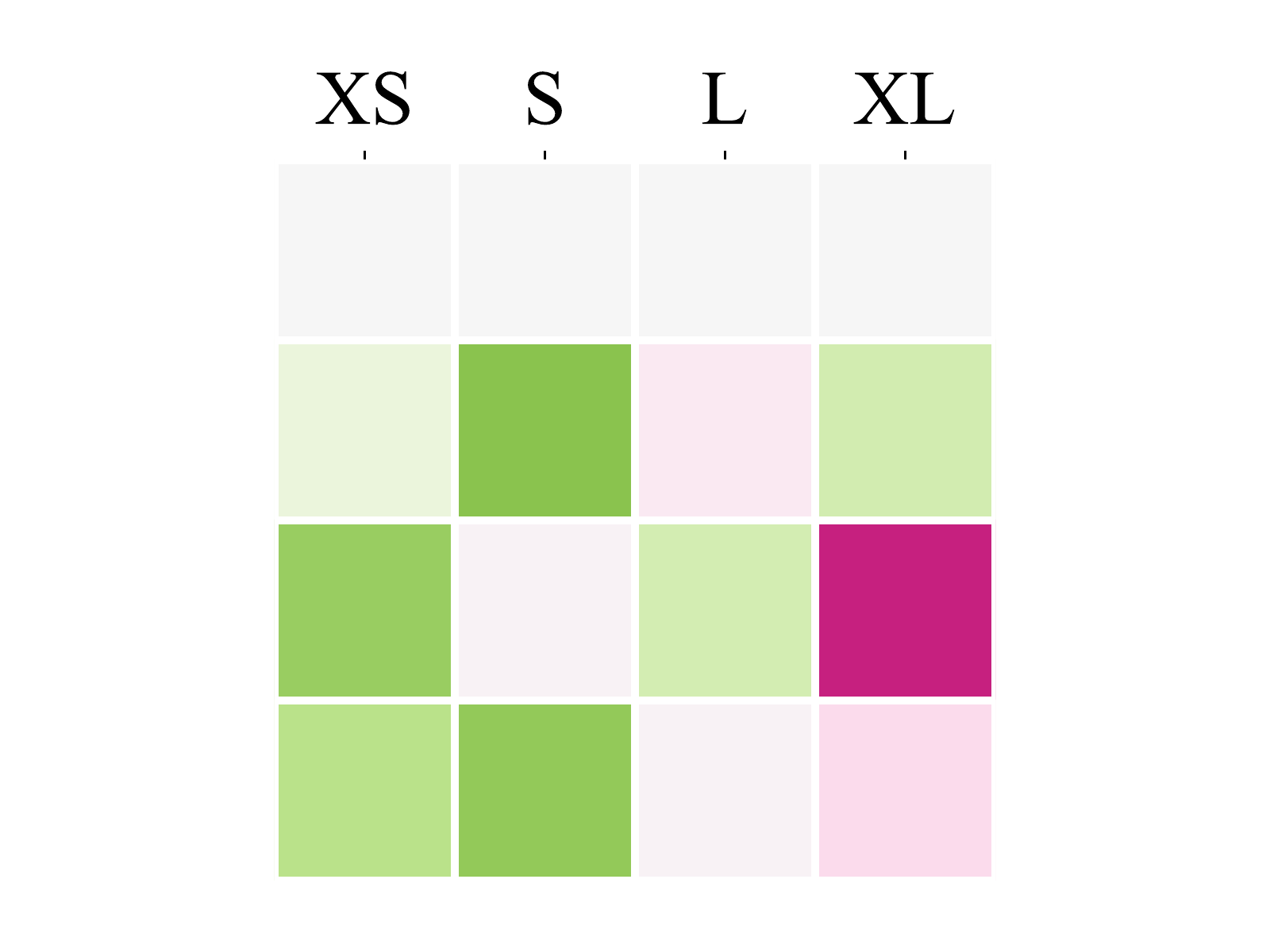}}                 
        & \multicolumn{2}{c}{\includegraphics[scale=0.182]{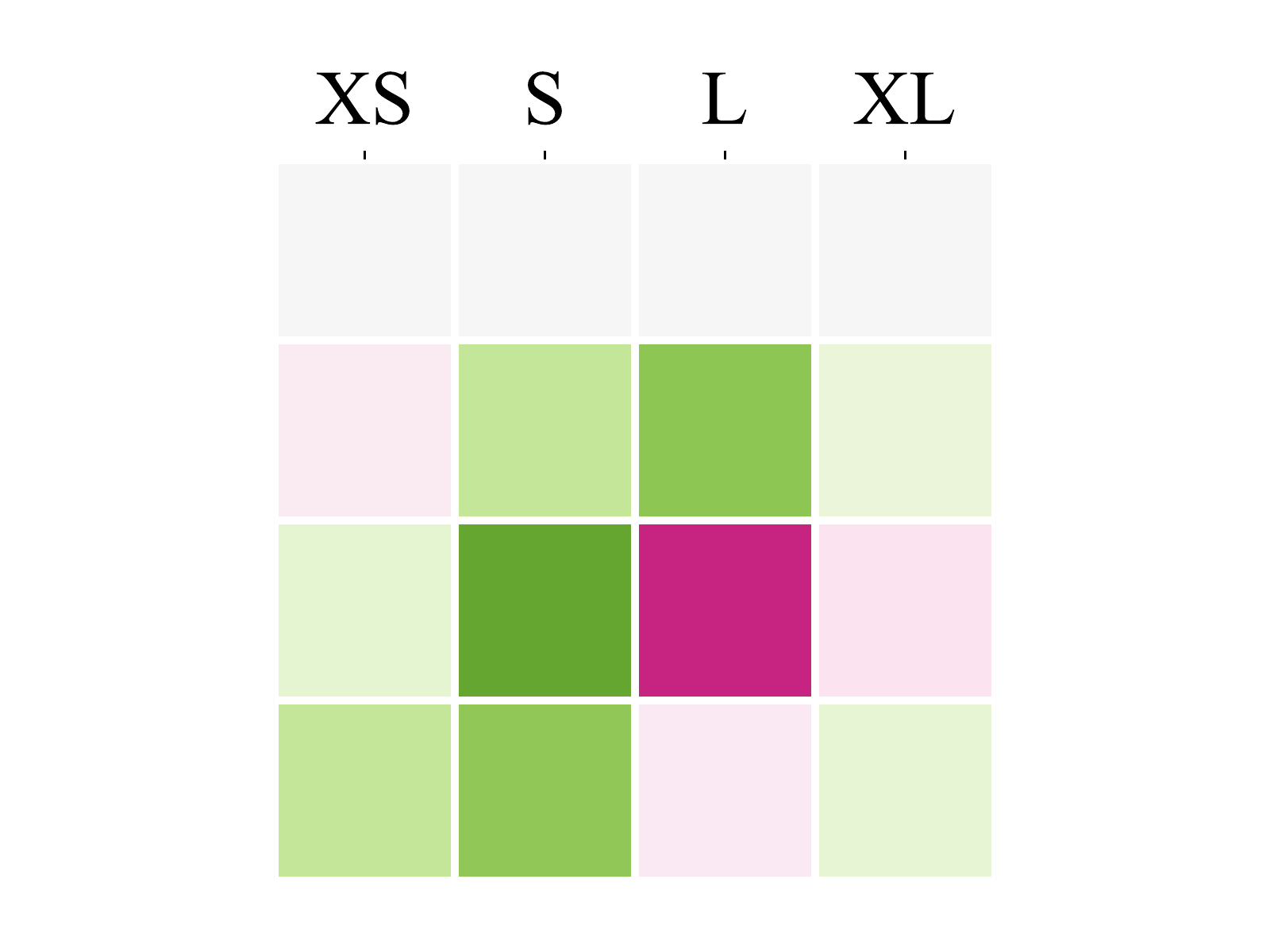}} 
        & \multicolumn{3}{c}{\includegraphics[scale=0.182]{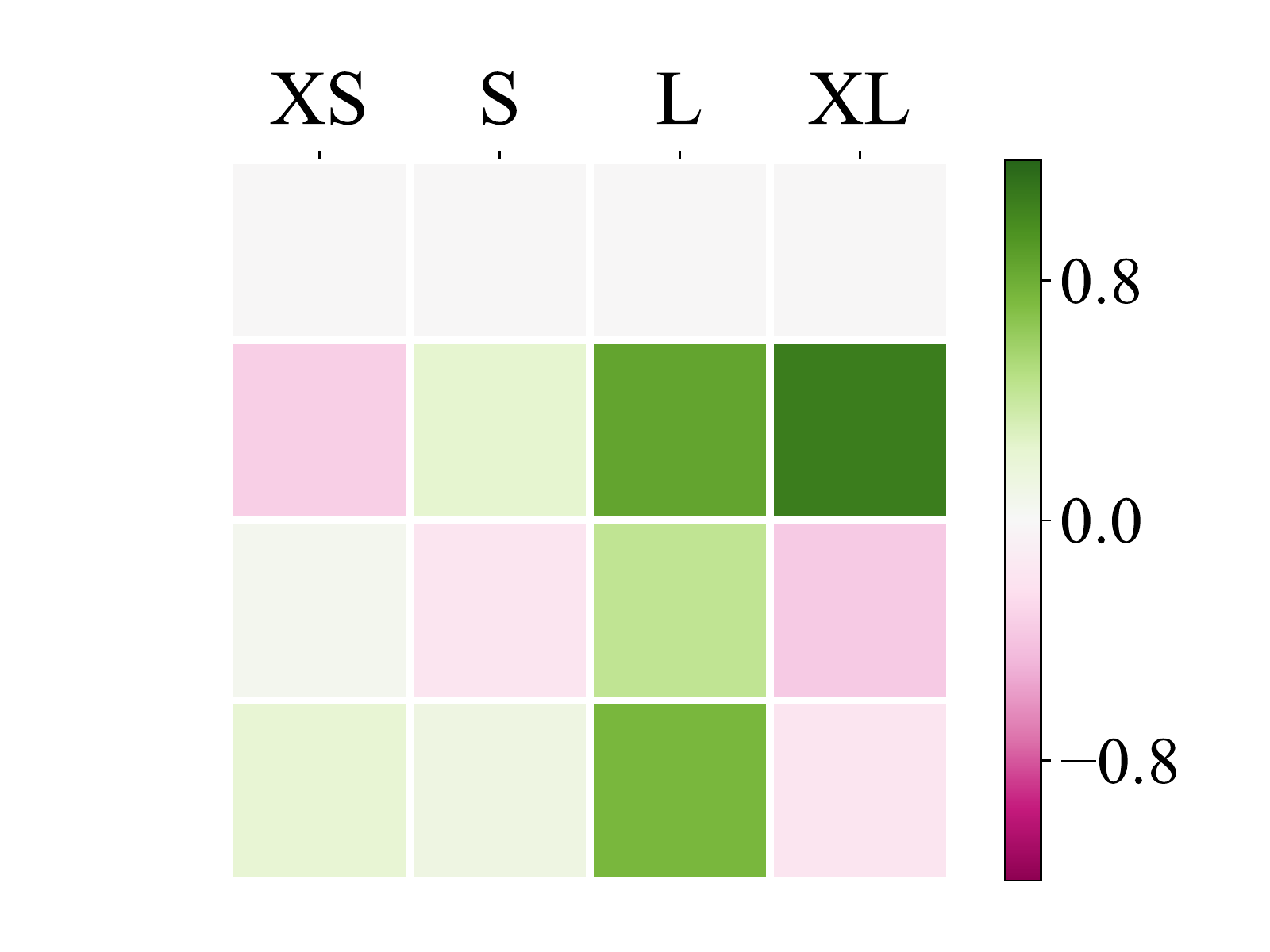}} 
        \\
        \midrule \midrule
	    \multicolumn{11}{c}{\textbf{BC5CDR}} & \multicolumn{11}{c}{\textbf{SciERC}}\\
    \cmidrule(lr){1-11}\cmidrule(lr){12-22}
    \multicolumn{4}{c}{\texttt{eLen}}      & \multicolumn{2}{c}{\texttt{tLen}} & \multicolumn{2}{c}{\texttt{eCon}} & \multicolumn{3}{c}{\texttt{eDen}} 
    & \multicolumn{4}{c}{\texttt{eLen}}      & \multicolumn{2}{c}{\texttt{tLen}} & \multicolumn{2}{c}{\texttt{eCon}} & \multicolumn{3}{c}{\texttt{eDen}}\\
    \cmidrule(lr){1-11}\cmidrule(lr){12-22}
           \multicolumn{4}{c}{\includegraphics[scale=0.182]{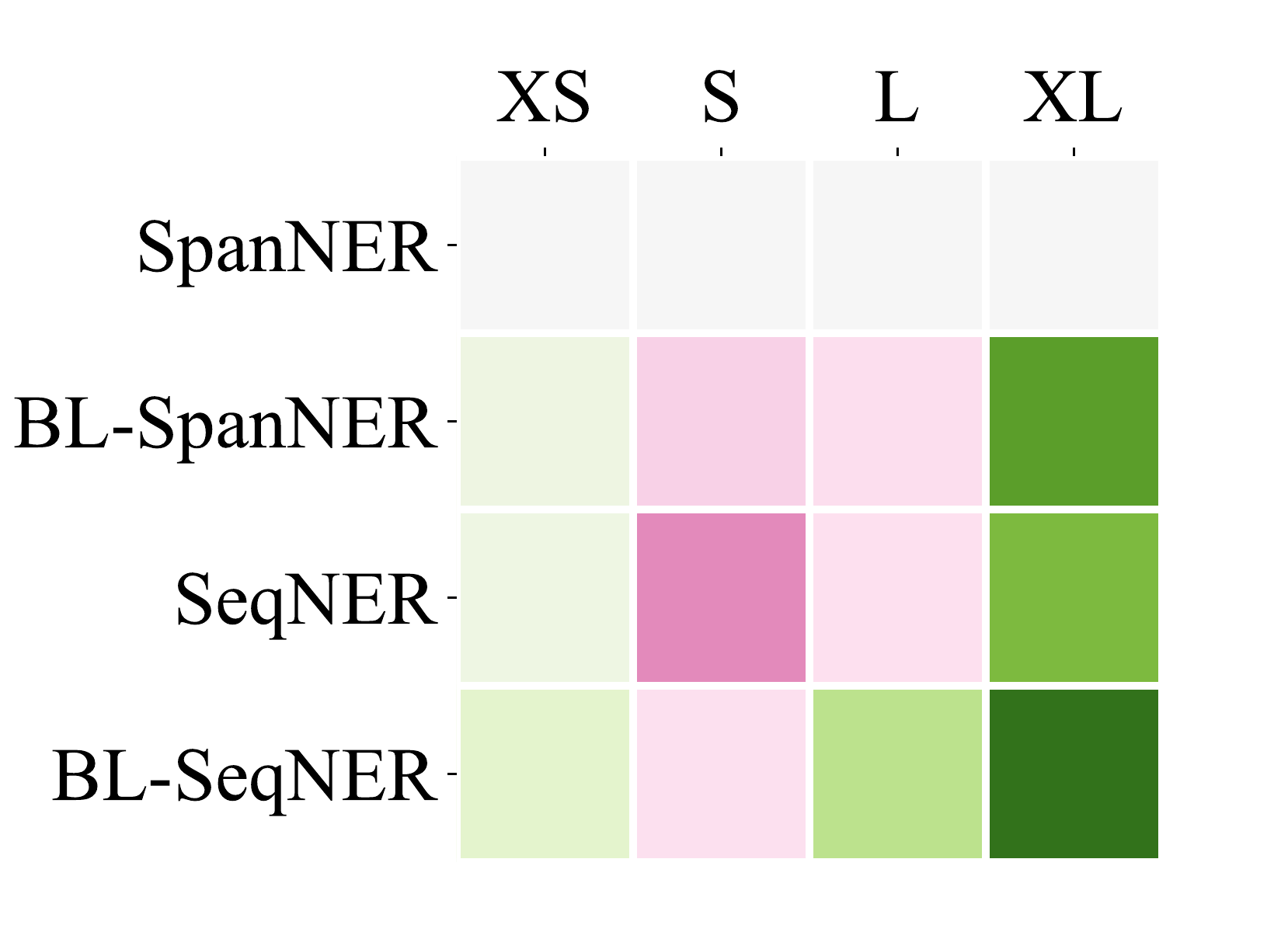}}  
        & \multicolumn{2}{c}{\includegraphics[scale=0.182]{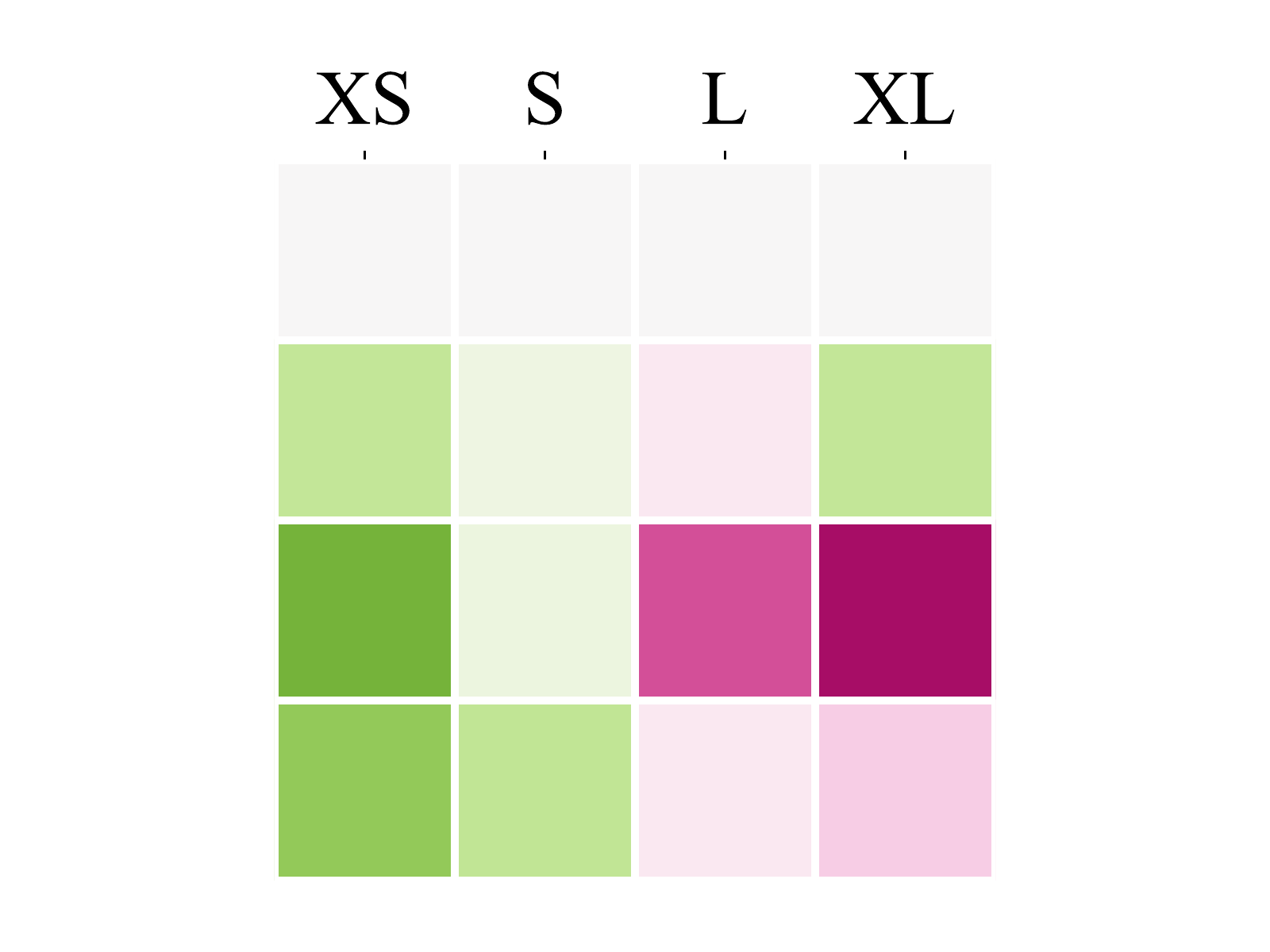}}                 
        & \multicolumn{2}{c}{\includegraphics[scale=0.182]{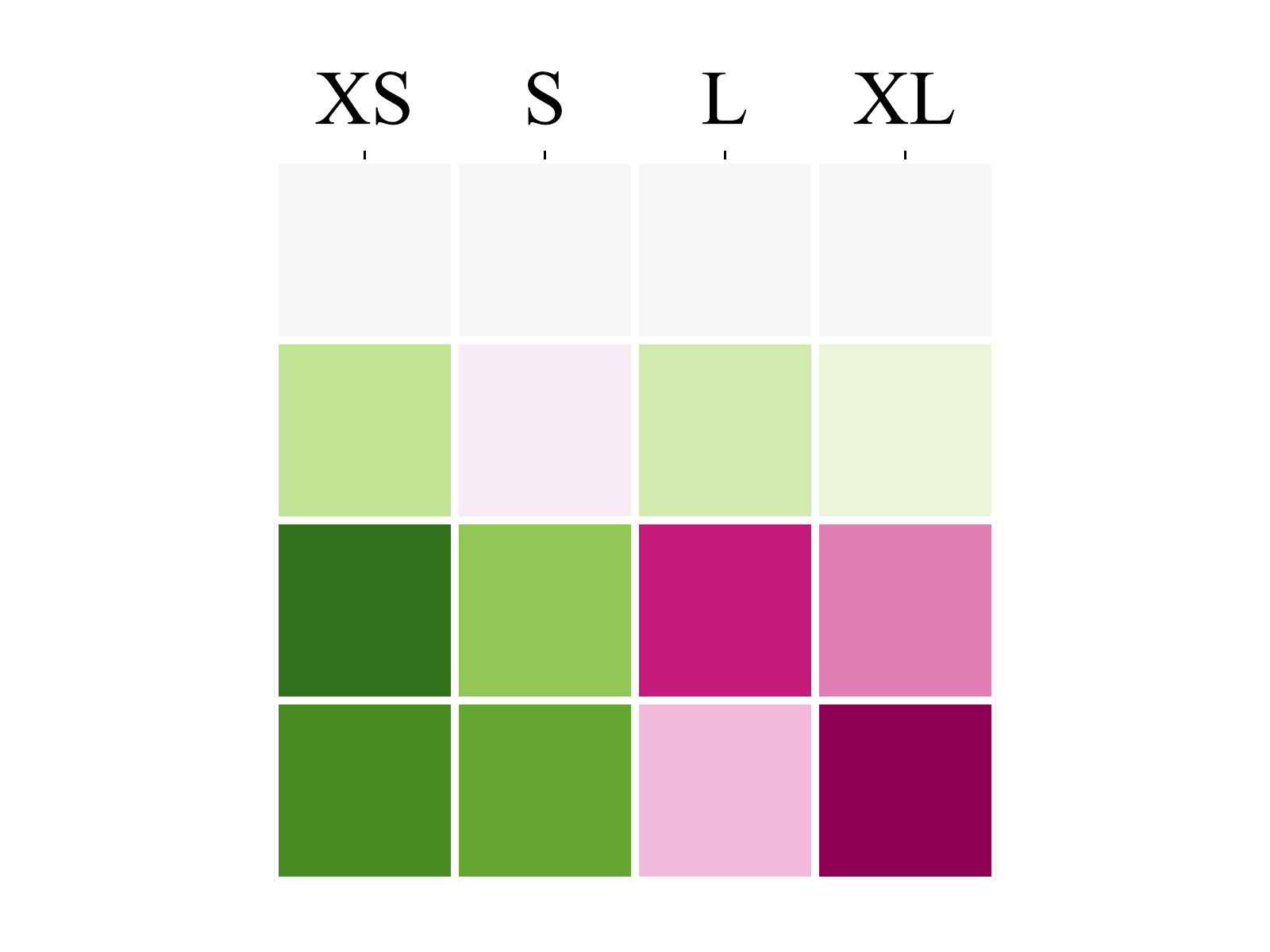}} 
        & \multicolumn{3}{c}{\includegraphics[scale=0.182]{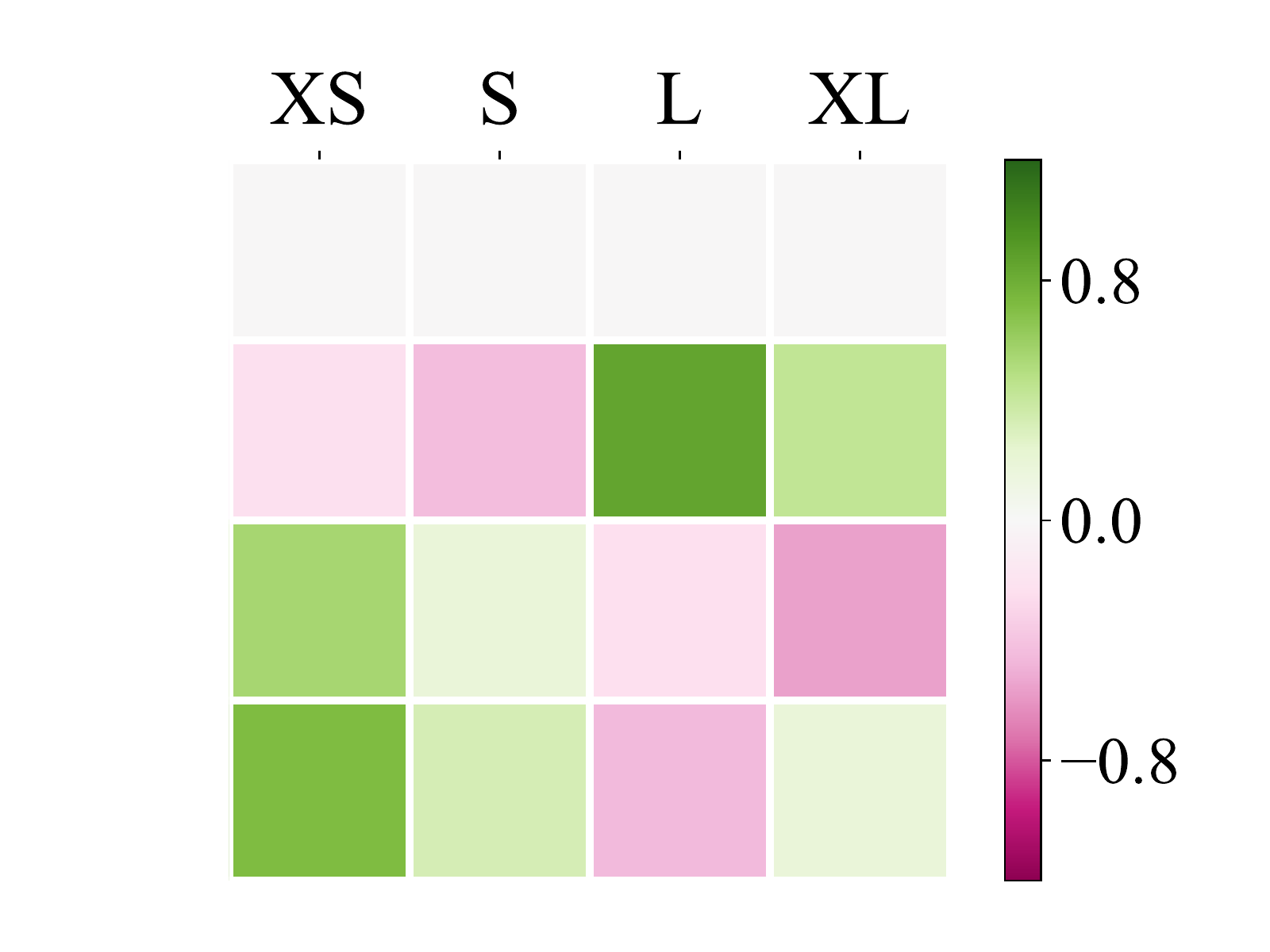}} 
     	  &\multicolumn{4}{c}{\includegraphics[scale=0.182]{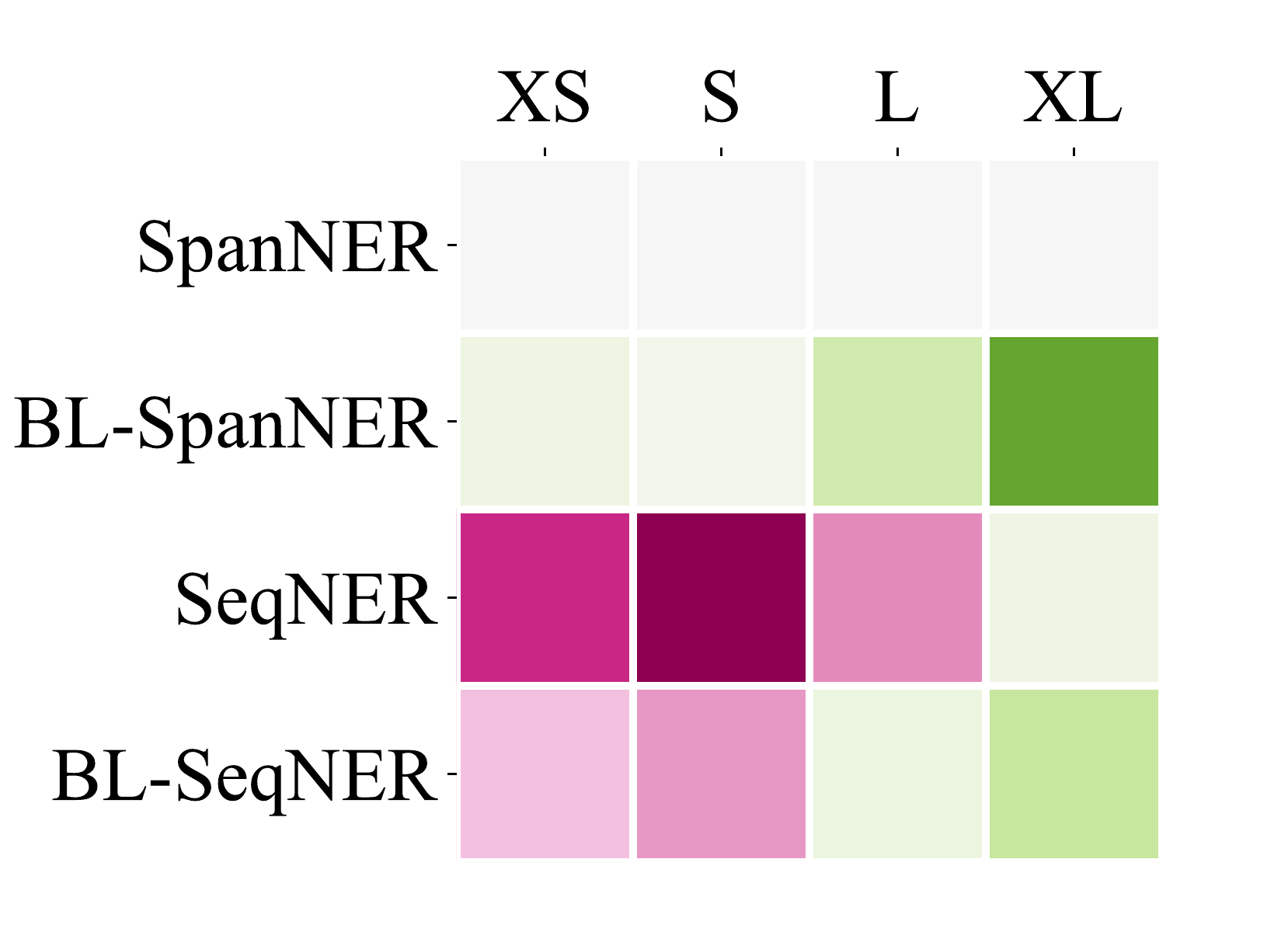}}  
        & \multicolumn{2}{c}{\includegraphics[scale=0.182]{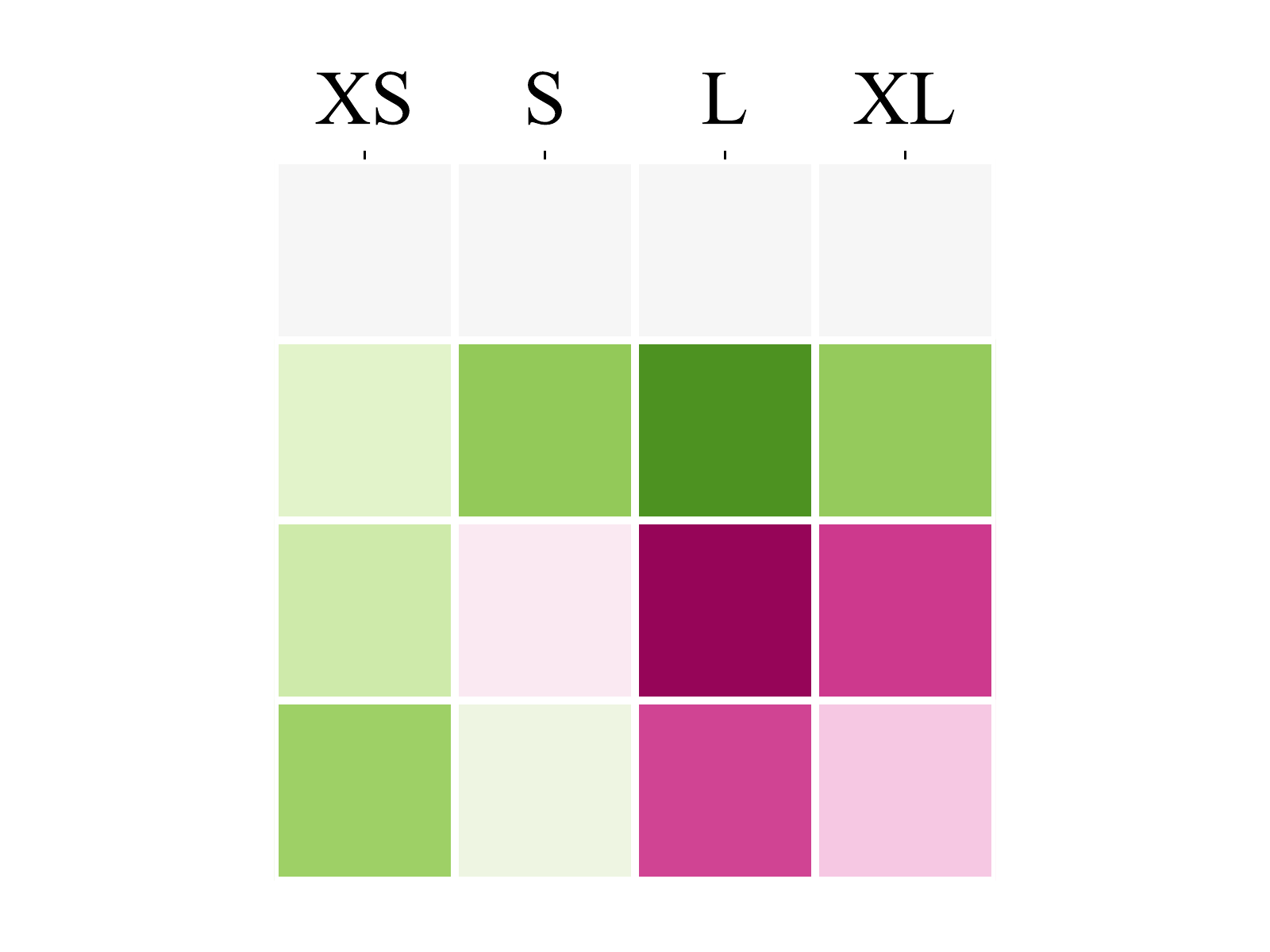}}                 
        & \multicolumn{2}{c}{\includegraphics[scale=0.182]{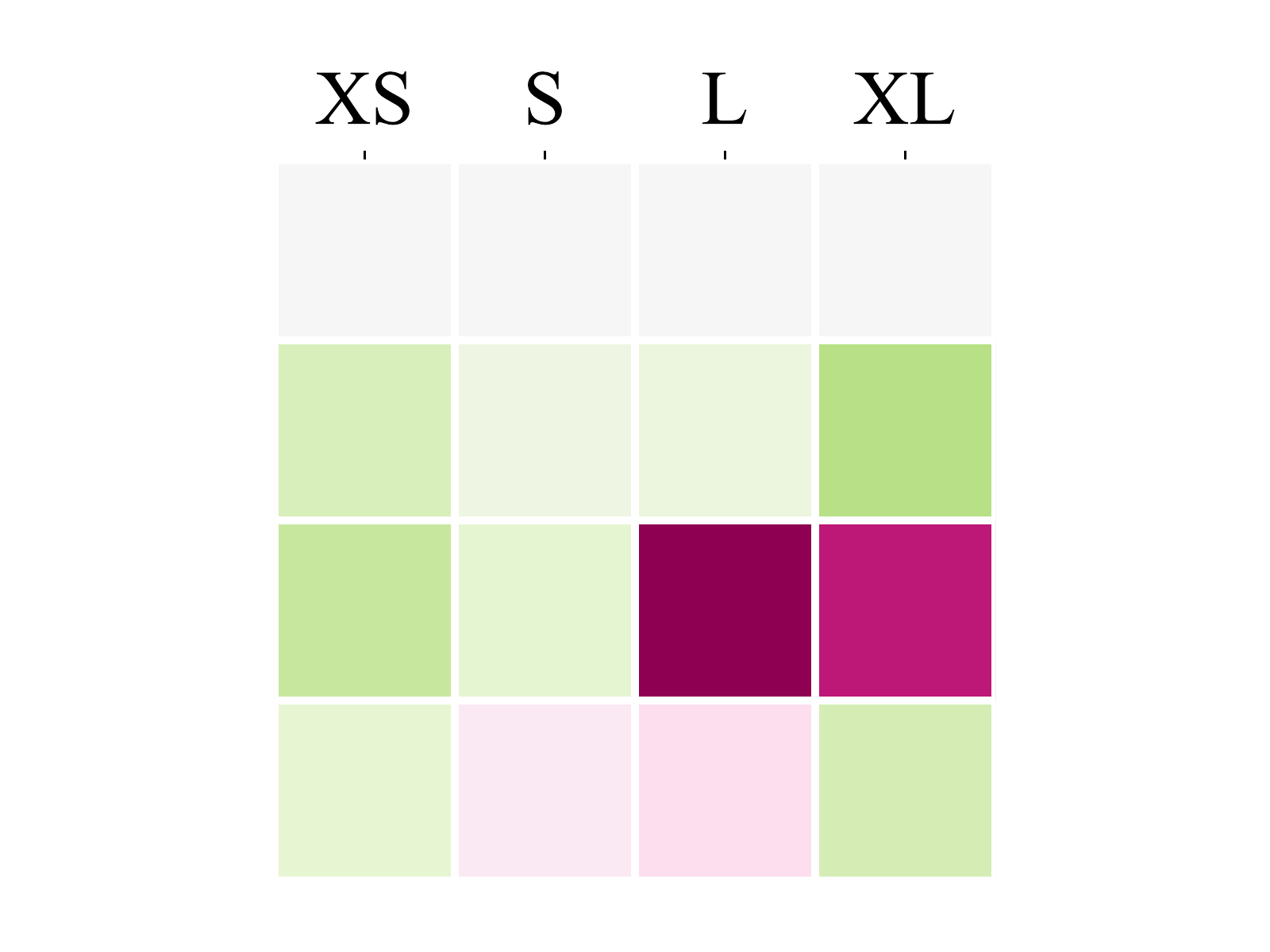}} 
        & \multicolumn{3}{c}{\includegraphics[scale=0.182]{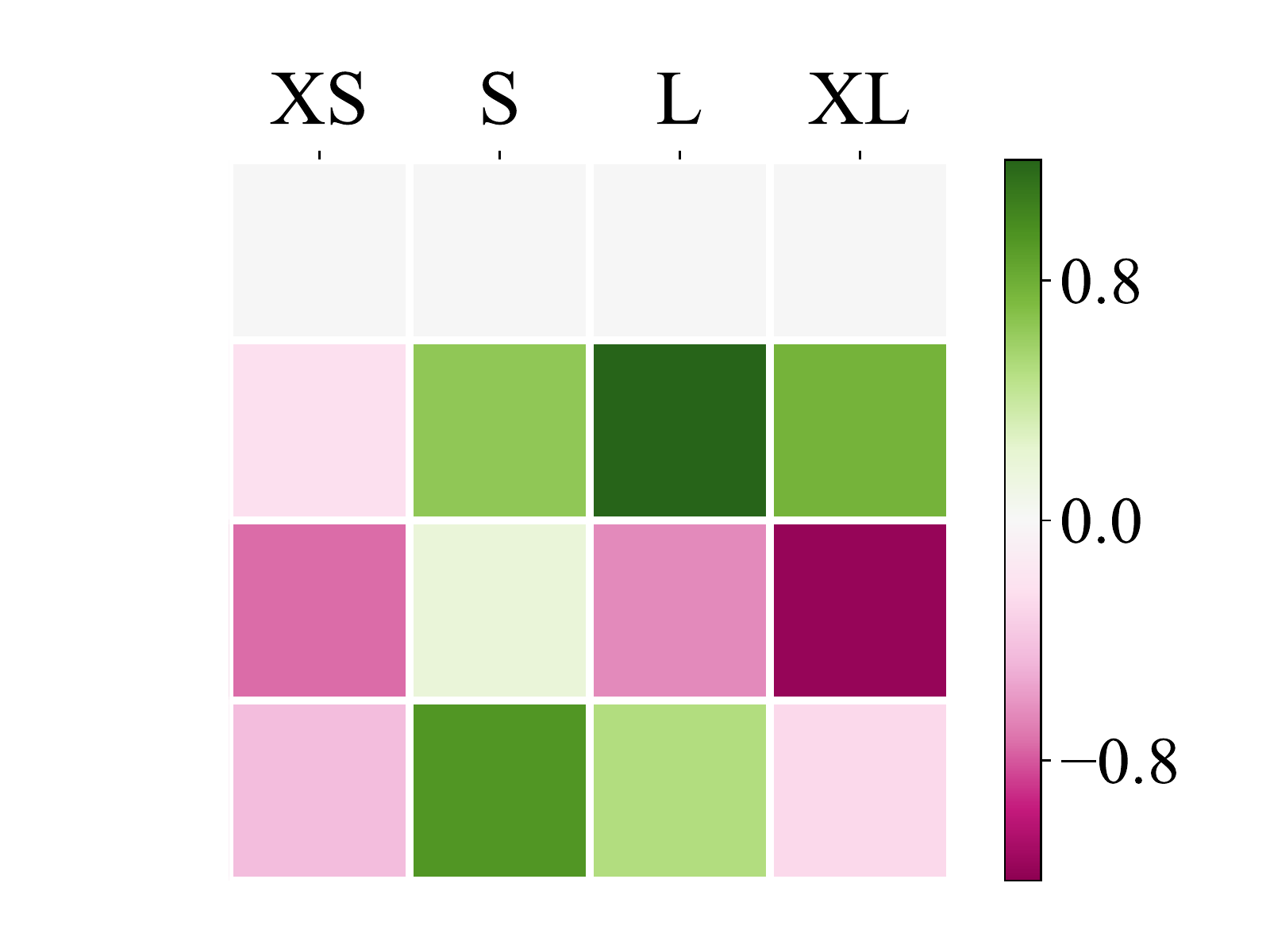}} 
        \\ \midrule \midrule
	\multicolumn{11}{c}{\textbf{W16}} & \multicolumn{11}{c}{\textbf{W17}}\\
    \cmidrule(lr){1-11}\cmidrule(lr){12-22}
    \multicolumn{4}{c}{\texttt{eLen}}      & \multicolumn{2}{c}{\texttt{tLen}} & \multicolumn{2}{c}{\texttt{eCon}} & \multicolumn{3}{c}{\texttt{eDen}} 
    & \multicolumn{4}{c}{\texttt{eLen}}      & \multicolumn{2}{c}{\texttt{tLen}} & \multicolumn{2}{c}{\texttt{eCon}} & \multicolumn{3}{c}{\texttt{eDen}}\\
    \cmidrule(lr){1-11}\cmidrule(lr){12-22}
           \multicolumn{4}{c}{\includegraphics[scale=0.182]{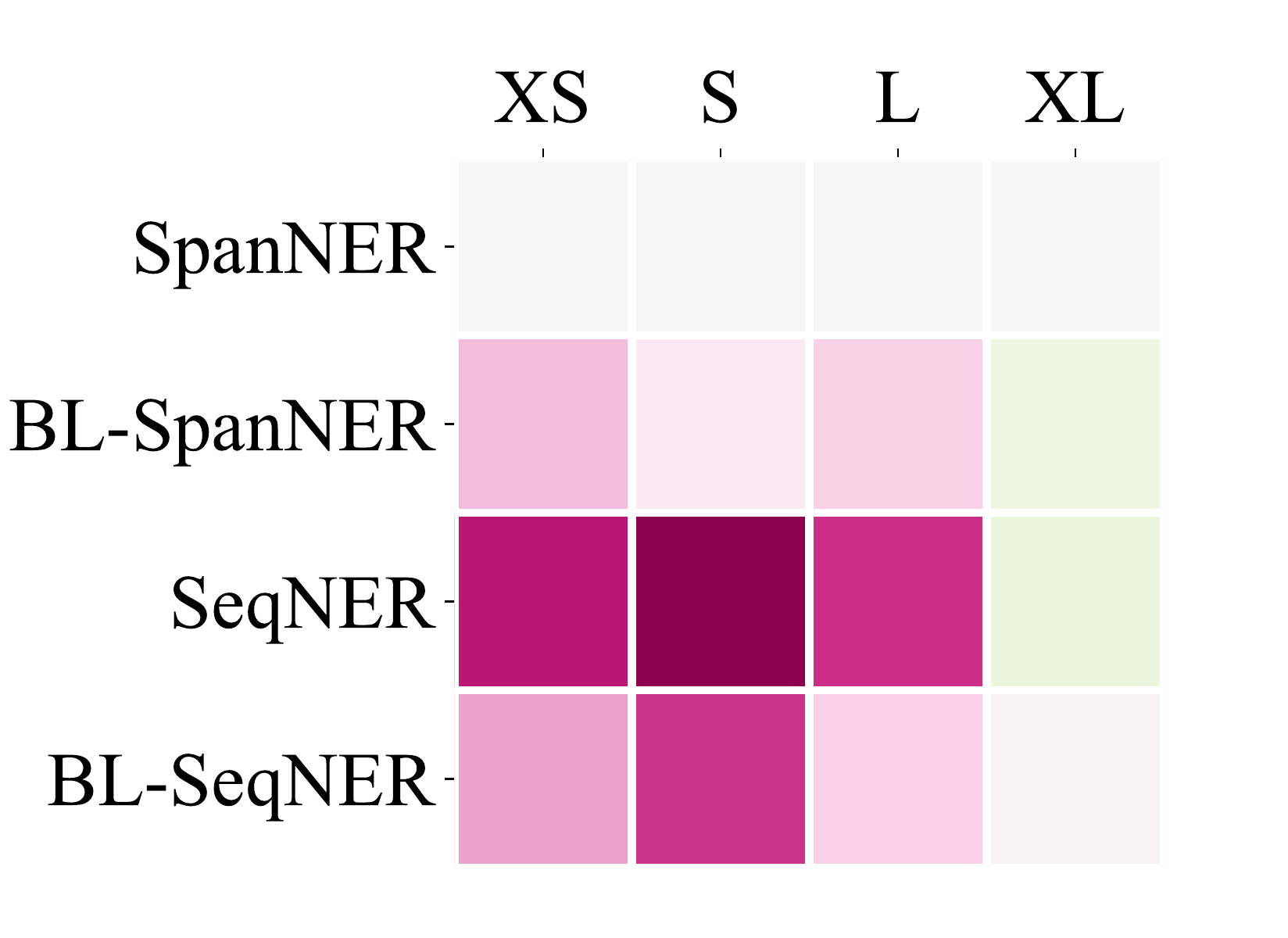}}  
        & \multicolumn{2}{c}{\includegraphics[scale=0.182]{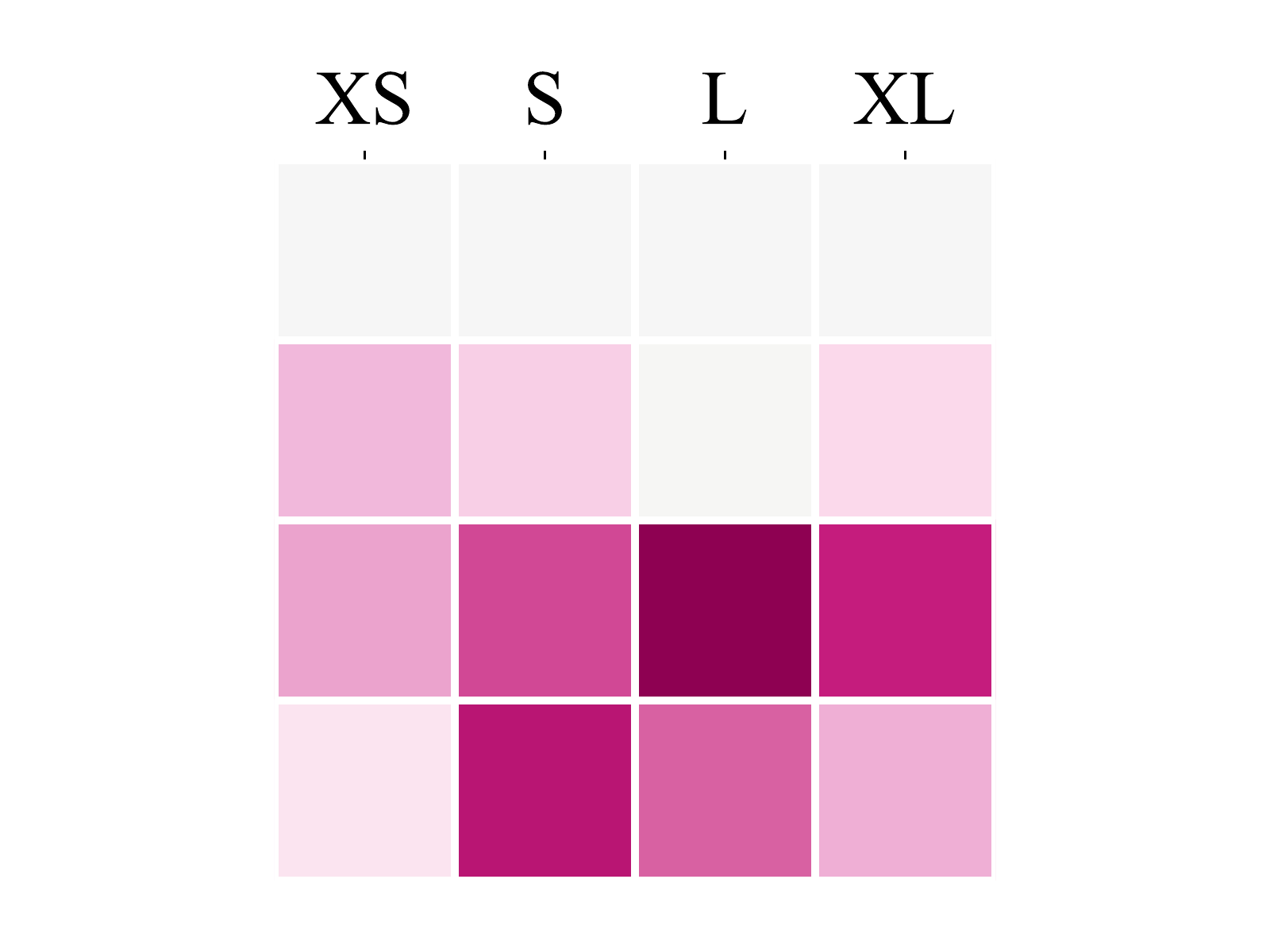}}                 
        & \multicolumn{2}{c}{\includegraphics[scale=0.182]{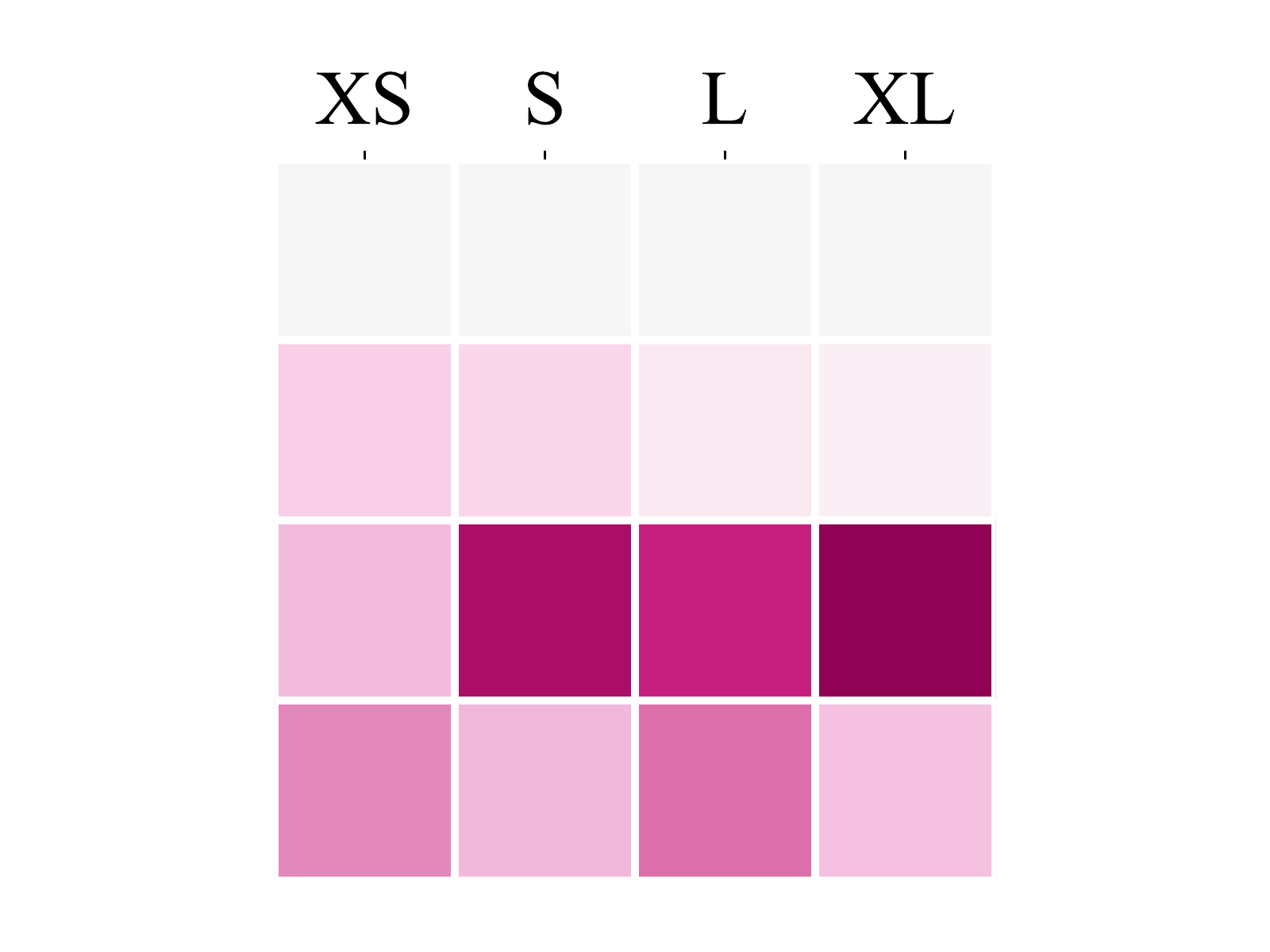}} 
        & \multicolumn{3}{c}{\includegraphics[scale=0.182]{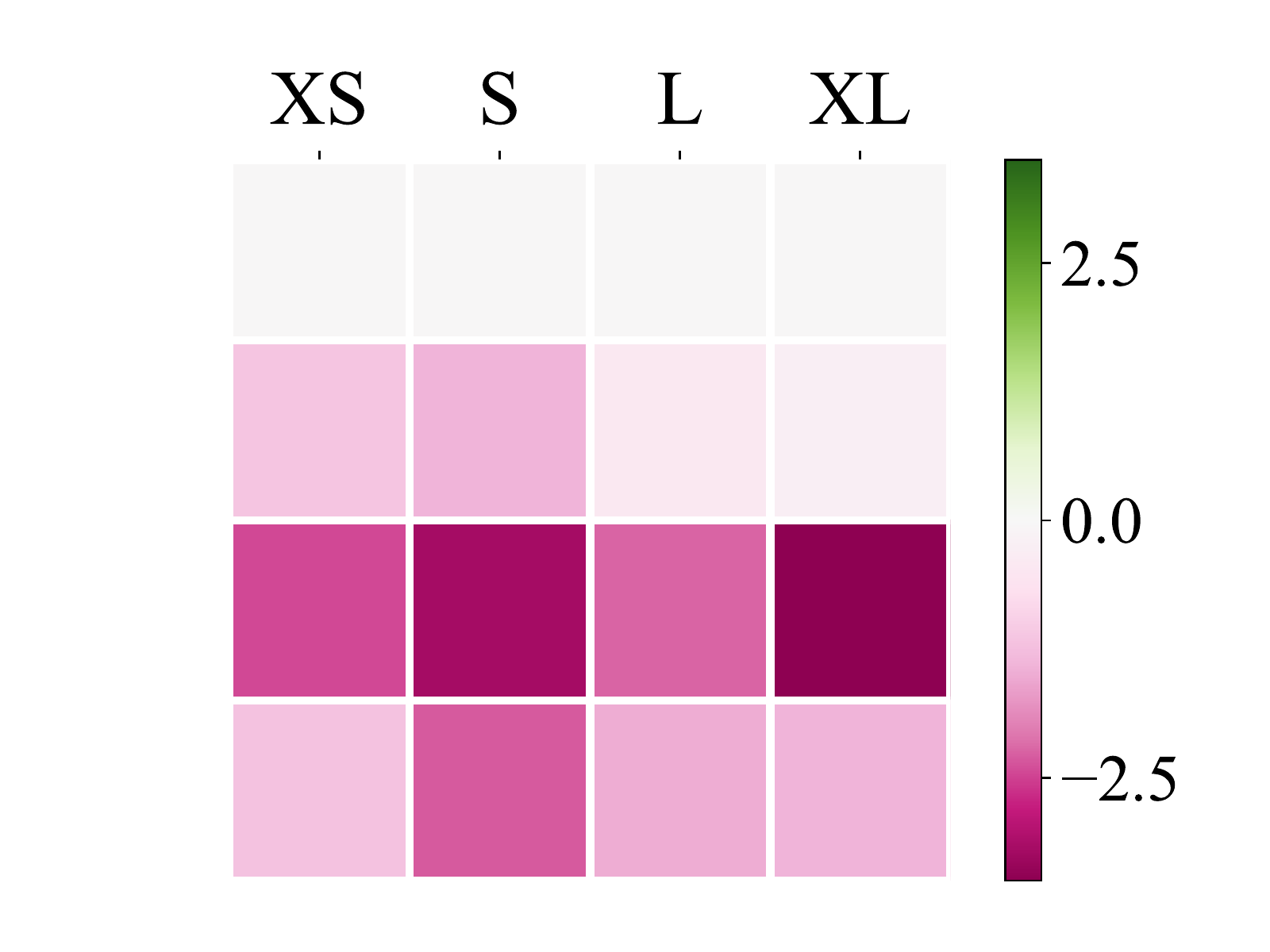}} 
     	  &\multicolumn{4}{c}{\includegraphics[scale=0.182]{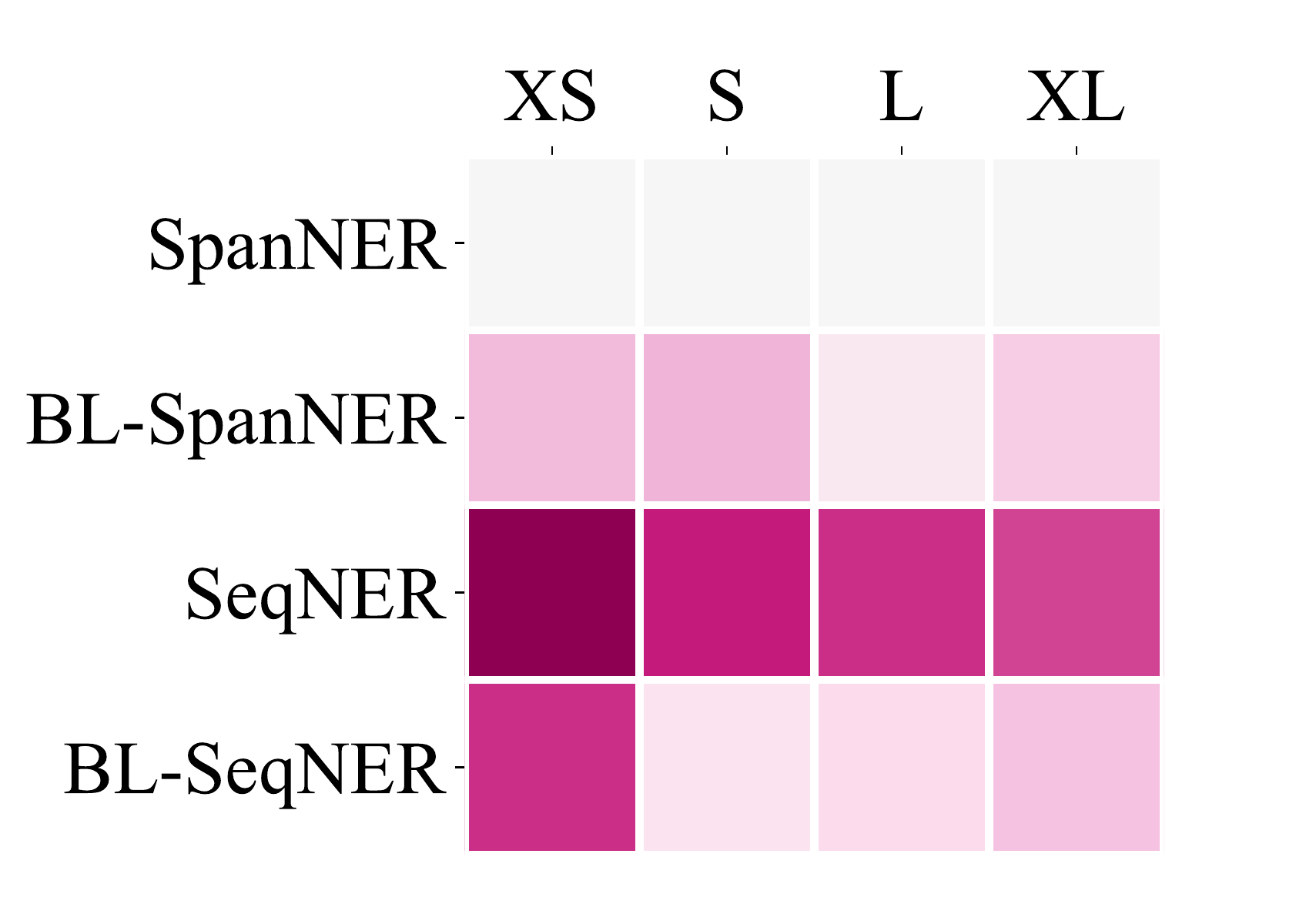}}  
        & \multicolumn{2}{c}{\includegraphics[scale=0.182]{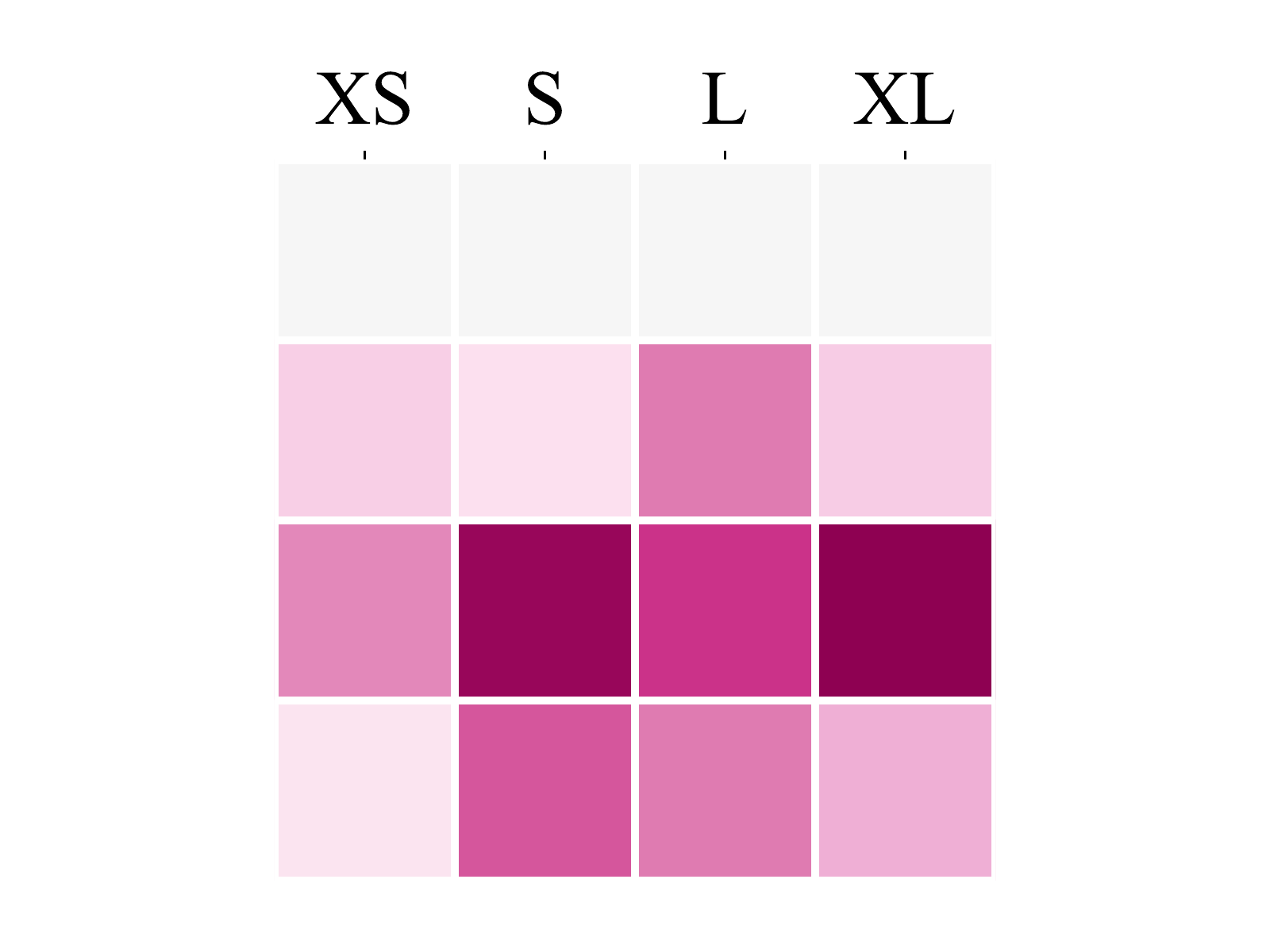}}                 
        & \multicolumn{2}{c}{\includegraphics[scale=0.182]{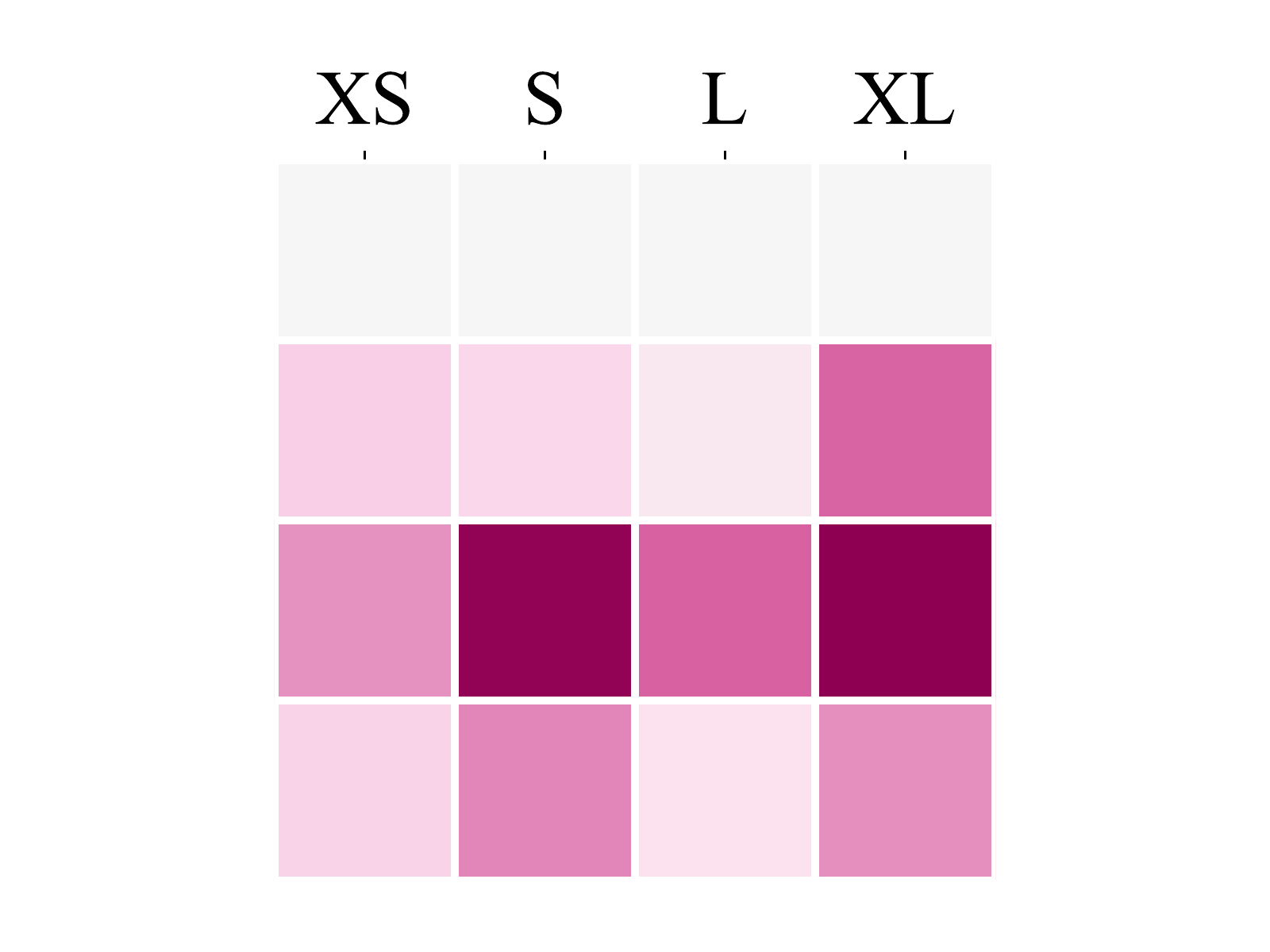}} 
        & \multicolumn{3}{c}{\includegraphics[scale=0.182]{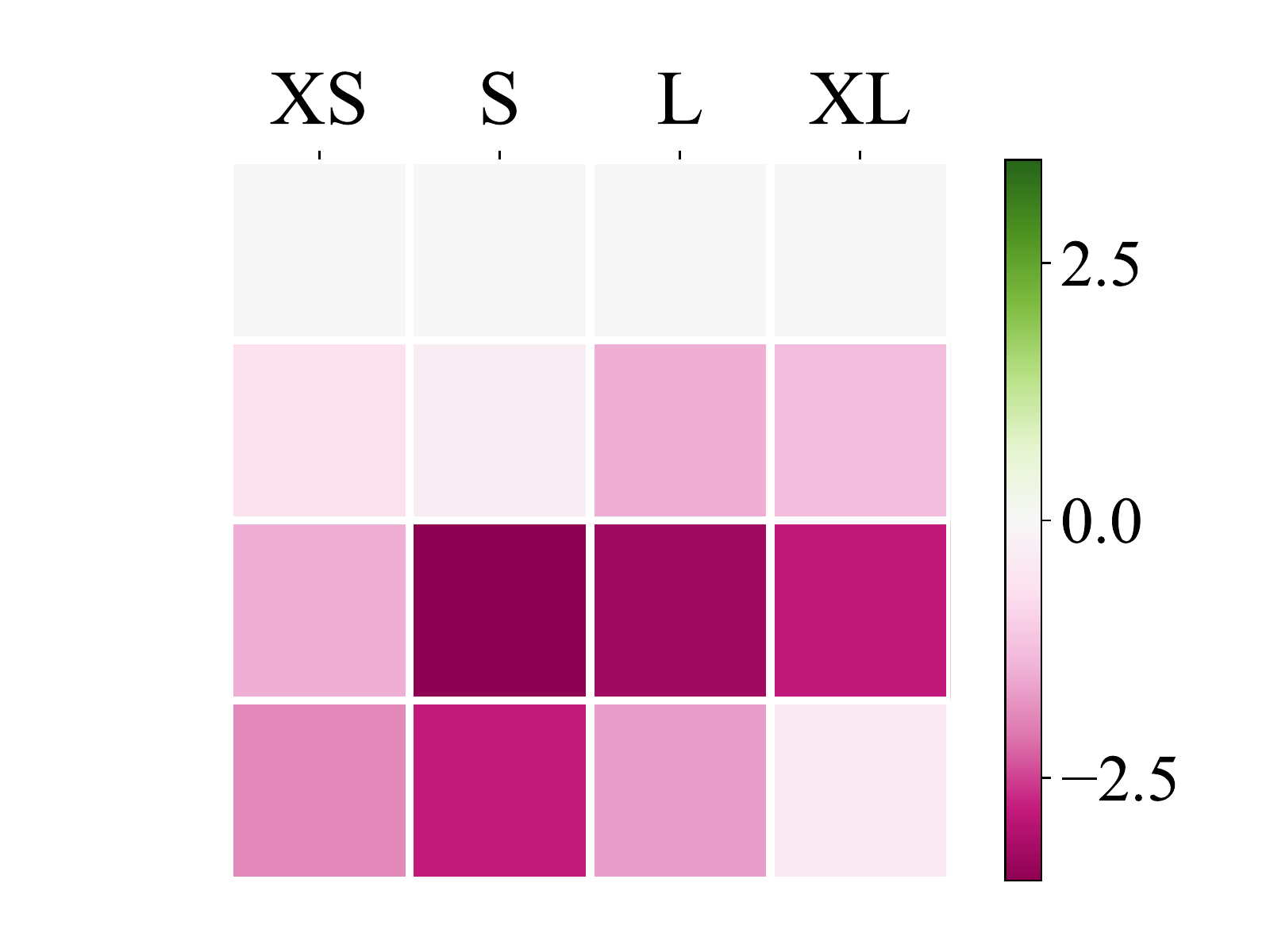}} 
        \\
    \bottomrule
    \end{tabular}     
\end{table*}

\noindent \textbf{Analysis.}
We report performance visualization results in Table \ref{complementaryheatmap}. We can observe that:
\begin{compactitem}
\item \textbf{$<$SpanNER, SeqNER$>$}. SpanNER shows clear complementary advantages with SeqNER. On the first four datasets, (1) SpanNER generally outperforms SeqNER when dealing with medium and long texts, short and medium entities, as well as large entity label consistency and entity density; (2) in contrast, SeqNER shows better performance than SpanNER when texts are short with long entities annotated, as well as small entity label consistency and entity {density}. On W16 and W17, we find that SpanNER significantly outperforms SeqNER in almost all cases. We attribute this to the fact that the test sets of two datasets have a larger OOV density (see the discussion of Section \ref{sectiondatasets}), and SpanNER is better at dealing with this scenario.

\item \textbf{$<$SpanNER, BL-SpanNER$>$}. The BL paradigm enables SpanNER to leverage almost all the complementary advantages of SeqNER on the first four datasets. For example, BL-SpanNER shows obvious performance gains over SpanNER when entities are long and texts are short. 
In some cases, however, BL-SpanNER shows slight performance declines, such as for entities with medium lengths and low densities.
While on W16 and W17, we observe that BL-SpanNER performs worse than SpanNER on almost all buckets of the four attributes. We attribute this to the severe disadvantages of SeqNER on the two datasets, suggesting that the BL paradigm can deliver disadvantages from one model to another.

\item \textbf{$<$SeqNER, BL-SeqNER$>$}. The BL paradigm allows SeqNER to leverage almost all the complementary advantages of SpanNER on all six datasets. For example, BL-SeqNER explicitly performs better than SeqNER in the case of short entities and long texts, and large entity density on the first four datasets. On W16 and W17, we can see that BL-SeqNER obviously outperforms SeqNER in almost all cases. Moreover, we discover that the disadvantages of SpanNER have imperceptive effects on SeqNER across all six datasets, which is attributed to the fact that the holistic better performance of SpanNER weakens the negative effects of these disadvantages.
\end{compactitem}

Based on the above observations, we draw the following conclusion:
\begin{compactitem}
\item \textbf{\textsc{Conclusion}} \#2:  The BL paradigm enables SeqNER and SpanNER to leverage their relative advantages that cover four attributes: entity length, text length, entity label consistency, and entity density. 
Additionally, the BL paradigm can deliver disadvantages from one model to another, leading to performance dropping on some attributes.
\end{compactitem}

\subsubsection{How Does BL Affect Entity Prediction Errors?}

In this section, we investigate how the BL paradigm affects two common entity prediction errors, which are boundary error (B.E.) and type error (T.E.). We formalize their definitions as follows:
\begin{compactitem}
\item \textsc{Boundary Error.} If a predicted entity contains more or fewer tokens than a ground truth entity, we call the unaligned entity boundary a boundary error.
\item \textsc{Type Error.} If a predicted entity and a ground truth entity have the same boundary but different types, we call the mistakenly predicted type a type error.
\end{compactitem}

Take the running example in Figure 1 as an example. Two ground truth entities, i.e., ``rain'' of the \texttt{Weather} type and ``this night'' of the \texttt{Date} type, are annotated in the text ``will it rain this night''. A boundary error happens if the ``it rain'' is predicted as a \texttt{Weather} entity. The same goes for ``will it rain'', ''rain this'' etc. And if the ''rain'' is predicted as any other type except for the \texttt{Weather}, a type error happens.

We calculate the rates of the two errors as follows:
\begin{subequations}
\begin{normalsize}
\begin{align}
\textrm{BE-Rate} &= \frac{\textrm{B.E.}}{\textrm{FP}+\textrm{FN}},\\
\textrm{TE-Rate} &= \frac{\textrm{T.E.}}{\textrm{FP}+\textrm{FN}},
\end{align}
\end{normalsize}
\end{subequations}
where we use the B.E. and T.E. to denote the instance number of the two errors, respectively.

We report the error analyses in Table \ref{erroranalysis}. We can observe that: (1) SpanNER shows much smaller type error rates than SeqNER, while SeqNER deals with boundary errors much better than SpanNER. These results reveal that the two NER models have complementary advantages regarding entity boundary and entity type predictions.
(2) BL-SpanNER generally reduces both errors compared to SpanNER on the first four datasets. And similar results can also be seen on BL-SeqNER compared to SeqNER. These results demonstrate that the BL paradigm enables the two NER models to leverage their relative complementary advantages.
(3) On W16 and W17, the BL paradigm reduces the two errors in SeqNER but increases the two errors in SpanNER. Similar phenomena have been observed in {Section} \ref{sectionmainresults} and \ref{sectioncomplementary}, which we attribute to the specific characteristic of the two datasets (see the discussion in Section \ref{sectiondatasets}).

\renewcommand\tabcolsep{1.5pt}
\begin{table}[h]
  \centering 
\caption{Performance heatmaps regarding entity prediction errors. SpanNER is taken as a standard. \textbf{The green (red) area indicates that the error rate is higher (lower) than SpanNER.}}
  \label{erroranalysis}
    \begin{tabular}{rrrrrrrrrrrrrrr}
    \toprule
    \multicolumn{4}{c}{\textbf{OntoNotes}}      & \multicolumn{2}{c}{\textbf{CoNLL03}} & \multicolumn{2}{c}{\textbf{BC5CDR}}&  \multicolumn{2}{c}{\textbf{SciERC}}      & \multicolumn{2}{c}{\ \ \ \ \ \textbf{W16}\ \ \ \ \ } & \multicolumn{3}{c}{\textbf{W17}}
   \\
    \midrule
           \multicolumn{4}{c}{\includegraphics[scale=0.182]{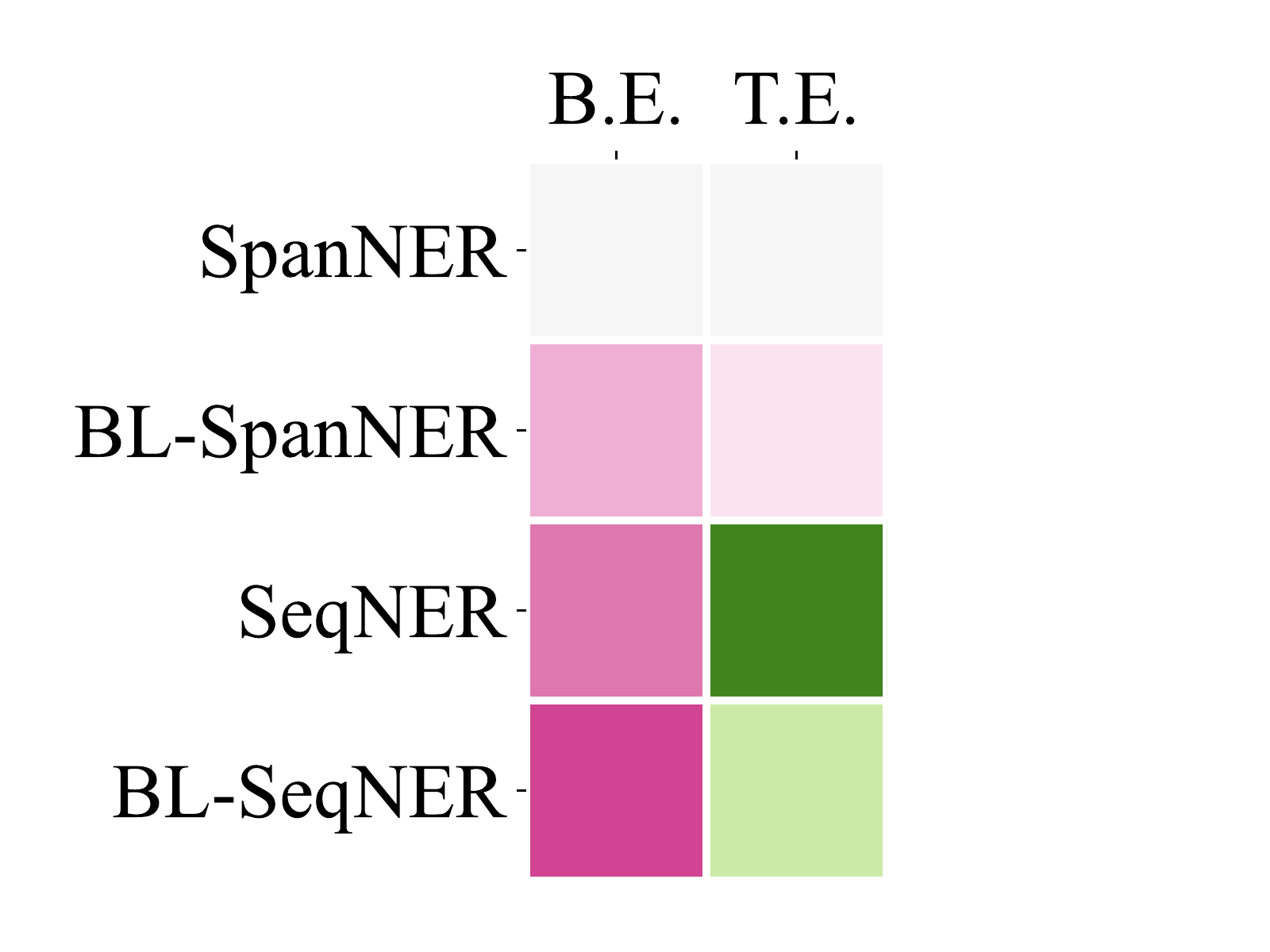}}  
        & \multicolumn{2}{c}{\includegraphics[scale=0.182]{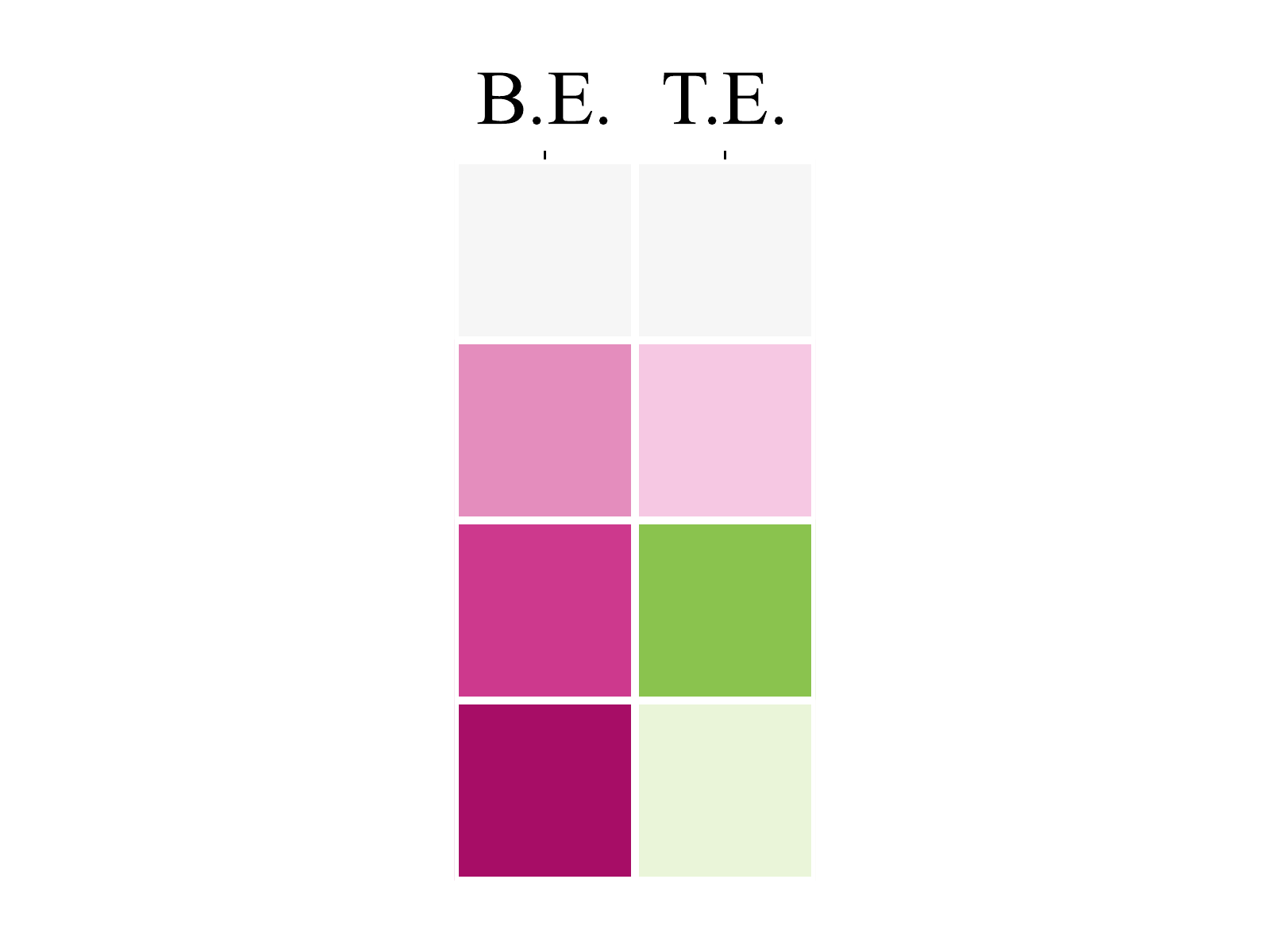}}                 
        & \multicolumn{2}{c}{\includegraphics[scale=0.182]{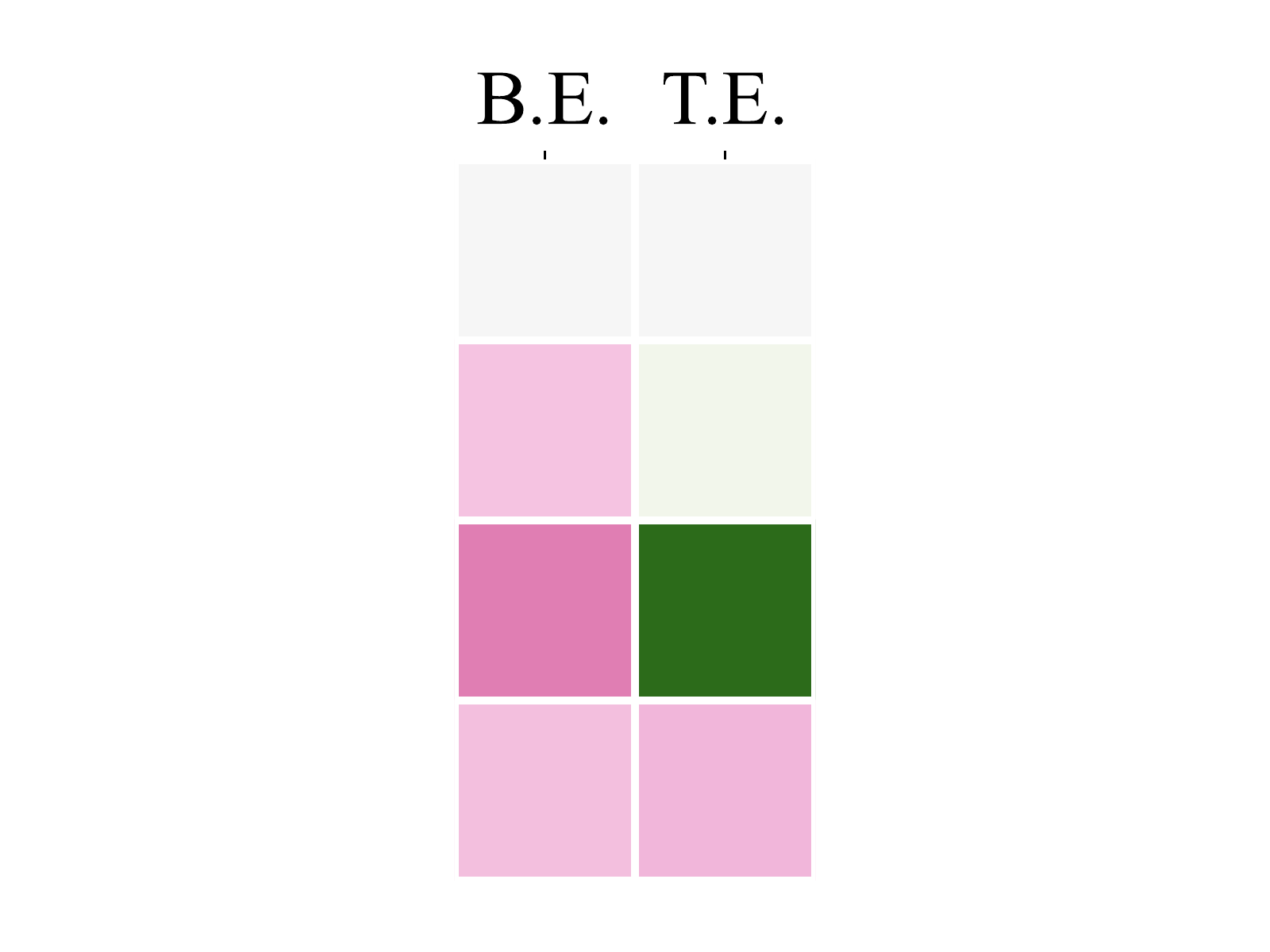}} 
     	  &  \multicolumn{2}{c}{\includegraphics[scale=0.182]{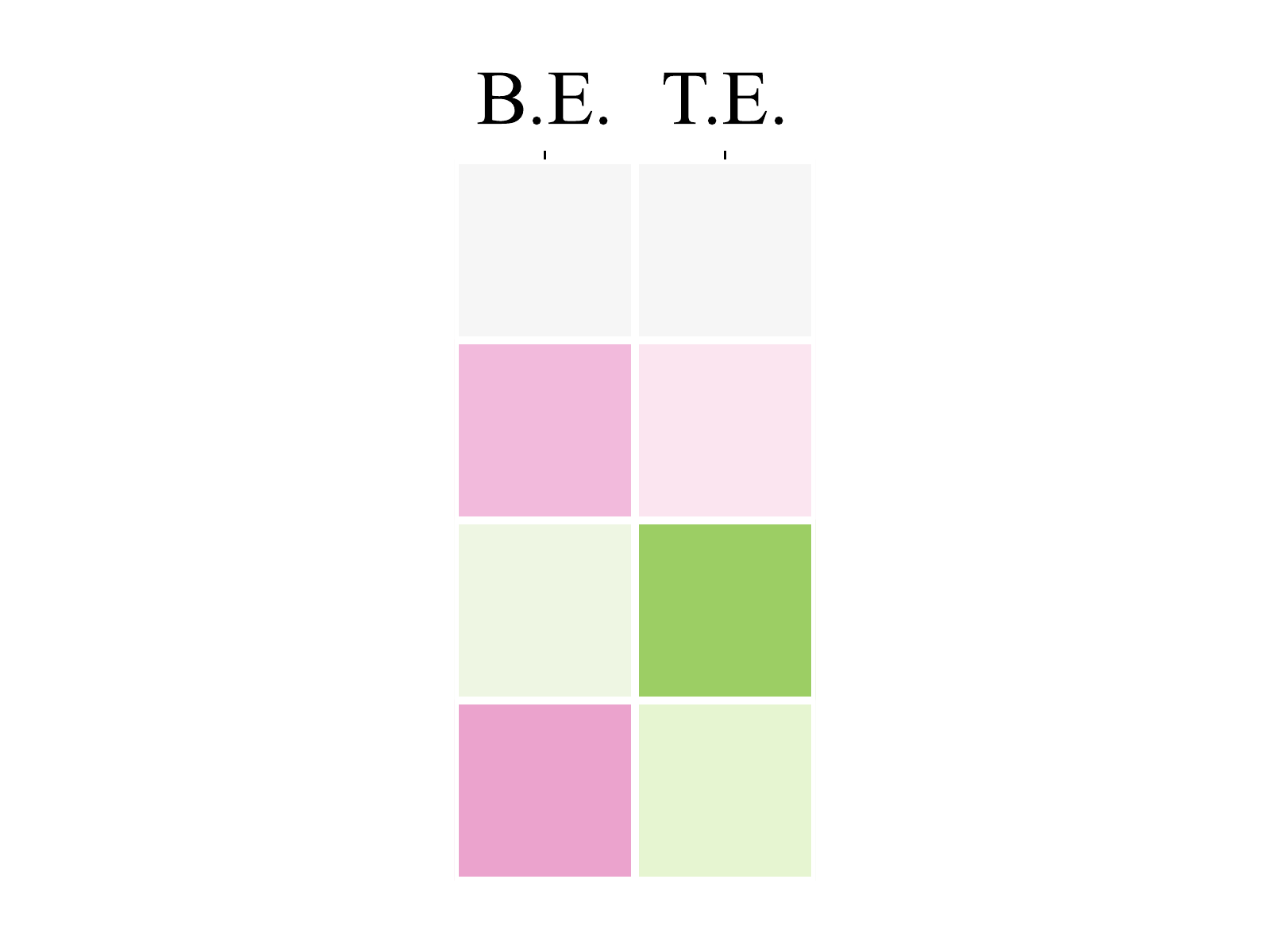}}  
        & \multicolumn{2}{c}{\includegraphics[scale=0.182]{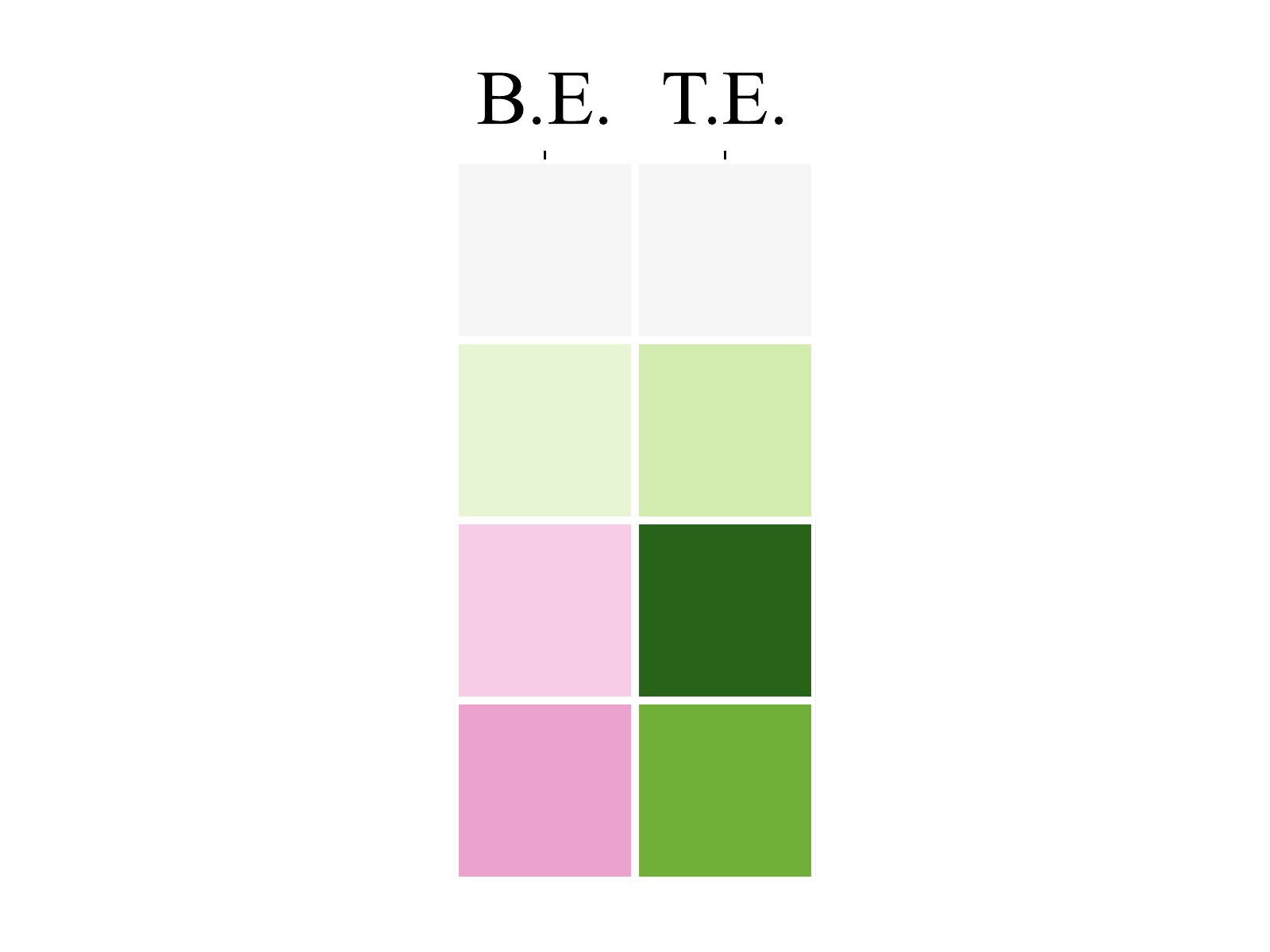}}                 
        & \multicolumn{3}{c}{\includegraphics[scale=0.182]{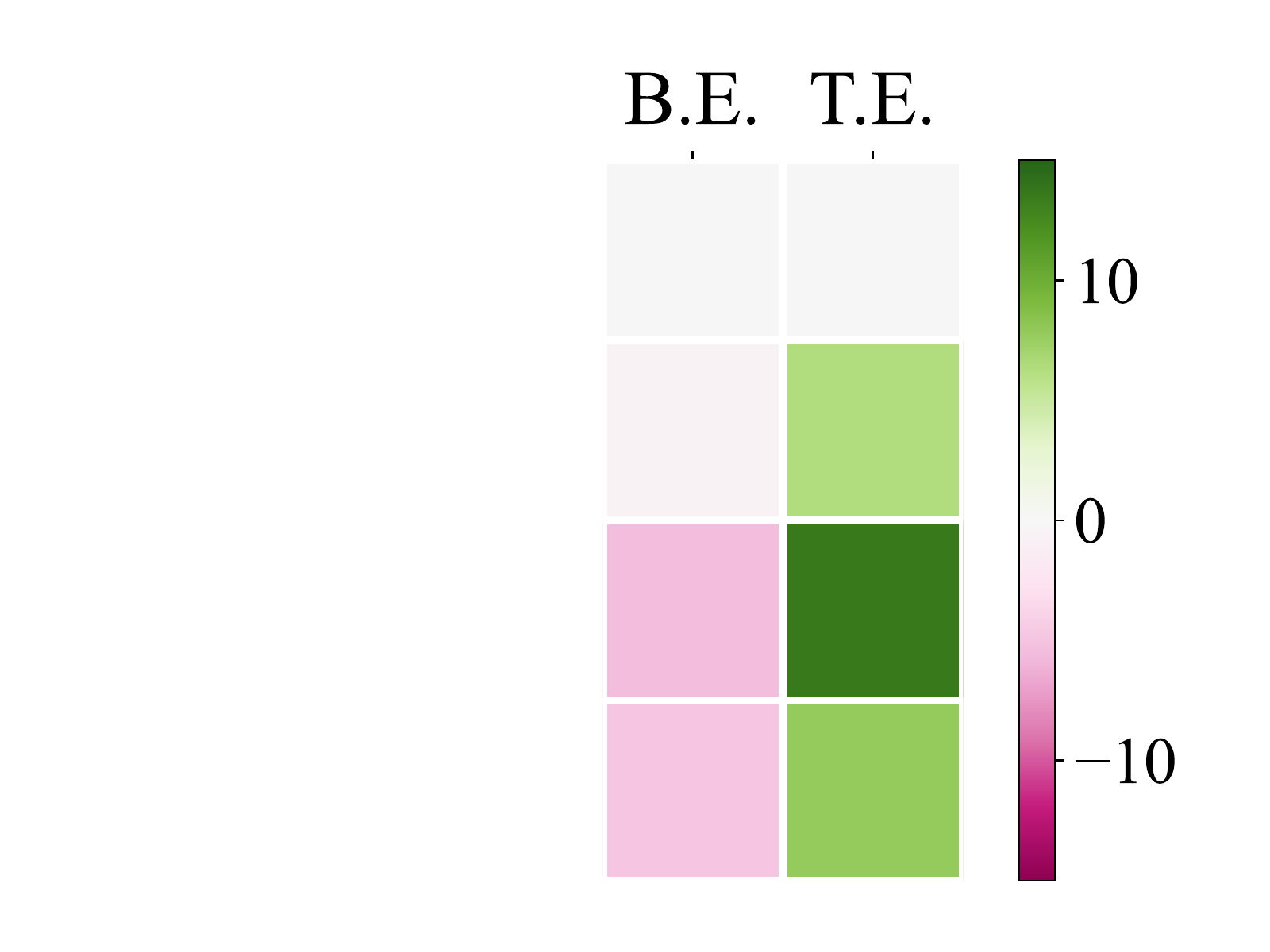}} 
        \\ 
    \bottomrule
    \end{tabular}     
\end{table}

\subsection{Applying BL to Existing SOTA NER Models}
In this section, we explore whether the BL paradigm works in existing SOTA NER models. Specifically, for those NER models adopting the sequence labeling-based paradigm, we add the Span Splitter and the Span Decoder module to the top of the model's encoder, which is similar to Figure \ref{bl-ner}-\ding{193}\&\ding{192}. 
We regard the combination of the two added modules and the encoder as a span-based NER model. 
For those NER models adopting the span-based paradigm, we add the Seq Decoder and the Label Combiner module to the top of the model's encoder, which is similar to Figure \ref{bl-ner}-\ding{195}\&\ding{194}. 
We regard the combination of the two added modules and the encoder as a sequence labeling-based NER model.
We then sum the losses of two bundled models to jointly train the encoder parameters. We set the loss weight $\alpha$ to 0.1, which is the same as the one described in Section \ref{details}.

We select five existing SOTA NER models for the exploration. Three are sequence labeling-based models, which are named RDANER \cite{rdaner}, CL-KL \cite{clkl}, and Hire-NER \cite{hirener}. Two are span-based models, which are named SodNER \cite{lifei} and DyGIE++ \cite{dygie++}. Note that DyGIE++ is a span-based model for joint entity and relation extraction, and we just use its NER part for the exploration. Moreover, SodNER takes overlapped and discontinuous entities into account, and we exclude these entities during the exploration. For each of the above NER models, we compare it with its BL enhanced model, where the two models use the same method to output entities for a fair comparison.

We report the model performance (measured by F1) in Table \ref{explorationresults}, from which we observe that the BL paradigm consistently improves the performance of the five NER models. Specifically, it brings +0.15\% to +1.73\% F1 scores to the three sequence labeling-based NER models, and it boosts the two span-based NER models by +0.31\% to +1.56\% F1 scores.

Based on the above results, we conclude that given a current SOTA NER system that adopts either the sequence labeling-based or the span-based paradigm, the BL paradigm makes it possible to obtain a new SOTA system without requiring additional resources, such as external data annotations.

\renewcommand\tabcolsep{8.5pt}
\begin{table}[h]
  \centering
  \caption{Results of applying the BL paradigm to five existing SOTA NER models, where \textsc{Seq} denotes sequence labeling-based models and \textsc{Span} denotes span-based models. $\dag$ represents we reproduce the results by removing overlapped or discontinuous entities from the datasets. $\ddag$ represents we only use the NER part of the model.}\label{explorationresults}
    \begin{tabular}{llccc}
    \toprule
    \multirow{8.6}[12]{*}{\textsc{Seq}} & Model & SciERC & BC5CDR & NCBI \\
\cmidrule{2-5}          & RDANER & 68.96 & 87.39 & 87.89 \\
          & BL-RDANER &\textbf{69.44}       & \textbf{89.12}       & \textbf{88.47}  \\
\cmidrule{2-5}
\specialrule{0em}{-1pt}{-1pt}
\cmidrule{2-5}          
& Model & CoNLL03 & BC5CDR & NCBI \\
\cmidrule{2-5}          & CLKL  & 93.21 & 90.73 & 89.24 \\
          & BL-CLKL & \textbf{93.86}       &\textbf{90.88}       &\textbf{89.94}  \\
\cmidrule{2-5}
\specialrule{0em}{-1pt}{-1pt}
\cmidrule{2-5}          
& Model & CoNLL02 & CoNLL03 & OntoNotes \\
\cmidrule{2-5}          & HireNER & 87.08 & 93.37 & 90.30 \\
          & BL-HireNER &\textbf{88.16}       &\textbf{93.68}       &\textbf{90.84}  \\
    \midrule
    \multirow{5}[8]{*}{\textsc{Span}} & Model & CLEF  & CADEC & ACE05 \\
\cmidrule{2-5}          & SodNER  &\ \  82.42 $\dag$ &\ \  60.38 $\dag$ &\ \  88.58 $\dag$ \\
          & BL-SodNER & \textbf{83.17}        & \textbf{60.84}       & \textbf{88.89}   \\
\cmidrule{2-5}
\specialrule{0em}{-1pt}{-1pt}
\cmidrule{2-5}          
& Model & ACE05 & SciERC & GENIA \\
\cmidrule{2-5}          & DyGIE++ $\ddag$ &\ \  88.96 $\dag$ & 67.23 &\ \  72.24 $\dag$ \\
          & BL-DyGIE++ &\textbf{89.44}       &\textbf{68.79}       & \textbf{73.12}  \\
    \bottomrule
    \end{tabular}%
  \label{tab:addlabel}%
\end{table}%

\subsection{Performance against Model variants}\label{sectionparameter}
The purpose of this section is to examine the correlation between model performance and three model variants.
We conduct explorations on the six datasets selected in Section \ref{sectionwhy}.

\subsubsection{Performance against the Loss Weight}
The loss weight ($\alpha$, see Eq. \ref{equloss}) is a critical hyperparameter that controls the contributions of SeqNER (the weight score is $\alpha$) and SpanNER (the weight score is 1-$\alpha$) to BL-NER, and we demonstrate that the larger the weight score, the larger the model contribution. When setting $\alpha$ to 0, SeqNER contributes nothing to BL-NER. At this moment, BL-NER is SpanNER in essence. Similarly, BL-NER is SeqNER when setting $\alpha$ to 1. 

To explore the influence of various $\alpha$ acores, we run BL-NER with setting $\alpha$ to 0.1, 0.2, 0.3,..., 0.9 and report the performance of both BL-SeqNER and BL-SpanNER, as shown in Figure 5. We have the following observations: (1) Performance of the two models generally decreases when the $\alpha$ score consistently increases. (2) When setting the $\alpha$ score to 0.1, BL-SeqNER performs the best on five of the six datasets, and BL-SpanNER shows the best performance on three of the six datasets. 

We attribute the above observations to the fact that SpanNER consistently surpasses SeqNER on the six datasets (see Table \ref{mainresults}). Thus a larger contribution of SpanNER ensures a better model performance, and a minor contribution of SeqNER allows the two models to leverage their complementary advantages. For simplicity, we set $\alpha$ to 0.1 in all the other experiments.

\begin{figure}[h]
\centering
\caption{Performance against various loss weight ($\alpha$) scores.}
\includegraphics[width=0.486\textwidth]{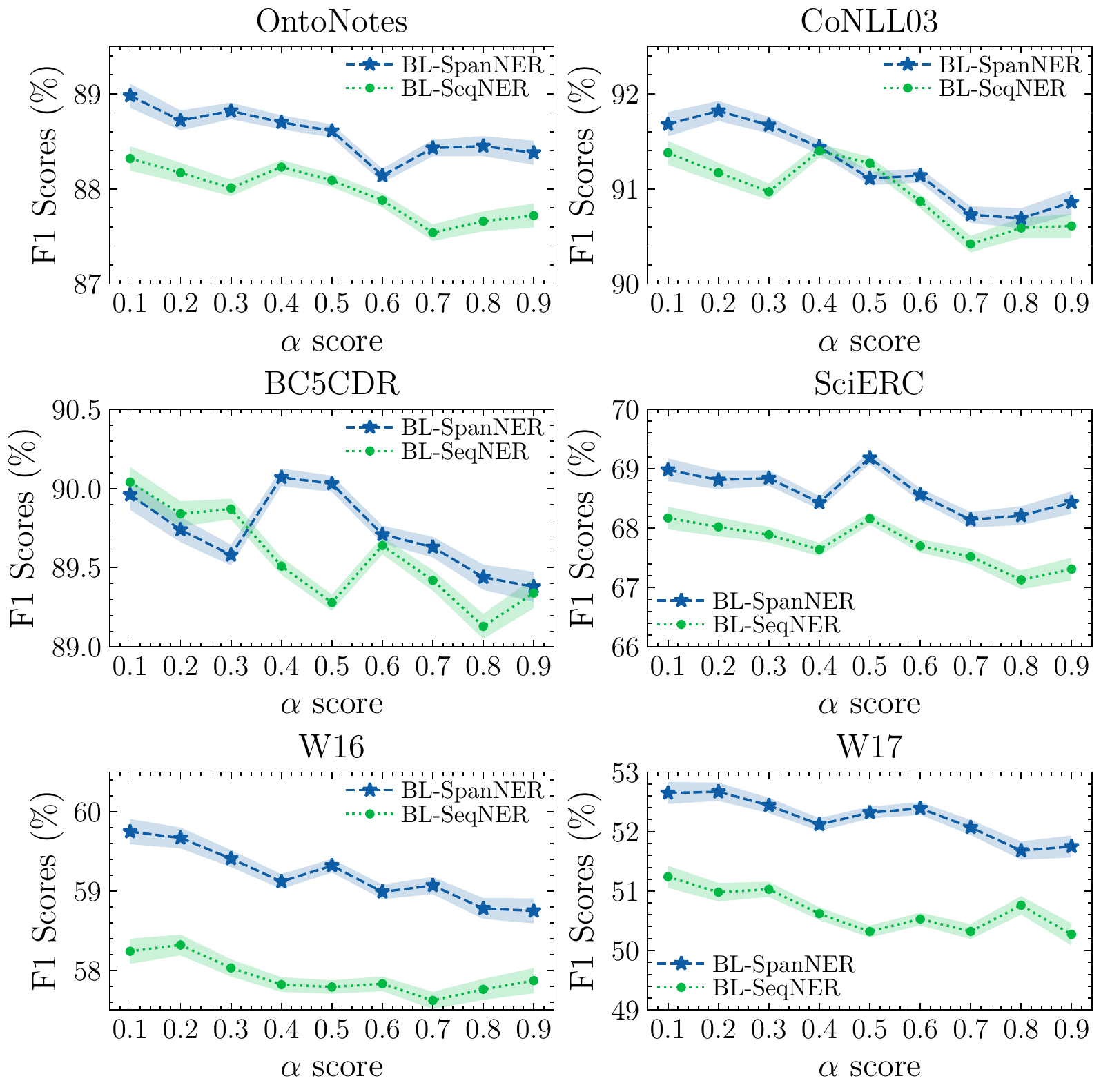} 
\end{figure}

\subsubsection{Performance against the Label Tagging}
In Section \ref{sectionseqner} we propose two label tagging methods: softmax-based and CRF-based. In this section, we investigate their effectivenesses in both BL-SeqNER and BL-SeqNER. Since BL-SpanNER can be affected by the bundled SeqNER model, we also take BL-SpanNER into consideration.

\begin{figure}[h]
\centering
\caption{Performance against various label tagging methods.}\label{labeltagging}
\includegraphics[width=0.486\textwidth]{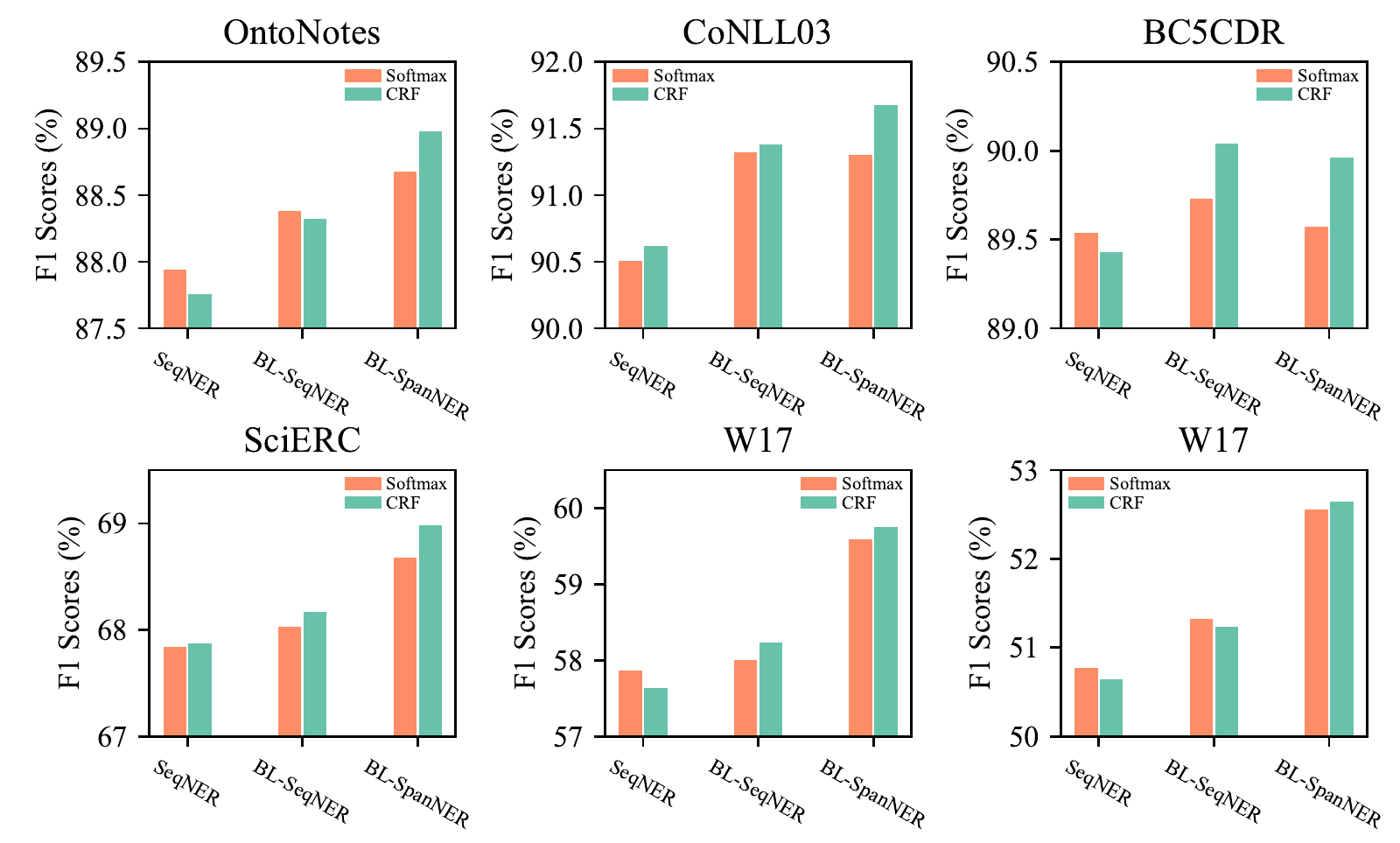} 
\end{figure}

We report the investigation results in Figure \ref{labeltagging}, from which we observe that: (1) For SeqNER, the CRF-based tagging performs better than the softmax-based tagging on two of the six datasets. And for BL-SeqNER, the CRF-based tagging beats the softmax-based tagging on four datasets. 
These results indicate that the complex CRF-based tagging does not always lead to better performance in comparison with simple softmax-based tagging, which conforms to the conclusion drawn by Hanh et al. \cite{hanh2021named}. 
As for BL-SpanNER, the CRF-based tagging consistently performs the best across the six datasets. (2) Despite the CRF-based tagging showing advantages in most cases, it actually delivers slight performance gains (averaged +0.11\% F1).

Moreover, we find that the CRF-based tagging takes much more time than the softmax-based tagging. 
In all the other experiments, we actually use the softmax-based tagging, which is a trade-off between the model performance and the training efficiency. 

\subsubsection{Performance against the Span Representation}

In Section \ref{sectionspanner}, we design three methods to obtain span semantic representations. In this section, we conduct experiments on these methods to explore their effectiveness. Specifically, we use SpanNER, BL-SpanNER, and BL-SeqNER for the exploration. It is obvious that BL-SpanNER and BL-SpanNER are directly affected by these methods. We demonstrate that BL-SeqNER can also be affected indirectly due to its bundled SpanNER model.

\begin{figure}[h]
\centering
\caption{Performance against various span representations.}\label{spanrepresentation}
\includegraphics[width=0.486\textwidth]{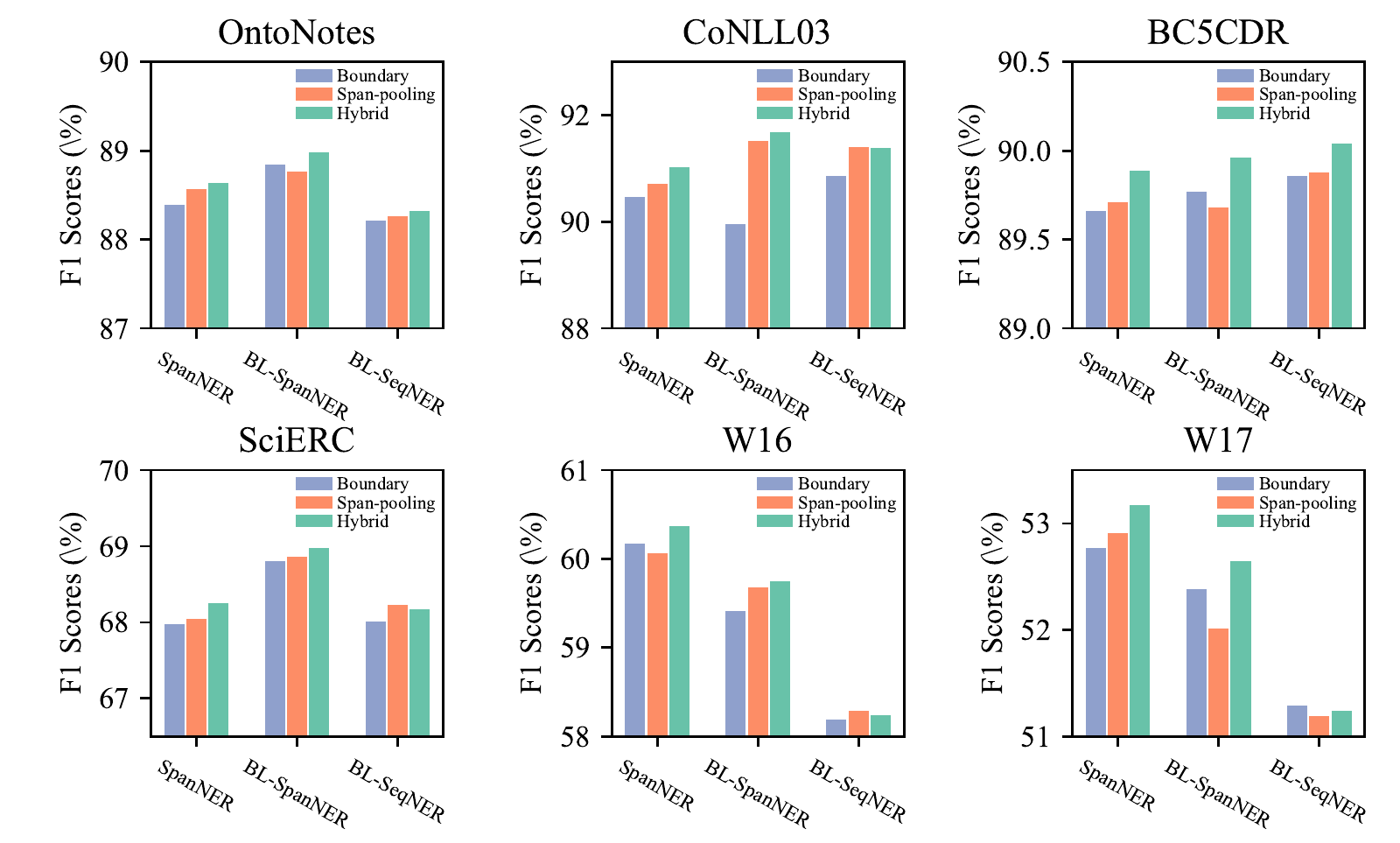}
\end{figure}

We report the exploration results in Figure \ref{spanrepresentation}, from which we can see that: (1) In SpanNER and BL-SpanNER, the hybrid method shows consistent superiority across the six datasets. And in SeqNER, the hybrid method performs the best on two of the six datasets. (2) In SpanNER, the Span-pooling method performs better than the boundary method on five of the six datasets. And in BL-SpanNER, the boundary and span-pooling methods show comparable performance.

We attribute the superiority of the hybrid method to the fact that it combines the advantages of boundary and span-pooling methods, making it take the structure, boundary, length, and context of spans into consideration. Moreover, we find that the hybrid method does not require extra time overhead compared to the other two methods. Based on the above facts, we use the hybrid method to generate span representations in all the other experiments.

\section{Conclusion}
In this paper, we explore three critical issues of the BL paradigm, a win-win NER approach achieved by bundling sequence labeling-based and span-based models. 
We first design two NER models: SeqNER and SpanNER, and then we bundle them together to form a BL enhanced model: BL-NER, to answer the first two issues, i.e., when and why BL works.
Experimental results on eleven datasets validate the effectiveness of BL, and detailed analyses provide answers to both issues. 
In order to address the third issue, i.e., whether BL enhances existing SOTA NER models, we apply BL to five previous NER models. 
Extensive results indicate that BL consistently improves their performance. 
Moreover, we find that SeqNER leads to fewer entity boundary prediction errors than SpanNER, but more entity type prediction errors. Furthermore, we make comparisons between two label tagging methods, as well as three span representations.

\bibliographystyle{IEEEtran}
\bibliography{custom}
\end{document}